\documentclass[letterpaper]{article}
\usepackage{aaai25}  % DO NOT CHANGE THIS
\usepackage{times}  % DO NOT CHANGE THIS
\usepackage{helvet}  % DO NOT CHANGE THIS
\usepackage{courier}  % DO NOT CHANGE THIS
\usepackage[hyphens]{url}  % DO NOT CHANGE THIS
\usepackage{graphicx} % DO NOT CHANGE THIS
\usepackage{natbib}  % DO NOT CHANGE THIS AND DO NOT ADD ANY OPTIONS TO IT
\usepackage{caption} % DO NOT CHANGE THIS AND DO NOT ADD ANY OPTIONS TO IT
\usepackage{algorithm}
\usepackage{algorithmic}

\usepackage{newfloat}
\usepackage{listings}
\DeclareCaptionStyle{ruled}{labelfont=normalfont,labelsep=colon,strut=off} % DO NOT CHANGE THIS
\floatstyle{ruled}
\newfloat{listing}{tb}{lst}{}
\floatname{listing}{Listing}
\usepackage{mdframed}
\usepackage[utf8]{inputenc} % allow utf-8 input
\usepackage[T1]{fontenc}    % use 8-bit T1 fonts
\usepackage{url}            % simple URL typesetting
\usepackage{booktabs}       % professional-quality tables
\usepackage{nicefrac}       % compact symbols for 1/2, etc.
\usepackage{microtype}      % microtypography
\usepackage{xcolor}         % colors

\usepackage{graphicx}
\usepackage{subfigure}
\usepackage{marvosym}
\usepackage{threeparttable}
\usepackage{amsthm}

\usepackage{amsmath}
\usepackage{multirow} 
\usepackage{multicol} 

\usepackage{caption}

\usepackage{enumitem}

\usepackage{bm}
\usepackage{pifont}
\usepackage{amssymb}
\usepackage{todonotes}

\title{LLM-TS Integrator: 

Integrating LLM for Enhanced Time Series Modeling}

\author{
    Can (Sam) Chen\thanks{Work done during an internship at Borealis AI.},
    Gabriel L. Oliveira,
    Hossein Sharifi-Noghabi,
    Tristan Sylvain
}
\affiliations{
    Borealis AI\\
}

\begin{document}

\maketitle

\begin{abstract}
Time series~(TS) modeling is essential in dynamic systems like weather prediction and anomaly detection.
Recent studies utilize Large Language Models (LLMs) for TS modeling, leveraging their powerful pattern recognition capabilities.
These methods primarily position LLMs as the predictive backbone, often omitting the mathematical modeling within traditional TS models, such as periodicity.
However, disregarding the potential of LLMs also overlooks their pattern recognition capabilities.
To address this gap, we introduce \textit{LLM-TS Integrator}, a novel framework that effectively integrates the capabilities of LLMs into traditional TS modeling.
Central to this integration is our \textit{mutual information} module.
The core of this \textit{mutual information} module is a traditional TS model enhanced with LLM-derived insights for improved predictive abilities.
This enhancement is achieved by maximizing the mutual information between traditional model's TS representations and LLM's textual representation counterparts, bridging the two modalities.
Moreover, we recognize that samples vary in importance for two losses: traditional prediction and mutual information maximization.
To address this variability, we introduce the \textit{sample reweighting} module to improve information utilization.
This module assigns dual weights to each sample: one for prediction loss and another for mutual information loss, dynamically optimizing these weights via bi-level optimization.
Our method achieves state-of-the-art or comparable performance across five mainstream TS tasks, including short-term and long-term forecasting, imputation, classification, and anomaly detection. 
%Our code is available at: \url{https://anonymous.4open.science/r/llm_ts_anonymous-F07D/README.MD}
\end{abstract}

\vspace{-.3cm}

\section{Introduction}
%
% \tristan{TODOs:
% \begin{itemize}
%   %\item[$\Square$] Tables should be stand-alone. This means all captions should include some interpretation of results.
%   %\item[$\Square$] Table 4 fit to page or shrink.
%   %\item[$\Square$] Related Work on sample reweighting.
%   %\item[$\Square$] Fig 1 caption expand.
%   %\item[$\Square$] Fig 2. I like the figure, but I would probably try to rescale a bit the amplitude of each point (e.g. exponential rather than linear scale?) to add more differentiability between the methods.
%   %\item[$\bigstar$] Comment: For eq (6) we can probably put all the 3 gradient descent update equations in the same line. what do you think? \Sam{I do not understand. What are three grad update?}
%   %\item[$\bigstar$] Comment: We are taking a lot of space in sec 2 to discuss TimesNet. I get where this is coming from but we may want to make this a bit more concise. What do you think? \Sam{agree}
%   %\item[$\bigstar$] Comment: We have too many figures. Thoughts on pruning, and which to prune? \Sam{I suppose no need to prune. Just place them in appendix}
%   %\item[$\bigstar$] Comment: Posted an intro candidate. what do you think? \Sam{Merged.}
% \end{itemize}
% }
% \hossein{We should mention that all results are averaged over x number of runs with seed values in appendix}\Sam{got u. It is in main paper now.}

% \hossein{We should have a footnote for LLM methods that we needed to rerun due to longer history compared to timesnet setup}\Sam{got u. Will mention it in the main paper.}

Time series (TS) modeling, as emphasized in \cite{hyndman2018forecasting}, is crucial for a variety of real-world applications. It is instrumental in forecasting meteorological factors for weather prediction \cite{wu2021autoformer}, imputing missing data in economic TS \cite{friedman1962interpolation}, detecting anomalies in industrial monitoring data for maintenance \cite{gao2020robusttad}, and classifying trajectories for action recognition \cite{franceschi2019unsupervised}.
 Given its significant practical impact, TS analysis continues to attract substantial attention~\cite{Lim2021TimeseriesFW, wen2022robust}.

Recent efforts in TS modeling have increasingly adopted Large Language Models (LLMs) to leverage their exceptional pattern recognition capabilities~\cite{jiang2024empowering, zhou2024one, jin2024timellm, sun2023test, gruver2024large, cao2023tempo}.
While these innovative approaches validate the potential of LLMs in TS modeling, they primarily position LLMs as the core predictive model.
Consequently, they often omit the mathematical modeling tailored specifically for TS models, such as employing the Fourier Transform to capture periodic patterns \cite{wu2023timesnet}.

On the other hand, fully disregarding the potential of LLMs also overlooks their powerful pattern recognition capabilities. It is important to recognize the balance between leveraging LLMs for their advanced capabilities and utilizing traditional TS models for their mathematical modeling, to enhance the overall performance and accuracy of TS predictions.
In response, we propose \textit{LLM-TS Integrator}, a novel framework that effectively integrates the capabilities of LLMs into traditional TS modeling.

\begin{figure*}[t!]
    \centering
    \includegraphics[scale=0.35]{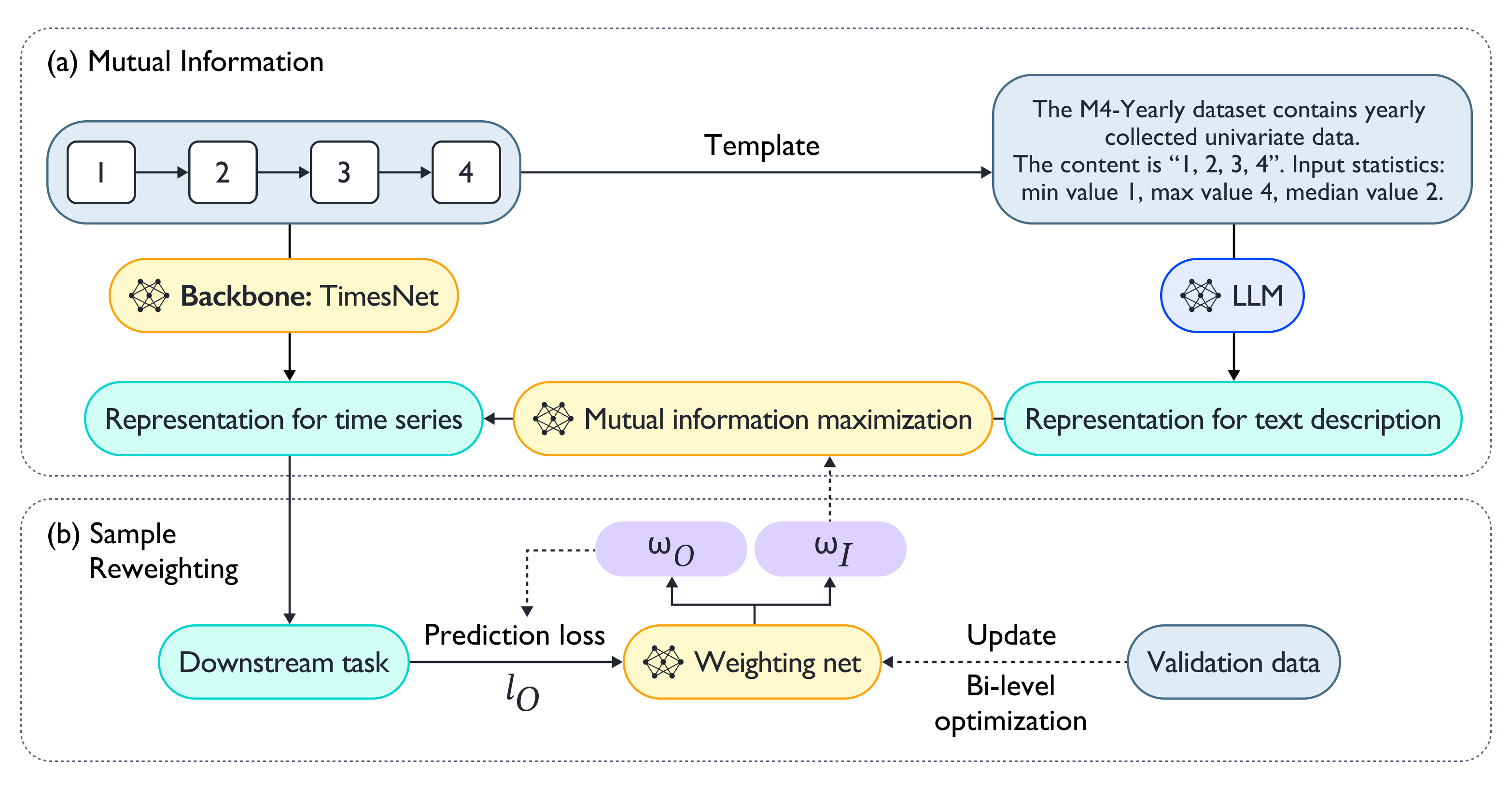}
    \caption{Illustration of \textit{LLM-TS Integrator}.
    Module (a) enhances the traditional TS model~(TimesNet) with LLM-derived insights by mutual information maximization.
    Module (b) optimizes sample importance for both prediction loss and mutual information loss to improve information utilization.
    %\hossein{If we have space I would add more text to this caption basically if someone reads the abstract and look at this figure they should have a rough idea about the framework} \gabriel{The caption should be as self-contained as possible.}
    }
    %\vspace{-10pt}
    \label{fig: overall_method}
\end{figure*}

Central to our framework is a \textit{mutual information} module, as depicted in Figure~\ref{fig: overall_method}(a).
The core of this module is a traditional predictive model, which we enhance with insights derived from LLMs to improve its predictive abilities.
In this work, we primarily utilize TimesNet~\cite{wu2023timesnet} as the traditional predictive model due to its exceptional performance and insight into periodic modeling, and our framework is also applicable to other traditional TS models (see in Section~\ref{subsec: model_analysis}).
%
% \hossein{perhaps we can say: Although in this work we primarily utilize TimesNet~\cite{wu2023timesnet} as the traditional predictive model due to its exceptional performance and insight into periodic modeling, our framework is also applicable to other traditional TS models (see in Section~\ref{subsec: model_analysis}).}
%
We achieve this enhancement by maximizing the mutual information~\cite{sun2019infograph} between the TS representations from traditional models and their textual counterparts from LLMs, thereby bridging these two modalities.
Despite its established use, mutual information maximization has not been previously applied to the intersection of TS and text domain.
With textual descriptions often missing from TS data, we propose generating such descriptions via a carefully designed template.
This template is enriched with essential background and statistical details pertinent to the TS, thereby enriching the LLM's comprehension of the TS context. 
%

%\hossein{Should we briefly cite our results here? something like we outperformed baselines in task A (see section ??) etc.}

Our first module introduces a dual loss framework: traditional prediction and mutual information, and we recognize that the importance of samples differs between the two losses.
For instance, a large prediction loss for a sample highlights its learning potential, emphasizing the need to focus on its prediction loss.
This scenario also implies that the model’s learning for this sample is inadequate and its hidden representation is suboptimal for mutual information computation.
Consequently, the sample's contribution to the mutual information calculation should be reduced.
To manage this variability, we have introduced a novel \textit{sample reweighting} module powered by a simple MLP~(multilayer perceptron) network, as depicted in Figure~\ref{fig: overall_method}(b).
This module processes the sample prediction loss to produce dual weights for each sample, one for the prediction loss and another for the mutual information loss. 
These weights are optimized through bi-level optimization, thereby enhancing the efficacy of information utilization.

% \begin{wrapfigure}[13]{r}{0.5\textwidth}
%     \centering
%     \includegraphics[scale=0.30]{Figures/overall_perf.pdf}
%     \caption{Comparative analysis of model performance across different tasks.}
%     \label{fig: overall_perf}
% \end{wrapfigure}

Our primary contributions are as follows:
\begin{itemize}[leftmargin=*]
\item We introduce \textit{LLM-TS Integrator}, which consists of \textit{mutual information} and \textit{sample reweighting}.
The first module enhances traditional TS modeling with capabilities from LLMs through mutual information maximization.
\item The second module optimizes sample importance for both prediction loss and mutual information loss, which improves information utilization.
\item Extensive experiments across five mainstream TS tasks—short-term and long-term forecasting, imputation, classification, and anomaly detection—demonstrate the effectiveness of our framework.
\end{itemize}

% \hossein{so far you are using "our framework" and "our model" to refer to your work - I suggest to pick one and stick to it or if you want to use both make sure it is used in the right type for example if you are talking about loss or optimizing parameters model is the better word but if you are talking about general aspect then framework is a better word}
% \Sam{let's use our framework for consistency. I may also use our method for word variation.}
%\input{sections/introduction2.tex}

\section{Preliminaries}
\label{sec: pre}
\noindent\textbf{TimesNet.}
In this paper, we mainly choose TimesNet as the traditional predictive model due to its exceptional performance~\cite{wu2023timesnet} and also explore other additional traditional models including ETSformer~\cite{woo2022etsformer}, Stationary~\cite{liu2022non}, and FreTS~\cite{yi2024frequency} in Section~\ref{subsec: model_analysis}.
Previous studies to modeling temporal variations in 1D time series often struggle with complex temporal patterns. TimesNet addresses this challenge by decomposing these complex variations into multiple intra-period and inter-period variations. This is achieved by transforming the 1D time series into a series of 2D tensors, each corresponding to different periods.
%
% Given a 1D time series $\mathbf{X}_{\text{1D}}$, TimesNet begins by determining period lengths via the following FFT operation:
% %\tristan{Specify the period operator, i.e. FFT.}
% \begin{equation}
%     \mathbf{A},\{f_{1},\cdots,f_{k}\},\{p_{1},\cdots,p_{k}\}=\operatorname{Period}\left({\mathbf{X}}_{\text{1D}}\right)
% \end{equation}
% In this equation, $\mathbf{A}$ represents the amplitudes, while $f_{i}$ and $p_{i}$ denote the frequencies and period lengths, respectively.
% %
% The next step involves reshaping the 1D time series into a 2D format for each identified period, as shown below:
% %
% \begin{equation}
%     \mathbf{X}^{i}_{\text{2D}} =\operatorname{Reshape}_{p_{i},f_{i}}\left(\operatorname{Padding}(\mathbf{X}_{\text{1D}})\right),\ i\in\{1,\cdots, k\}
% \end{equation}
% %
% These 2D tensors are then processed using an Inception block:
% \begin{equation}
%     \widehat{\mathbf{X}}^{i}_{\text{2D}} =\operatorname{Inception}\left(\mathbf{X}^{i}_{\text{2D}}\right),\ i\in\{1,\cdots, k\}
% \end{equation}
% %
% The output from the Inception block is subsequently converted back into a 1D format:
% \begin{equation}
%     \widehat{\mathbf{X}}^{i}_{\text{1D}} =\operatorname{Trunc}\left(\operatorname{Reshape}_{1,(p_{i}\times f_{i})}\left(\widehat{\mathbf{X}}^{i}_{\text{2D}}\right)\right),\ i\in\{1,\cdots, k\}
% \end{equation}
%
% By employing this method, TimesNet adeptly identifies and encapsulates the nuanced variations within and between periods. This sophisticated modeling forms the cornerstone of our predictive framework as detailed in this study.
%
For the time series $\boldsymbol{x}$,  we derive its representation $\boldsymbol{h}^{m}_{\boldsymbol{\theta}}(\boldsymbol{x})$ using the TimesNet model parameterized by $\boldsymbol{\theta}$ where $m$ represents \textit{model}.
%\gabriel{Did we define $m$?}

\noindent\textbf{Large Language Models.}
Language models are trained on extensive collections of natural language sequences, with each sequence consisting of multiple tokens.
Notable large language models such as GPT-3~\cite{brown2020language} and Llama2~\cite{touvron2023llama} aim to predict the next token based on preceding tokens, demonstrating their capabilities through improvements in model parameter size and the amount of training data.
Each language model uses a tokenizer that breaks down an input string into a sequence of recognizable tokens.
However, the training of current large language models is solely focused on natural language, not encompassing time series data.
This limitation presents challenges for the direct application of large language models to time series analysis.

\vspace{-10pt}
\section{Method}
In this section, we present the \textit{LLM-TS Integrator} framework, which effectively integrates the capabilities of LLMs into traditional TS modeling. 
This framework consists of two modules: \textit{mutual information} and \textit{sample reweighting}.
The first module enhances a traditional TS model with LLM-derived insights for improved predictive abilities, as explored in Section~\ref{subsec: mutual}.
The second module optimizes weights for prediction loss and mutual information loss via bi-level optimization, improving information utilization, as covered in Section~\ref{subsec: reweight}. 
The overall algorithm is shown in Algorithm~\ref{alg: overall}.

\begin{algorithm}[tb]
\caption{\textbf{LLM-TS Integrator}}\label{alg: overall}
\textbf{Input}: The TS dataset $\mathcal{D}$, number of training iterations $T$.\\
\textbf{Output}: Trained TS model parameterized by $\boldsymbol{\theta}^{*}$.

\begin{algorithmic}[1]
\STATE \texttt{\textcolor{gray}{/* \textit{Mutual Information Module} */}}
\STATE Train a traditional TS model~(e.g., TimesNet) parameterized by $\boldsymbol{\theta}$ using $\mathcal{D}$.
\STATE Generate text description $\boldsymbol{t}$ for TS sample $\boldsymbol{x}$ via a designed template.
\STATE Derive hidden representations $\boldsymbol{h}_{\boldsymbol{\theta}}^m(\boldsymbol{x})$ from the TS model and $\boldsymbol{h}^l(\boldsymbol{t})$ from the LLM.
\WHILE{$\tau <= T-1$}
\STATE Sample $\boldsymbol{x}$, $\boldsymbol{t}$, $\boldsymbol{y}$ from $\mathcal{D}$, where $\boldsymbol{y}$ are the labels.
\STATE Optimize a discriminator model $T_{\boldsymbol{\beta}}$ to estimate mutual information as per Eq~.(\ref{eq: discriminator}).
\STATE \texttt{\textcolor{gray}{/* \textit{Sample Reweighting Module} */}}
\STATE Process sample loss $l_O$ with the weighting net to produce dual weights as per Eq.~(\ref{eq: weight1}),~(\ref{eq: weight2}).
\STATE Adopt bi-level optimization to update the weighting net following Eq.~(\ref{eq: bilevel1}),~(\ref{eq: bilevel2}).
\STATE Re-calculate dual weights using the updated weighting net per Eq.~(\ref{eq: weight1}),~(\ref{eq: weight2}).
\STATE Calculate the overall loss to update the TS model as per Eq.~(\ref{eq: overall_loss}).
\ENDWHILE
\STATE Return the trained TS model parameterized by $\boldsymbol{\theta}^{*}$.
\end{algorithmic}
\end{algorithm}

\subsection{Mutual Information}
\label{subsec: mutual}

% \gabriel{probably the third reference is from another paper, right? Otherwise just remove it.} 

% Below old version, then new version to replace it.
% Prior research~\cite{zhou2024one, jin2023time} at the nexus of Large Language Models (LLMs) and TS analysis has predominantly cast LLMs as the core predictive mechanism, often overlooking the essential mathematical principles fundamental to traditional TS models. 

% Previous studies~\cite{zhou2024one, jin2023time} focusing on the integration of Large Language Models (LLMs) with TS analysis have mainly emphasized the use of LLMs as primary predictive tools. This approach frequently neglects the specificities of time-series as a modality.

% %
% Contrary to this approach, our methodology leverages a traditional model as the primary predictive framework, augmenting it with the sophisticated insights offered by LLMs. 
% %
% In this paper, we adopt the TimesNet detailed in Section~\ref{sec: pre} as the prediction backbone and we also explore the transformer backbone in Appendix.
% %
% This hybrid approach allows us to harness the strengths of both worlds.
% %
% This enrichment is facilitated by maximizing the mutual information between the TS representations generated by the traditional model and their corresponding textual representations obtained from LLMs. 

Previous studies~\cite{zhou2024one, jin2023time} have predominantly highlighted the use of Large Language Models (LLMs) as the core predictive model in the TS analysis, often omitting the mathematical modeling within traditional TS models, such as periodicity.

In contrast, our framework utilizes a traditional TS model as the predictive backbone, enhanced by the advanced capabilities of LLMs.
In this paper, we employ TimesNet, as outlined in Section~\ref{sec: pre}, as the traditional predictive model, and we further examine other models in Section~\ref{subsec: model_analysis}.
This hybrid methodology combines the advantages of both traditional TS models and modern LLMs.
We achieve this integration via a \textit{mutual information} module, which maximizes the mutual information between the TS data representations derived from the traditional model and their corresponding textual representations derived from LLMs.

\subsubsection{Mutual Information Estimation.}
Estimating the mutual information between hidden representations of a time series (TS) sample $\boldsymbol{x}$ and its corresponding textual description $\boldsymbol{t}$ is essential. For the TS sample $\boldsymbol{x}$, we derive its representation $\boldsymbol{h}^{m}_{\boldsymbol{\theta}}(\boldsymbol{x})$ using TimesNet, a \textbf{m}odel parameterized by $\boldsymbol{\theta}$. For the text $\boldsymbol{t}$, its representation $\boldsymbol{h}^{l}(\boldsymbol{t})$ is extracted using a pre-trained Large Language Model (LLM), where $l$ denotes the language model. In this study, we employ the LLaMA-3b model~\cite{touvron2023llama} as our primary LLM, while also evaluating other LLMs as detailed in Section~\ref{subsec: model_analysis}.

We estimate mutual information using the Jensen-Shannon MI estimator~\cite{sun2019infograph, nowozin2016f} and additionally explore the MINE estimator~\cite{hjelm2018learning} as detailed in Appendix~\ref{appendix: further_ablation}. 
Specifically, let $(\boldsymbol{x}, \boldsymbol{t})$ represent a sample from the TS set $\mathbb{S}$, and $(\boldsymbol{\tilde{x}}, \boldsymbol{\tilde{t}})$ denote a sample from $\mathbb{\tilde{S}} = \mathbb{S}$ where $(\boldsymbol{x}, \boldsymbol{t})\neq (\boldsymbol{\tilde{x}}, \boldsymbol{\tilde{t}})$.
Within this context, $\mathbb{S}$ denotes the TS training distribution while the product ${\mathbb{S}\times\mathbb{\tilde{S}}}$ represents pairs of distinct samples within $\mathbb{S}$.
Then the lower bound of mutual information can be estimated as:
\begin{equation}
    \label{eq: mutual_comp}
    \begin{aligned}
    I(\boldsymbol{\theta}, \boldsymbol{\beta}) =  \mathbb{E}_{\mathbb{S}}[-{sp}(-{T}_{\boldsymbol{\beta}}(\boldsymbol{h}^{m}_{\boldsymbol{\theta}}(\boldsymbol{x}), \boldsymbol{h}^{l}(\boldsymbol{t}))] - \\
    \mathbb{E}_{\mathbb{S}\times\mathbb{\tilde{S}}}[{sp}({T}_{\boldsymbol{\beta}}(\boldsymbol{h}^{m}_{\boldsymbol{\theta}}(\boldsymbol{x}), \boldsymbol{h}^{l}(\boldsymbol{\tilde{t}}))]\,,
    %\label{eq: mututual}
    \end{aligned}
\end{equation}
where ${T}_{\boldsymbol{\beta}}$ denotes the discriminator parameterized by $\boldsymbol{\beta}$ and $sp$ is the softplus function.
For the details of ${T}_{\beta}$, we feed the positive and negative examples into a 1-layered fully-connected network, and then output the dotproduct of the two representations.
The mutual information estimation begins by fixing the model parameters $\boldsymbol{\theta}$, followed by training $\boldsymbol{\beta}$ as the estimator. 
Specifically, we optimize $\boldsymbol{\beta}$ via the following:
\begin{equation}
    \label{eq: discriminator}
    \hat{\boldsymbol{\beta}} = \boldsymbol{\beta} - \eta_0 \cdot \frac{\partial I(\boldsymbol{\theta}, \boldsymbol{\beta})}{\partial \boldsymbol{\beta}}
\end{equation}
where $\eta_0$ denotes the learning rate.
Subsequently, we refine the model parameters $\boldsymbol{\theta}$ to maximize mutual information, thereby enriching the traditional TS model with insights derived from LLMs. This alternating optimization procedure between model and discriminator is repeated each epoch.

\subsubsection{Text Description for Time Series.}
In our approach, we assume each time series (TS) sample $\boldsymbol{x}$ is paired with a corresponding textual description, $\boldsymbol{t}$.
However, textual descriptions are frequently unavailable for many TS datasets. To bridge this gap, we introduce a methodology for generating textual descriptions of TS data.
We propose creating textual representations that capture essential background and statistical details of the TS inspired by~\cite{jin2023time}. This process can be systematically implemented using the following carefully designed template template:
\begin{quote}
\begin{scriptsize}\begin{verbatim}
template = (
    "{task_description}. The content is: {TS}. "
    "Input statistics: 
    min value {min(TS)}, max value {max(TS)}, "
    "median value {median(TS)}, 
    top 5 lags {compute_lags(TS)}."
)
\end{verbatim}\end{scriptsize}
\end{quote}

%\Sam{The way we display this template can be better.}
% \tristan{Proposed change, lmk what you think} \hossein{I really like this design of the template} \gabriel{I liked the current version.}

%
% \textbf{How to obtain text desciption for 2D data}
% %
% To enrich the textual descriptions with detailed statistical insights, we utilize the tsfresh Python package. This tool is distinguished for its exhaustive feature extraction capabilities tailored specifically for TS analysis. It integrates a plethora of algorithms from various domains such as statistics, TS analysis, signal processing, and nonlinear dynamics. Moreover, it includes an efficient feature selection process to enhance the relevance and quality of the generated features.

\subsection{Sample Reweighting}
\label{subsec: reweight}

Our \textit{mutual information} module introduces two distinct loss functions: (1) the original sample prediction loss, ${l}_{O}(\boldsymbol{x}, \boldsymbol{y})$, hereafter referred to as ${l}_{O}$, which corresponds to the prediction loss for a TS sample $\boldsymbol{x}$ and its label $\boldsymbol{y}$, and (2) the mutual information maximization loss, denoted as $-I(\boldsymbol{\theta}, \boldsymbol{\beta})$. We acknowledge that the significance of samples varies between these two losses. Specifically, a large prediction loss $l_O$ indicates a sample's substantial learning potential, thereby justifying a higher weight $\omega_{O}$ for its prediction loss. Conversely, this suggests that the sample's representation may be suboptimal for mutual information computation, warranting a lower weight $\omega_{I}$.

To address this disparity, we have developed a novel \textit{sample reweighting} module, described as follows.

\subsubsection{Weighting Network.}
To automate weight assignment, our module employs a two-layer MLP network parameterized by $\boldsymbol{\alpha}$, which processes the sample prediction loss to produce a pair of weights:
\begin{equation}
    \label{eq: weight1}
    \omega_O(\boldsymbol{\alpha}), \omega_I(\boldsymbol{\alpha}) = MLP_{\boldsymbol{\alpha}}(l_{O})
\end{equation}
This process involves converting the sample loss $l_{O}$ into a latent code $z$ through a hidden layer. The network then outputs dual weights:
\begin{equation}
    \label{eq: weight2}
    \omega_O(\boldsymbol{\alpha}), \omega_I(\boldsymbol{\alpha}) = \sigma(m_{O}\cdot z), \sigma(m_{I}\cdot z)
\end{equation}
where $m_{O} > 0$ and $m_{I} < 0$ to ensure a negative correlation between $\omega_O$ and $\omega_I$. The function $\sigma(\cdot)$ denotes sigmoid.

For a batch of $N$ samples, the weight vector $\boldsymbol{\omega_{O}(\alpha)} \in \mathbb{R}^{N}$ is directly applied to the original prediction loss vector $\boldsymbol{l_{O}} \in \mathbb{R}^{N}$, resulting in the weighted average loss calculated as ${mean}(\boldsymbol{\omega_O}(\boldsymbol{\alpha}) \cdot \boldsymbol{l_{O}})$.
Similarly, the weight vector $\boldsymbol{\omega_I(\alpha)} \in \mathbb{R}^{N}$ not only reflects the overall significance of the mutual information but also each sample's individual contribution to this metric. The mean of these weights ${mean}(\boldsymbol{\omega_I}(\boldsymbol{\alpha}))$ represents the overall importance.
For mutual information computations, $\boldsymbol{\omega_I(\alpha)}$ is transformed into a probability distribution $p_{I}^{i} = \frac{\omega_{I}^{i}}{\sum_{i=1}^{N} \omega_{I}^{i}}$.
This adjustment affects the distribution used in mutual information calculations, necessitating a recalculation of mutual information as $I(\boldsymbol{\theta}, \boldsymbol{\beta}, \boldsymbol{\alpha})$, with details provided in Appendix~\ref{appendix: mutual}.
As a result, the overall loss is formulated as:
\begin{equation}
\begin{split}
    \label{eq: overall_loss}
    \mathcal{L}(\boldsymbol{\theta}, \boldsymbol{\alpha}) = mean(\boldsymbol{\omega_O(\alpha)} \cdot \boldsymbol{l_{O}}) \\
    + mean(\boldsymbol{\omega_I(\alpha)}) \cdot [-I(\boldsymbol{\theta}, \boldsymbol{\beta}, \boldsymbol{\omega_I(\alpha)})]
\end{split}
\end{equation}

\subsubsection{Bi-level Optimization.}
The ensuing challenge is optimizing the weighting network $\boldsymbol{\alpha}$.
We achieve this by
leveraging the supervision signals from a small validation dataset~\cite{chen2022gradient}.
If the weighting network is properly optimized, the model trained with these weights is expected to show improved performance on the validation dataset in terms of the validation loss $\mathcal{L}_{V}(\boldsymbol{\theta}) = \frac{1}{M}\sum_{j}^{M} {l}_{O}^j(\boldsymbol{x}_j, \boldsymbol{y}_j)$, where $M$ denotes the size of the validation set. This constitutes a bi-level optimization problem.
At the inner level, model training is conducted through gradient descent:
\begin{equation}
    \label{eq: bilevel1}
\hat{\boldsymbol{\theta}}(\boldsymbol{\alpha}) = \boldsymbol{\theta} - \eta_1 \cdot \frac{\partial \mathcal{L}(\boldsymbol{\theta}, \boldsymbol{\alpha})}{\partial \boldsymbol{\theta}}
\end{equation}
The objective is to ensure that the model performs optimally on the validation dataset:
\begin{equation}
    \label{eq: bilevel2}
    \hat{\boldsymbol{\alpha}} = \boldsymbol{\alpha} - \eta_2 \cdot \frac{\partial \mathcal{L}_V(\boldsymbol{\theta}(\boldsymbol{\alpha}))}{\partial \boldsymbol{\alpha}}
\end{equation}
Both $\eta_1$ and $\eta_2$ represent the learning rates for the respective optimization steps. 
Through the minimization of the validation loss, we aim to optimize the weighting network $\boldsymbol{\alpha}$.

\section{Experimental Results}
\label{sec: exp}
To demonstrate the versatility of our \textit{LLM-TS Integrator}, we conduct extensive experiments across five main tasks: short- and long-term forecasting, imputation, classification, and anomaly detection. 
To maintain experimental integrity, our methodology adheres to the setup in~\cite{wu2023timesnet}. 
We detail the experimental setting in Appendix~\ref{appendix: settings}.
%
%
%\hossein{perhaps make this clear that by setup you meant train and test split, hyper-parameters, optimizer, etc. to make it more clear based on the checklist}
%
% \Sam{Weighting network size, learning rates,bs, optimizer}

\subsubsection{Baselines.}
Our evaluation employs a comprehensive array of baseline models across several architectural designs
\textbf{(1)} CNN-based models, specifically TimesNet~\cite{wu2023timesnet};
\textbf{(2)} MLP-based models, including LightTS~\cite{zhang2022less} and DLinear~\cite{zeng2023transformers};
\textbf{(3)} Transformer-based models, such as Reformer~\cite{kitaev2020reformer}, Informer~\cite{zhou2021informer}, Autoformer~\cite{wu2021autoformer}, FEDformer~\cite{zhou2022fedformer}, Nonstationary Transformer~\cite{liu2022non}, ETSformer~\cite{woo2022etsformer}, and PatchTST~\cite{nie2022time};
\textbf{(4)} LLM-based models, represented by GPT4TS~\cite{zhou2024one}.
While we assess a wide range of models, we focus our discussion on the top-performing ones as highlighted in~\cite{zhou2024one}.

Additional comparisons for forecasting tasks include LLM-based models like Time-LLM~\cite{jin2023time} and TEST~\cite{sun2023test}. 
For short-term forecasting, models like N-HiTS~\cite{challu2023nhits} and N-BEATS~\cite{oreshkin2019n} are included. Anomaly detection tasks are assessed using Anomaly Transformer~\cite{xu2021anomaly}, and for classification tasks, models such as XGBoost~\cite{chen2016xgboost}, Rocket~\cite{dempster2020rocket}, LSTNet~\cite{lai2018modeling}, LSSL~\cite{gu2021efficiently}, Pyraformer~\cite{liu2021pyraformer}, TCN~\cite{franceschi2019unsupervised}, and Flowformer~\cite{huang2022flowformer} are considered.
This broad selection of baselines enables a rigorous and fair comparison across various tasks, highlighting the capabilities of our method.

\subsection{Main Results}
\begin{figure}[htb]
    \centering
    \includegraphics[scale=0.35]{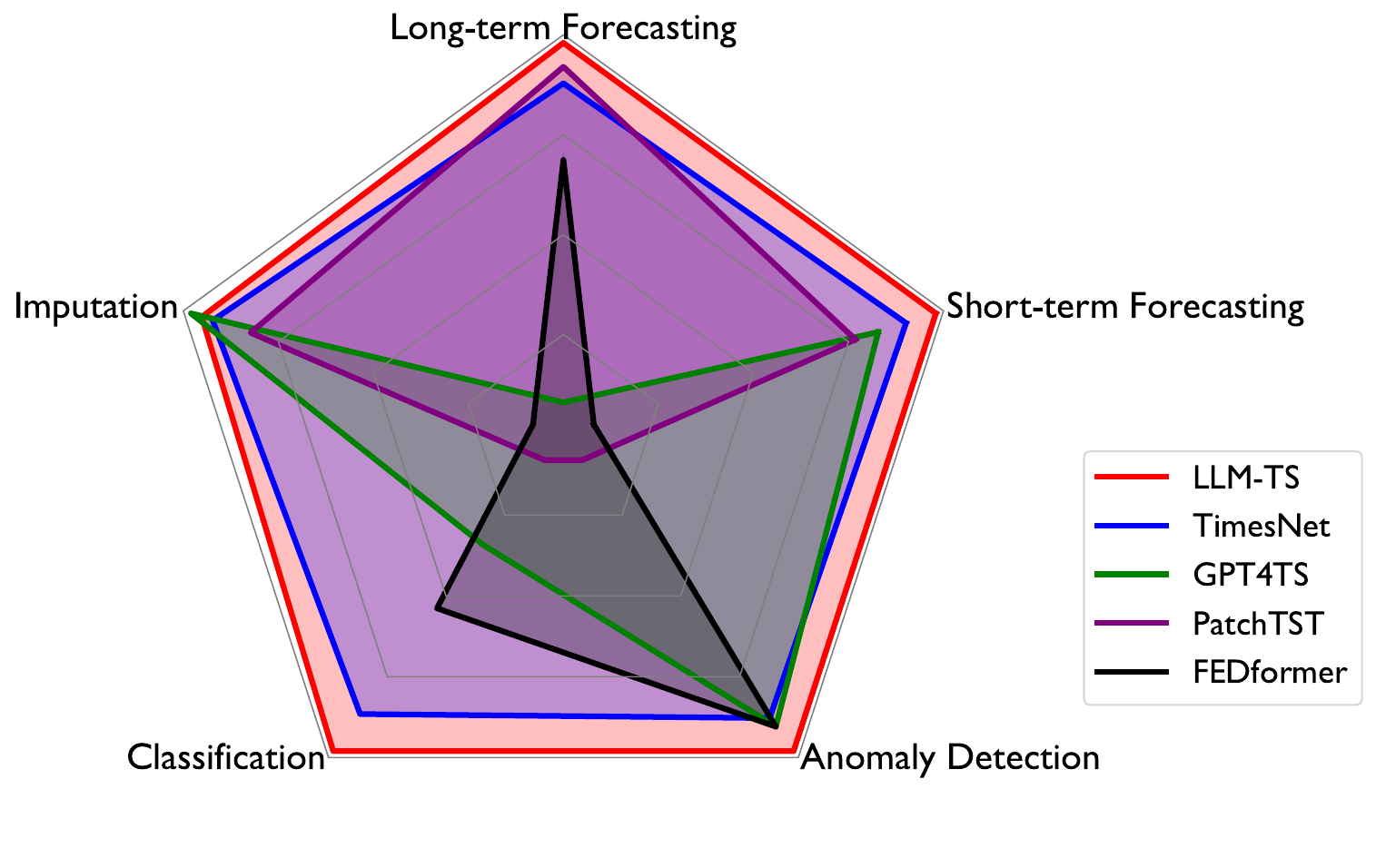}
    \vspace{-15pt}
    \caption{Model performance across different tasks. 
    %\gabriel{Figure 2 is not a bit too far from its citation, is it a strategy to show the figure that early in the paper?} 
    %\hossein{This figure is a bit overwhelming and kinda hard to compare - does it make sense to keep it focused to top 3 models on average and report detailed results in a table in appendix} \Sam{I follow previous work to report around 9 methods. They key here may be that not all models can be applied to five tasks. One method is good at one task and may not be able to apply to or not good at another task.}
    }
    \vspace{-5pt}
    \label{fig: overall_perf}
\end{figure}
Figure~\ref{fig: overall_perf} demonstrates that our \textit{LLM-TS Integrator} consistently outperforms other methods in various tasks, underscoring its efficacy.
We will refer to our method as \textit{LLM-TS} in the tables for brevity.
Unless otherwise indicated, we cite results from TimesNet~\cite{wu2023timesnet}.
We reproduce TimesNet and GPT4TS~\cite{zhou2024one} experiments for all tasks. 
All results are averages from three runs with different seeds. Standard deviations for ablation studies are detailed in Appendix~\ref{tab: ablation_std}.
The best results are highlighted in bold, with the second-best underlined.
We also (1) present several showcases of our method in Appendix~\ref{appendix: show} and (2) discuss the model efficiency in Appendix~\ref{appendix: efficiency}

% \hossein{For some section we mention the evaluation metric(s) and for some other they are reported in table caption. It's better to make it consistent to finish each evaluation setup with corresponding metrics}
%\Sam{I check TimesNet and GPT4TS. They do not introduce the metrics, and I suppose this is because this is common knowledge in this field.}

% \hossein{reminder to highlight which results are obtained from other resources and we are reporting their numbers}

\subsection{Short- and Long- Term Forecasting}
\subsubsection{Setup.} 
To comprehensively assess our framework's forecasting capabilities, we engage it in both short- and long-term forecasting settings.
In the realm of short-term forecasting, we utilize the M4 dataset \citep{M4team2018dataset}, which aggregates univariate marketing data on a yearly, quarterly, and monthly basis.
For long-term forecasting, we examine five datasets following~\cite{zhou2024one}: ETT \citep{haoyietal-informer-2021}, Electricity \citep{ecldata}, Traffic \citep{trafficdata}, Weather \citep{weatherdata}, and ILI \citep{ilidata}. We adhere to the TimesNet setting with an input length of $96$.
For LLM-based methods like GPT4TS and Time-LLM, which use different input lengths, we rerun the experiments using their code. For PatchTST, we cite the results from~\cite{wang2023timemixer}, as the original PatchTST uses an input length of $512$.
Due to shorter input lengths in this study compared to the original, the reported performance is lower.

\subsubsection{Results.} 
As shown in Tables~\ref{tab:short_term} and~\ref{tab:long_term}, our \textit{LLM-TS} performs exceptionally well in both short- and long-term settings. 
It consistently surpasses TimesNet, highlighting the effectiveness of incorporating LLM-derived insights.
Furthermore, it generally outperforms other LLM-based methods such as GPT4TS, TIME-LLM, and TEST, underscoring the advantages of integrating traditional TS modeling.\\

\begin{table*}[htb]
\captionsetup{font=small} 
\caption{Short-term M4 forecasting. The prediction lengths are in [$6$, $48$] and results are obtained by weighting averages across multiple datasets with varying sampling intervals.
Full results are in Appendix~\ref{appendix:short-term_full}.}
\label{tab:short_term}
\vspace{-5pt}
%\vskip 0.15in
\begin{center}
\begin{small}
\scalebox{0.7}{
\setlength\tabcolsep{3pt}
\begin{tabular}{c|ccccccccccc}
\toprule

Methods& LLM-TS & TimesNet & GPT4TS & TIME-LLM & TEST & PatchTST & N-HiTS&N-BEATS& FEDformer &Stationary &Autoformer  \\

\midrule

SMAPE &\boldsymbol{$ 11.819 $}&$11.908  $&$ {11.991} $&$ 11.983 $&$ 11.927 $&$ 12.059$&$ 11.927$&\underline{$11.851$}&$12.840 $&$12.780 $&$12.909$ \\
MASE &\boldsymbol{$ 1.588 $}&$ 1.612 $&$ {1.600}$&\underline{$ 1.595 $}&$  1.613$&$1.623 $&$ 1.613 $&$1.599$&$1.701 $&$1.756 $&$1.771$ \\
OWA &\boldsymbol{$0.851$} &$0.860$ & $0.861$ & $0.859$ & $0.861$ & $0.869$ & $0.861$ & \underline{$0.855$} & $0.918$ & $0.930$ & $0.939$ \\

\bottomrule

\end{tabular}
}
\end{small}
\end{center}
\end{table*}

\vspace{-10pt}
\begin{table*}[htb]
\captionsetup{font=small} 
\caption{Long-term forecasting: Averages over $4$ lengths: {$24$, $36$, $48$, $60$} for ILI, and {$96$, $192$, $336$, $720$} for others. Full results in Appendix~\ref{appendix:long-term_full}.}
\label{tab:long_term}
\vspace{-10pt}
\begin{center}
\begin{small}
\scalebox{0.66}{
\setlength\tabcolsep{3pt}
\begin{tabular}{c|cc|cc|cc|cc|cc|cc|cc|cc|cc|cc}
\toprule

\multirow{2}{*}{Methods} 
&\multicolumn{2}{c|}{LLM-TS}
&\multicolumn{2}{c|}{TimesNet}  & \multicolumn{2}{c|}{TIME-LLM} & \multicolumn{2}{c|}{DLinear} & \multicolumn{2}{c|}{PatchTST} &\multicolumn{2}{c|}{GPT4TS}&\multicolumn{2}{c|}{FEDformer}&\multicolumn{2}{c|}{TEST}&\multicolumn{2}{c|}{Stationary}&\multicolumn{2}{c}{ETSformer} \\
&MSE&MAE&MSE&MAE&MSE&MAE&MSE&MAE&MSE&MAE&MSE&MAE&MSE&MAE&MSE&MAE&MAE&MSE&MAE&MSE \\

\midrule

Weather &\boldsymbol{$ 0.257 $}&\boldsymbol{$ 0.285 $}&\underline{$ 0.265$}&$ 0.290 $&$0.279 $&$ 0.296 $&\underline{$ 0.265 $}&$ 0.317 $&\underline{$ 0.265 $}&\boldsymbol{$ 0.285 $}&$ 0.275 $&$ 0.292 $&$ 0.309 $&$ 0.360 $&$ 0.291 $&$ 0.315 $&$ 0.288 $&$ 0.314 $&$0.271 $&$ 0.334$\\
ETTh1   &$ 0.454 $&\boldsymbol{$ 0.451 $}&$ 0.470$&$ 0.462 $&$0.474 $&$ 0.459 $&$ 0.456 $&$ 0.452 $&$ 0.516 $&$ 0.484 $&$ 0.473 $&\boldsymbol{$ 0.451 $}&\boldsymbol{$ 0.440 $}&$ 0.460 $&\boldsymbol{$ 0.440 $}&$ 0.460 $&$ 0.570 $&$ 0.537 $&$ 0.542$&$ 0.510$\\
ETTh2   &$ 0.396 $&\underline{$ 0.413 $}&$ 0.413$&$ 0.426 $&$0.398 $&$ 0.415 $&$ 0.559 $&$ 0.515 $&\underline{$ 0.391 $}&\boldsymbol{$ 0.411 $}&\boldsymbol{$ 0.383 $}&$ 0.410 $&$ 0.437 $&$ 0.449 $&$ 0.414 $&$ 0.432 $&$ 0.526 $&$ 0.516 $&$ 0.439$&$ 0.452$\\
ETTm1   &\boldsymbol{$ 0.401 $}&$ 0.409 $&$ 0.414$&$ 0.418 $&$0.437 $&$ 0.421 $&$ 0.403 $&\underline{$ 0.407 $}&$ 0.406 $&$ 0.407 $&$ 0.408 $&\boldsymbol{$ 0.400 $}&$ 0.448 $&$ 0.452 $&\underline{$ 0.402 $}&$ 0.411 $&$ 0.481 $&$ 0.456 $&$ 0.429$&$ 0.425$\\
ETTm2   &$ 0.295 $&\boldsymbol{$ 0.331 $}&$ 0.294$&\boldsymbol{$ 0.331 $}&$0.298 $&$ 0.342 $&$ 0.350 $&$ 0.401 $&\boldsymbol{$ 0.290 $}&$ 0.334 $&\boldsymbol{$ 0.290 $}&$ 0.335 $&$ 0.305 $&$ 0.349 $&$ 0.323 $&$ 0.359 $&$ 0.306 $&$ 0.347 $&$ 0.293$&$ 0.342$\\
ILI     &\boldsymbol{$ 1.973 $}&\boldsymbol{$ 0.894 $}&$ 2.266$&$ 0.974 $&$2.726 $&$ 1.098 $&$ 2.616 $&$ 1.090 $&$ 2.184 $&\underline{$ 0.906 $}&$ 5.117 $&$ 1.650 $&$ 2.847 $&$ 1.144 $&$ 3.324 $&$ 1.232 $&\underline{$ 2.077 $}&$ 0.914 $&$ 2.497$&$ 1.004$\\
ECL     &\underline{$ 0.194 $}&$ 0.299 $&$ 0.198$&\underline{$ 0.298 $}&$0.229 $&$ 0.315 $&$ 0.212 $&$ 0.300 $&$ 0.216 $&$ 0.318 $&$ 0.206 $&\boldsymbol{$ 0.285 $}&$ 0.214 $&$ 0.327 $&$ 0.237 $&$ 0.324 $&\boldsymbol{$ 0.193 $}&$ 0.296 $&$ 0.208$&$ 0.323$\\
Traffic &$ 0.618 $&\boldsymbol{$ 0.333 $}&$ 0.627$&$ 0.335 $&$0.606 $&$ 0.395 $&$ 0.625 $&$ 0.383 $&\boldsymbol{$ 0.529 $}&$ 0.341 $&\underline{$ 0.561 $}&$ 0.373 $&$ 0.610 $&$ 0.376 $&$ 0.581 $&$ 0.388 $&$ 0.624 $&\underline{$ 0.340 $}&$ 0.621$&$ 0.396$\\
\midrule

Average &\boldsymbol{$ 0.574 $}&\boldsymbol{$ 0.427 $}&$ 0.618$&$ 0.442 $&$0.681 $&$ 0.468 $&$ 0.686 $&$ 0.483 $&\underline{$ 0.600 $}&\underline{$ 0.436 $}&$ 0.964 $&$ 0.525 $&$ 0.701 $&$ 0.489 $&$ 0.756 $&$ 0.491 $&$ 0.633 $&$ 0.465 $&$ 0.662$&$ 0.473$\\

\bottomrule

\end{tabular}
}
\end{small}
\end{center}
\vspace{-5pt}

\end{table*}

\subsection{Imputation}

\subsubsection{Setup.}
%
%Imputation plays a critical role in addressing missing data issues commonly encountered in time series analyses, often resulting from system malfunctions. 
To assess our method's imputation capabilities, we employ three datasets: ETT \citep{haoyietal-informer-2021}, Electricity \citep{ecldata}, and Weather \citep{weatherdata}, serving as our benchmarks. To simulate various degrees of missing data, we randomly obscure time points at proportions of $\{12.5\%, 25\%, 37.5\%, 50\%\}$ following~\cite{wu2023timesnet}.

\subsubsection{Results.}
Table~\ref{tab:imputation} illustrates that our method achieves performance comparable to GPT4TS and surpasses other baselines, highlighting its effectiveness. We attribute the robust performance of GPT4TS primarily to its backbone feature extractor: the pre-trained language model, which excels at capturing time series patterns, enhancing its imputation proficiency.

\begin{table*}[htb]
\captionsetup{font=small} 
\caption{Imputation task: Randomly masked \{$12.5$\%, $25$\%, $37.5$\%, $50$\%\} of points in $96$-length series, averaging results over $4$ mask ratios.
Full results are in Appendix \ref{appendix:imputation_full}.}
\label{tab:imputation}
\vspace{-10pt}
%\vskip 0.15in
\begin{center}
\begin{small}
\scalebox{0.66}{
\setlength\tabcolsep{3pt}
\begin{tabular}{c|cc|cc|cc|cc|cc|cc|cc|cc|cc|cc}
\toprule

\multirow{2}{*}{Methods} 
&\multicolumn{2}{c|}{LLM-TS} &\multicolumn{2}{c|}{TimesNet} & \multicolumn{2}{c|}{GPT4TS}&\multicolumn{2}{c|}
{PatchTST}&\multicolumn{2}{c|}{LightTS}&\multicolumn{2}{c|}{DLinear}&\multicolumn{2}{c|}{FEDformer}&\multicolumn{2}{c|}{Stationary}&\multicolumn{2}{c|}{Autoformer}&\multicolumn{2}{c}{Reformer} \\
&MSE&MAE&MSE&MAE&MSE&MAE&MSE&MAE&MSE&MAE&MSE&MAE&MSE&MAE&MSE&MAE&MSE&MAE&MSE&MAE \\

\midrule

ETTm1&\boldsymbol{${0.025} $}&\boldsymbol{${0.103}$}&\underline{${0.028} $}&${0.109}$&\underline{${0.028}$}&\underline{${0.108}$}&$0.047 $&$0.140 $&$ 0.104$&$ 0.218$&$ 0.093$&$ 0.206$&$ 0.062$&$ 0.177$&$ 0.036$&$ 0.126$&$0.051$&$ 0.150$&$ 0.055 $&$ 0.166$ \\
ETTm2&\boldsymbol{${0.021} $}&\boldsymbol{${0.087}$}&\underline{${0.022} $}&${0.089}$&${0.023}$&\underline{${0.088}$}&$0.029 $&$0.102 $&$ 0.046$&$ 0.151$&$ 0.096$&$ 0.208$&$ 0.101$&$ 0.215$&$ 0.026$&$ 0.099 $&$0.029$&$ 0.105$&$ 0.157 $&$ 0.280$\\
ETTh1&\underline{${0.087} $}&\underline{${0.198}$}&${0.090} $&${0.199}$&\boldsymbol{${0.069}$}&\boldsymbol{${0.174}$}&$0.115 $&$0.224 $&$ 0.284$&$ 0.373$&$ 0.201$&$ 0.306$&$ 0.117$&$ 0.246$&$ 0.094$&$ 0.201 $&$0.103$&$ 0.214$&$0.122$&$ 0.245$\\
ETTh2&\boldsymbol{${0.050} $}&\underline{${0.148}$}&${0.051} $&${0.150}$&\boldsymbol{${0.050}$}&\boldsymbol{${0.144}$}&$0.065 $&$0.163 $&$ 0.119$&$ 0.250$&$ 0.142$&$ 0.259$&$ 0.163$&$ 0.279$&$ 0.053$&$ 0.152$&$ 0.055$&$ 0.156$&$0.234$&$ 0.352$\\
ECL&${0.094} $&${0.211}$&${0.095} $&${0.212}$&\underline{${0.091}$}&\underline{${0.207} $}&\boldsymbol{$0.072 $}&\boldsymbol{${0.183}$}&$ 0.131$&$ 0.262$&$ 0.132$&$ 0.260$&$ 0.130$&$ 0.259$&$ 0.100$&$ 0.218 $&$0.101$&$ 0.225$&$0.200$&$ 0.313$ \\
Weather&\boldsymbol{${0.030} $}&\underline{${0.056}$}&\underline{${0.031} $}&${0.059}$&${0.032}$&${0.058}$&$0.034 $&\boldsymbol{$0.055 $}&$ 0.055$&$ 0.117$&$ 0.052$&$ 0.110$&$ 0.099$&$ 0.203$&$ 0.032$&$ 0.059 $&\underline{$0.031$}&$ 0.057$&$0.038$&$ 0.087$ \\

\midrule

Average &\underline{${0.051} $}&\underline{${0.134}$}&${0.053} $&${0.136}$&\boldsymbol{${0.049}$}&\boldsymbol{${0.130} $}&$ 0.060 $&$ 0.144  $&$ 0.123 $&$ 0.228 $&$ 0.119 $&$ 0.224 $&$ 0.112 $&$ 0.229 $&$ 0.056 $&$ 0.142 $&$ 0.061 $&$ 0.151 $&$ 0.134 $&$ 0.240$ \\

\bottomrule

\end{tabular}
}
\end{small}
\end{center}
\end{table*}

\subsection{Classification}

\subsubsection{Setup.}
%
%Time series classification has significant applications in fields such as recognition technologies and medical diagnostics \citep{Moody2011PhysioNetPS}.
%
We focus on the application of our method to sequence-level time series classification tasks, a crucial test of its ability to learn high-level representations from data. 
Specifically, we employ $10$ diverse multivariate datasets sourced from the UEA Time Series Classification repository~\cite{bagnall2018uea}.
%
% \begin{wrapfigure}[18]{r}{0.55\textwidth}
%     \centering
%     \includegraphics[scale=0.5]{Figures/classification_overall.pdf}
%     \vspace{-15pt}
%     \caption{Model comparison in classification.}
%     \label{fig: classification_perf}
% \end{wrapfigure}
%
These datasets encompass a wide range of real-world applications, including gesture and action recognition, audio processing, medical diagnosis, among other practical domains.
We reproduce the results of TEST based on their code~\cite{sun2023test}.

\subsubsection{Results.}
\begin{figure}[htb]
\centering
\begin{minipage}[t]{.51\textwidth}
  \centering
    \includegraphics[scale=0.43]{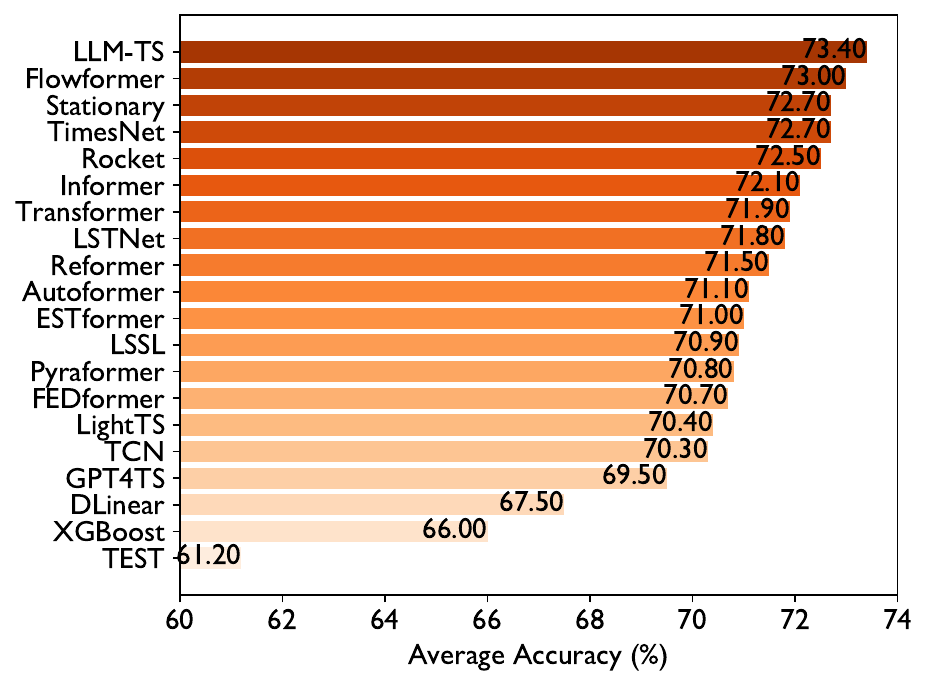}
    \caption{Model comparison in classification.}
    \label{fig: classification_perf}
\end{minipage}
\begin{minipage}[t]{.47\textwidth}
  \centering
    \includegraphics[scale=0.42]{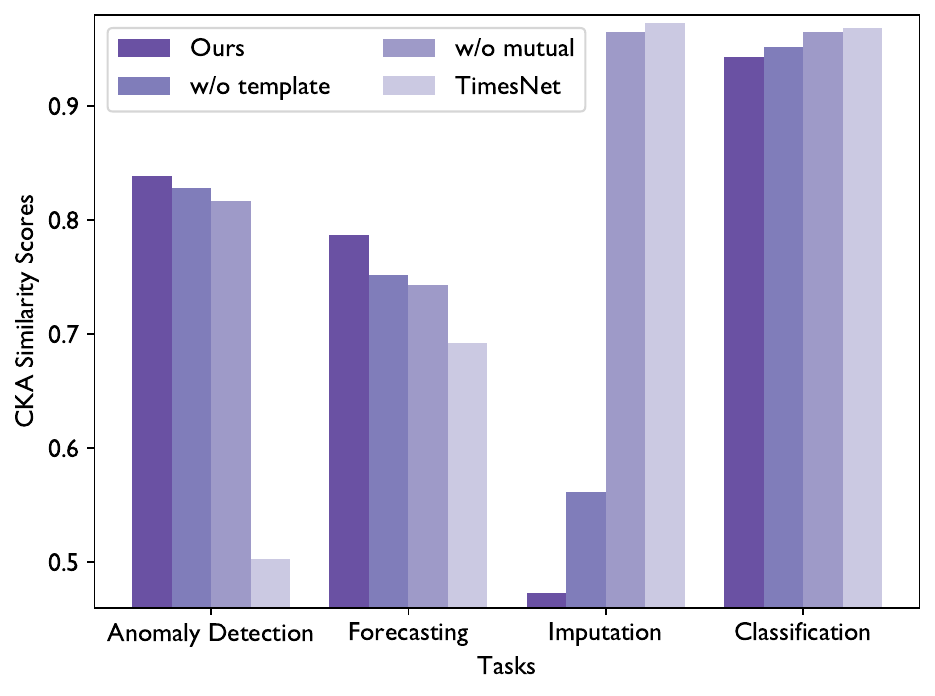}
    \caption{CKA by Task.}
    \label{fig: cka_simi}
\end{minipage}%
\end{figure}

As depicted in Figure~\ref{fig: classification_perf}, our \textit{LLM-TS Integrator} achieves superior performance with an average accuracy of $73.4\%$. As detailed in Appendix~\ref{appendix: classification}, it consistently outperforms other LLM-based methods across most tasks, including GPT4TS and TEST. We attribute this enhanced capability to the traditional TS modeling techniques in our framework, which effectively capture classification characteristics more adeptly than LLMs.

\subsection{Anomaly Detection}
\subsubsection{Setup.}
%
%The identification of anomalies within monitoring data plays a crucial role in ensuring industrial maintenance and reliability. 
Our study concentrates on unsupervised time series anomaly detection, aiming to identify aberrant time points indicative of potential issues. We benchmark our method against five established anomaly detection datasets: SMD \citep{Su2019RobustAD}, MSL \citep{Hundman2018DetectingSA}, SMAP \citep{Hundman2018DetectingSA}, SWaT \citep{DBLP:conf/cpsweek/MathurT16}, and PSM \citep{DBLP:conf/kdd/AbdulaalLL21}. These datasets span a variety of applications, including service monitoring, space and earth exploration, and water treatment processes. For a consistent evaluation framework across all experiments, we employ the classical reconstruction error metric to determine anomalies following~\cite{wu2023timesnet}.
%

% \hossein{If time allows, run our model n times with different seeds and report average+standard deviation for all tasks}

\subsubsection{Results.}
As indicated in Table~\ref{tab:anomaly}, our \textit{LLM-TS Integrator} exhibits superior performance with an average F1-score of $85.17\%$. This result underscores the versatility of \textit{LLM-TS}, demonstrating its capability not only in classifying complete sequences, as discussed previously, but also in effectively detecting anomalies in time series data.

\begin{table*}[htb]
\captionsetup{font=small} 
\caption{Anomaly detection task. F1-score (as \%). $\ast$. in the Transformers represents the name of $\ast$former. Full results are in Appendix \ref{appendix:anomaly_full}.}
\label{tab:anomaly}
\vspace{-10pt}
\begin{center}
\begin{small}
\scalebox{0.72}{
\setlength\tabcolsep{3pt}
\begin{threeparttable}[b]
\begin{tabular}{c|ccccccccccccccc}
\toprule

\multirow{2}{*}{Methods} & \multirow{2}{*}{LLM-TS}
& \multirow{2}{*}{TimesNet} & \multirow{2}{*}{GPT4TS} & \multirow{2}{*}{PatchTS.} & \multirow{2}{*}{ETS.} & \multirow{2}{*}{FED.} & \multirow{2}{*}{LightTS} & \multirow{2}{*}{DLinear} & \multirow{2}{*}{Stationary} & \multirow{2}{*}{Auto.} & \multirow{2}{*}{Pyra.} & \multirow{2}{*}{Anomaly.\tnote{**}} & \multirow{2}{*}{In.} & \multirow{2}{*}{Re.}  & \multirow{2}{*}{Trans.} \\
&&&&&&&&&&&&&&& \\

\midrule
SMD &$84.69 $&$84.57 $&$ 84.32 $&$84.62$&$83.13$&$ 85.08$&$ 82.53$&$ 77.10 $&$84.72 $&$85.11 $&$83.04$&$ 85.49 $&$81.65 $&$75.32$&$ 79.56$ \\
MSL &$81.11$&$80.34$&$ 81.73 $&$78.70$&$85.03$&$78.57$&$78.95$&$84.88$&$77.50$&$79.05$&$84.86$&$83.31$&$84.06$&$84.40$&$78.68 $\\
SMAP&$69.41 $&$69.18$&$ 68.86 $&$68.82$&$ 69.50$&$ 70.76$&$ 69.21$&$ 69.26$&$ 71.09$&$ 71.12$&$ 71.09$&$ 71.18$&$ 69.92$&$ 70.40$&$ 69.70 $\\
SWaT&$93.23$&$ 93.12$&$ 92.59 $&$85.72 $&$ 84.91$&$ 93.19 $&$93.33$&$ 87.52$&$ 79.88$&$ 92.74$&$ 91.78$&$ 83.10$&$ 81.43$&$ 82.80$&$80.37 $\\
PSM&$97.43$&$  97.27$&$ 97.34 $&$96.08$&$ 91.76$&$ 97.23$&$ 97.15$&$ 93.55$&$ 97.29$&$ 93.29$&$ 82.08$&$ 79.40$&$ 77.10$&$ 73.61 $&$76.07 $\\
\midrule
Average &\boldsymbol{$85.17$}&{${84.90}$}&\underline{$ 84.97 $}&$82.79$&$ 82.87$&$ 84.97$&$ 84.23$&$ 82.46$&$ 82.08$&$ 84.26$&$ 82.57 $&$80.50$&$ 78.83$&$ 77.31$&$ 76.88$\\

\bottomrule
\end{tabular}

\end{threeparttable}

}
\end{small}
\end{center}
\vspace{-10pt}
\end{table*}

\subsection{Ablations}
\label{subsec: model_analysis}
\begin{table}
\caption{Results averaged over $4$ prediction lengths.}
\label{tab: ablation_part}
\vspace{-10pt}
\begin{center}
\begin{small}
\scalebox{0.70}{
\setlength{\tabcolsep}{3pt} % 控制列间距
\begin{tabular}{c|cc|cc|cc|cc|cc}
\toprule
\multicolumn{1}{c|}{Methods} & \multicolumn{2}{c|}{Ours} & \multicolumn{2}{c|}{w/o mutual} & \multicolumn{2}{c|}{w/o template} & \multicolumn{2}{c|}{w/o reweight} & \multicolumn{2}{c}{TimesNet} \\
\midrule
Metric & MSE  & MAE & MSE & MAE  & MSE & MAE & MSE & MAE & MSE  & MAE \\
\midrule
Weather &\boldsymbol{$ 0.257 $}&\boldsymbol{$ 0.285 $}&$ 0.264 $&$ 0.290 $&$ 0.263 $&$ 0.288 $&$ 0.264 $&$ 0.291 $&$ 0.265 $&$ 0.290$ \\
ETTh1   &\boldsymbol{$ 0.454 $}&\boldsymbol{$ 0.451 $}&$ 0.467 $&$ 0.460 $&$ 0.465 $&$ 0.460 $&$ 0.464 $&$ 0.463 $&$ 0.470 $&$ 0.462$ \\
ETTm1   &\boldsymbol{$ 0.401 $}&\boldsymbol{$ 0.409 $}&$ 0.411 $&$ 0.417 $&$ 0.406 $&$ 0.415 $&$ 0.403 $&$ 0.411 $&$ 0.414 $&$ 0.418$ \\
ILI     &\boldsymbol{$ 1.973 $}&\boldsymbol{$ 0.894 $}&$ 2.221 $&$ 0.942 $&$ 2.173 $&$ 0.950 $&$ 2.173 $&$ 0.947 $&$ 2.266 $&$ 0.974$ \\
\bottomrule
\end{tabular}
}
\end{small}
\end{center}
\vspace{-10pt}
%\end{table*}
\end{table}
In this section, we first verify the effectiveness of our framework by sequentially removing key components: (1) \textit{mutual information} module and (2) \textit{sample reweighting} module. 
Additionally, for \textit{mutual information}, we explore the impact of removing the template while retaining the raw time series data inputs to the LLM.
We denote these variants as \textit{w/o mutual}, \textit{w/o reweight} and \textit{w/o template}. Our experiments span long-term forecasting tasks including Weather, ETTh1, ETTm1 and ILI. As detailed in Table~\ref{tab: ablation_part}, the removal of any component leads to a decrease in performance, confirming the value of each design element.
Additionally, we explore the use of the MINE estimator~\cite{hjelm2018learning} instead of the Jensen-Shannon MI estimator in our main paper, with further details provided in Appendix~\ref{appendix: further_ablation}.
Lastly, we showcase various case studies to demonstrate the enhancements facilitated by our method in Appendix~\ref{appendix: show} and explore template variations in Appendix~\ref{appendix: template}.

%\begin{table*}[htb]

\subsubsection{Mutual Information.}
We further explore the \textit{mutual information} module from a representation learning perspective, following the findings in~\cite{wu2023timesnet}.
They adopt a CKA~(Centered Kernel Alignment) metric which measures similarity between representations obtained from the first and last layer of a model and they find that forecasting and anomaly detection benefits from high CKA similarity, contrasting with that imputation and classification tasks benefits from lower CKA similarity.

Experiments are conducted using the MSL dataset for the anomaly detection task, the Weather dataset for forecasting, the ETTh1 dataset for imputation, and the PEMS-SF dataset for classification. 
As depicted in Figure~\ref{fig: cka_simi}, the removal of components in our method results in decreased CKA similarity in anomaly detection and forecasting tasks, but an increase in imputation and classification tasks. 
This observation further substantiates the effectiveness of our components.

\subsubsection{Sample Reweighting.}
Regarding the \textit{sample reweighting} module, we illustrate the behavior of the learned weighting network in Appendix~\ref{appendix: further_ablation}. The trend confirms our hypothesis: sample weight $\omega_O$ increases with the prediction loss $l_O$, and weight $\omega_I$ decreases as $l_O$ increases. This pattern validates our \textit{sample reweighting} module. Further discussion comparing this module to a fixed weight scheme are presented in Appendix~\ref{appendix: further_ablation}.

% TimesNet CKA similarity (kernel CKA):  
% \begin{itemize}
%     \item Anomaly detection (MSL): LLM-TS(0.839) >  LLM-TS wo/ text (0.828) >LLM-TS wo/ mutual (0.817) > TimesNet(0.503) 
%     \item Forecasting (Weather): LLM-TS(0.787) > LLM-TS wo/ text (0.752) > LLM-TS wo/ mutual (0.743) > TimesNet(0.692) 
%     \item Imputation (ETTh1): LLM-TS(0.473) < LLM-TS wo/ text (0.562) < LLM-TS wo/ mutual (0.965) < TimesNet(0.973)
    
%     \item Classification (PEMS-SF): LLM-TS(0.943) <LLM-TS wo/ text (0.952)< LLM-TS wo/ mutual (0.965) < TimesNet(0.969)
% \end{itemize}

To verify the effectiveness of our method, we conduct ablation studies focusing on (1) traditional time series (TS) models and (2) language models.

\subsubsection{Traditional Models.}
Although we utilize TimesNet as our primary model, our framework is applicable to other traditional models. We explored additional traditional models including ETSformer~\cite{woo2022etsformer}, Stationary~\cite{liu2022non}, and FreTS~\cite{yi2024frequency}.
As shown in Table~\ref{tab: trad_ablation_main}, integrating \textit{LLM-TS} generally enhances performance across all traditional models, underscoring the benefits of our method.

\subsubsection{Language Models.}
In the main paper, the LLaMA-3b model~\cite{touvron2023llama} is used to generate embeddings for the TS language description. We compare it with GPT2~\cite{radford2019language} and BERT~\cite{devlin2018bert} to assess different embeddings' performance. Table~\ref{tab: lm_ablation_main} reveals that LLaMA-3b generally outperforms the alternatives, and all LMs improve results compared to non-LLM approaches, validating the effectiveness of \textit{LLM-TS Integrator}.

\begin{table}[htb]

\caption{Ablation results on different traditional models. Full results are in Appendix~\ref{appendix: further_ablation}.}
\label{tab: trad_ablation_main}
\vspace{-10pt}
\begin{center}
\begin{small}
\scalebox{0.63}{
\setlength\tabcolsep{3pt}
\begin{tabular}{c|cc|cc|cc|cc|cc|cc}
\toprule

\multicolumn{1}{c|}{Methods}&\multicolumn{2}{c|}{ETSformer}&\multicolumn{2}{c|}{ETS LLM-TS}&\multicolumn{2}{c|}{Stationary}&\multicolumn{2}{c|}{Stat LLM-TS}&\multicolumn{2}{c|}{FreTS} &\multicolumn{2}{c}{FreTS LLM-TS} \\

\midrule

\multicolumn{1}{c|}{Metric}  & MSE  & MAE& MSE  & MAE& MSE & MAE& MSE & MAE& MSE  & MAE& MSE  & MAE  \\

\midrule

\multirow{1}{*}{\rotatebox{0}{$Weather$}}
 & 0.313 & 0.382 & 0.307 & 0.375& 0.282 & 0.307 & 0.284 & 0.309 & 0.262 & 0.306 & 0.255 & 0.302 \\
%\midrule

\multirow{1}{*}{\rotatebox{0}{$ETTh1$}}
 & 0.799 & 0.684 & 0.791 & 0.678& 0.667 & 0.582 & 0.653 & 0.572& 0.484 & 0.473& 0.478 & 0.466 \\
%\midrule

\multirow{1}{*}{\rotatebox{0}{$ETTm1$}}
 & 0.638 & 0.583 & 0.555 & 0.528 & 0.527 & 0.477 & 0.522 & 0.471 & 0.415 & 0.422 & 0.407 & 0.415\\
%\midrule

\multirow{1}{*}{\rotatebox{0}{$ILI$}}
 & 3.922 & 1.367 & 3.740 & 1.320 & 2.722 & 1.041 & 2.205 & 0.935 & 3.449 & 1.279 & 3.158 & 1.211\\
\bottomrule
\end{tabular}
}
\end{small}
\end{center}
\vspace{-15pt}
\end{table}

\vspace{-10pt}
\begin{table}[htb]
%\caption{Different LLM embeddings. We use prediction length $O \in \{24, 36, 48, 60\}$ for ILI and $O \in \{96, 192, 336, 720\}$ for others. }
\caption{Ablation results on different LLM embeddings. Full results are in Appendix~\ref{appendix: further_ablation}.}
\label{tab: lm_ablation_main}
\vspace{-10pt}
\begin{center}
\begin{small}
\scalebox{0.63}{
\setlength\tabcolsep{3pt}
\begin{tabular}{c|cc|cc|cc|cc|cc}
\toprule

\multicolumn{1}{c|}{Methods}&\multicolumn{2}{c|}{LLM-TS (LLaMA)}&\multicolumn{2}{c|}{LLaMA w/o template}&\multicolumn{2}{c|}{GPT2}&\multicolumn{2}{c|}{BERT}&\multicolumn{2}{c|}{No LLM} \\

\midrule

\multicolumn{1}{c|}{Metric} & MSE  & MAE& MSE  & MAE & MSE & MAE& MSE & MAE& MSE  & MAE  \\
\midrule

\multirow{1}{*}{\rotatebox{0}{$Weather$}}
 &\boldsymbol{$ 0.257 $}&\boldsymbol{$ 0.285 $}&$ 0.263 $&$ 0.288$&$ 0.261 $&$ 0.287 $&$ 0.260 $&$ 0.287 $&$ 0.264 $&$ 0.290$ \\
%\midrule

\multirow{1}{*}{\rotatebox{0}{$ETTh1$}}
 &\boldsymbol{$ 0.454 $}&\boldsymbol{$ 0.451 $}&$ 0.465 $&$ 0.460$&$ 0.464 $&$ 0.458 $&$ 0.467 $&$ 0.460 $&$ 0.467 $&$ 0.460$ \\
%\midrule

\multirow{1}{*}{\rotatebox{0}{$ETTm1$}}
 &\boldsymbol{$ 0.401 $}&\boldsymbol{$ 0.409 $}&$ 0.406 $&$ 0.415$&$ 0.406 $&$ 0.413 $&$ 0.406 $&$ 0.412 $&$ 0.411 $&$ 0.417$ \\
%\midrule

\multirow{1}{*}{\rotatebox{0}{$ILI$}}
 &\boldsymbol{$ 1.973 $}&\boldsymbol{$ 0.894 $}&$ 2.173 $&$ 0.950$&$ 2.169 $&$ 0.936 $&$ 2.193 $&$ 0.952 $&$ 2.221 $&$ 0.942$ \\

\bottomrule
\end{tabular}
}
\end{small}
\end{center}
% \vspace{-15pt}
\end{table}

\vspace{10pt}
\section{Related Work}
%
% In the second paragraph of the Introduction, we explore the intersection of Large Language Models (LLMs) with time series analysis. 
% %
% Our discussion begins with methods that transform time series data into textual formats, then moves to explore specialized pre-training models developed for time series.
%\hossein{I think we can remove this intro for related work}

\subsubsection{LLM for TS Modeling.}
FPT~\cite{zhou2024one} suggests utilizing pre-trained language models to extract features from time series for improved predictions.
TIME-LLM~\cite{jin2024timellm} and TEST~\cite{sun2023test} adapt LLMs for general time series forecasting by maintaining the original language model structure while reprogramming the input to fit time series data~\cite{zhou2024large}.
LLMTIME~\cite{gruver2024large} interprets time series as sequences of numbers, treating forecasting as a next-token prediction task akin to text processing, applying pre-trained LLMs for this purpose.
Given that it is not a state-of-the-art method and primarily targets zero-shot forecasting, it has not been incorporated into our experimental framework.
TEMPO~\cite{cao2023tempo} utilizes essential inductive biases of the TS task for generative pre-trained transformer models.

\subsubsection{Time Series to Text.}
PromptCast~\cite{xue2023promptcast} proposes to transform the numerical input and output into prompts, which enables forecasting in a sentence-to-sentence manner.
%
%Another study~\cite{xue2022translating} develops a mobility-to-language template that articulates numerical mobility data as descriptive natural language sentences. 
%
Time-LLM~\cite{jin2024timellm} incorporates background, instruction and statistical information of the time series data via natural language to facilitate time series forecasting in LLM.
LLMTIME~\cite{gruver2023large} converts time series data into a string of numbers and predicts future values as if completing a text.
%

%

% WE MAY NOT NEED THIS PART \gabriel{I agree that next paragraph is not actually related work. On the contrary the new references in the introduction can be added.}

% \subsubsection{Time Series Pre-training.}
% %
% Contrary to applying pre-trained LLMs directly to time series, some recent research focuses on specifically pre-training models on time series itself~\cite{ma2023survey}.
% %
% Zerveas et al.~\cite{zerveas2021transformer} pioneer a Transformer-based architecture for unsupervised learning of multivariate time series, employing an input denoising objective for enhanced downstream task performance. 
% %
% TS2Vec~\cite{yue2022ts2vec} advocates for contrastive learning across hierarchical context views to foster robust contextual representations at each timestamp. 
% %
% CoST~\cite{woo2022cost} innovates with a dual-contrastive loss approach, targeting both time and frequency domains to capture trend and seasonal patterns in long-term time series data. 
% %
% SimMTM~\cite{dong2024simmtm} suggests reconstructing masked time points through the weighted amalgamation of neighboring data points.
% %
% Ti-MAE~\cite{li2023ti} employs a strategy of masking embedded time series data to train an autoencoder that accurately reconstructs these points at the point level.

\subsubsection{Mutual Information}
The Infomax principle~\cite{Linsker1988SelforganizationIA,bell1995information}, applied in the context of neural networks, advocates for maximizing mutual information between the inputs and outputs of a network. Traditionally, quantifying mutual information was challenging outside a few specific probability distributions, as discussed in~\cite{shwartzziv2017opening}. This complexity led to the development of various heuristics and approximations~\cite{tishby2000information}. More recently, a breakthrough came with MINE~\cite{belghazi2021mine}, which introduced a neural estimator capable of assessing mutual information between two arbitrary quantities with a precision that depends on the capacity of the encoding network. This innovative approach has spearheaded advancements in the field of representation learning~\cite{hjelm2019learning,sylvain2019locality}. The estimator we utilize is based on the Jensen-Shannon divergence variant of the MINE mutual information estimator.

\subsubsection{Sample Reweighting.}
Sample reweighting is commonly used to improve training efficacy~\cite{fang2024generalizing, wang2024unified, yuan2024task, zhang2024uncertainty, yuan2024importance}.
Traditional approaches~\cite{freund1997decision, sun2007cost} assign larger weights to samples with higher loss values, as these hard samples have greater learning potential.
Recent studies~\cite{ren2018learning} suggest using a validation set to guide the learning of sample weights, which can  enhance model training.
Notably, meta-weight-net~\cite{shu2019meta} proposes learning a mapping from sample loss to sample weight.
In this work, we adopt an MLP network that takes sample prediction loss as input and outputs dual weights for prediction loss and mutual information loss.
%
%This modeling approach is motivated by the intrinsic relationship between loss and weight, and is also automatically guided by a validation set using a bi-level optimization framework.

\section{Conclusion and Discussion}
\label{sec: conclusion}

In conclusion, the \textit{LLM-TS Integrator framework} offers a promising approach to integrating Large Language Models (LLMs) with traditional TS methods.
This work extends our prior findings from the workshop~\cite{chen2024enhance}.
By encouraging high mutual information between textual and TS data, our method aims to maintain the distinct characteristics of TS while benefiting from the advanced pattern recognition capabilities of LLMs. The introduced sample reweighting module enhances performance by dynamically adjusting the relevance of each sample based on its predictive and informational contributions. Comprehensive empirical evaluations suggest that this framework improves accuracy across various TS tasks, including forecasting, anomaly detection, and classification. However, further research is necessary to confirm these results across more diverse datasets and to explore additional LLM features that could further enrich our model. Additionally, it is important to acknowledge current limitations, such as the need for computational resources and the potential challenges in aligning the two modalities effectively.

% \subsubsection{Limitations.} 
% %
% However, further research is necessary to confirm these results across more diverse datasets and to explore additional LLM features that could further enrich our model. Additionally, it is important to acknowledge current limitations, such as the need for computational resources and the potential challenges in aligning the two modalities effectively.
%

\bibliography{main.bib}

%\newpage

\appendix

\section{Reproducibility Checklist}
This paper:
\begin{itemize}

\item Includes a conceptual outline and/or pseudocode description of AI methods introduced (yes/partial/no/NA) YES
\item Clearly delineates statements that are opinions, hypothesis, and speculation from objective facts and results (yes/no) YES
\item Provides well marked pedagogical references for less-familiare readers to gain background necessary to replicate the paper (yes/no) YES
\end{itemize}

\noindent Does this paper make theoretical contributions? (yes/no) NO

\noindent Does this paper rely on one or more datasets? (yes/no) YES 

\noindent If yes, please complete the list below.

\begin{itemize}

\item A motivation is given for why the experiments are conducted on the selected datasets (yes/partial/no/NA) YES

\item All novel datasets introduced in this paper are included in a data appendix. (yes/partial/no/NA) NA

\item All novel datasets introduced in this paper will be made publicly available upon publication of the paper with a license that allows free usage for research purposes. (yes/partial/no/NA) NA

\item All datasets drawn from the existing literature (potentially including authors’ own previously published work) are accompanied by appropriate citations. (yes/no/NA) YES

\item All datasets drawn from the existing literature (potentially including authors’ own previously published work) are publicly available. (yes/partial/no/NA) YES

\item All datasets that are not publicly available are described in detail, with explanation why publicly available alternatives are not scientifically satisficing. (yes/partial/no/NA) NA

\end{itemize}

\noindent Does this paper include computational experiments? (yes/no) YES

\noindent If yes, please complete the list below.

\begin{itemize}

\item Any code required for pre-processing data is included in the appendix. (yes/partial/no). YES. The code is available at \url{https://anonymous.4open.science/r/llm_ts_anonymous-F07D/README.MD}

\item All source code required for conducting and analyzing the experiments is included in a code appendix. (yes/partial/no) YES

\item All source code required for conducting and analyzing the experiments will be made publicly available upon publication of the paper with a license that allows free usage for research purposes. (yes/partial/no) YES 

\item All source code implementing new methods have comments detailing the implementation, with references to the paper where each step comes from (yes/partial/no) YES

\item If an algorithm depends on randomness, then the method used for setting seeds is described in a way sufficient to allow replication of results. (yes/partial/no/NA) YES

\item This paper specifies the computing infrastructure used for running experiments (hardware and software), including GPU/CPU models; amount of memory; operating system; names and versions of relevant software libraries and frameworks. (yes/partial/no) YES

\item This paper formally describes evaluation metrics used and explains the motivation for choosing these metrics. (yes/partial/no) YES

\item This paper states the number of algorithm runs used to compute each reported result. (yes/no) YES

\item Analysis of experiments goes beyond single-dimensional summaries of performance (e.g., average; median) to include measures of variation, confidence, or other distributional information. (yes/no) YES

\item The significance of any improvement or decrease in performance is judged using appropriate statistical tests (e.g., Wilcoxon signed-rank). (yes/partial/no) NA

\item This paper lists all final (hyper-)parameters used for each model/algorithm in the paper’s experiments. (yes/partial/no/NA) YES

\item This paper states the number and range of values tried per (hyper-) parameter during development of the paper, along with the criterion used for selecting the final parameter setting. (yes/partial/no/NA) YES

\end{itemize}

\newpage
\section{Appendix}
%\newpage
\subsection{Mutual Information Recalculation}
\label{appendix: mutual}

Recall that mutual information can be calculated using the equation:
%\hossein{x and t should be reversed in Eq 12 to 14}
%
\begin{equation}
    \begin{aligned}
    I(\boldsymbol{\theta}, \boldsymbol{\beta}) =  \mathbb{E}_{\mathbb{S}}[-{sp}(-{T}_{\boldsymbol{\beta}}(\boldsymbol{h}^{m}_{\boldsymbol{\theta}}(\boldsymbol{x}), \boldsymbol{h}^{l}(\boldsymbol{t}))] - \\
    \mathbb{E}_{\mathbb{S}\times\mathbb{\tilde{S}}}[{sp}({T}_{\boldsymbol{\beta}}(\boldsymbol{h}^{m}_{\boldsymbol{\theta}}(\boldsymbol{x}), \boldsymbol{h}^{l}(\boldsymbol{\tilde{t}}))]\,,
    %\label{eq: mututual}
    \end{aligned}
\end{equation}
where $T_{\boldsymbol{\beta}}$ signifies the discriminator characterized by parameters $\boldsymbol{\beta}$, and $\text{sp}$ denotes the softplus function.
Notably, $(\boldsymbol{x}, \boldsymbol{t})$ symbolizes a sample from the dataset $\mathbb{S}$, while $(\boldsymbol{\tilde{x}}, \boldsymbol{\tilde{t}})$ represents a different sample from the dataset $\mathbb{\tilde{S}} = \mathbb{S}$.

This formulation presumes a uniform distribution of samples. However, we have already computed probabilities $p_{I}^{i}$ for each sample, which introduces a non-uniform distribution.
For a batch of $N$ samples, the expected value is computed as
\begin{equation}
\begin{split}
    \mathbb{E}_{\mathbb{S}}[-{sp}(-{T}_{\boldsymbol{\beta}}(\boldsymbol{h}^{m}_{\boldsymbol{\theta}}(\boldsymbol{x}), \boldsymbol{h}^{l}(\boldsymbol{t}))] =
    \\ -\sum_{i=1}^{N} p_{I}^{i}{sp}(-{T}_{\boldsymbol{\beta}}(\boldsymbol{h}^{m}_{\boldsymbol{\theta}}(\boldsymbol{x^i}), \boldsymbol{h}^{l}(\boldsymbol{t^i})).
\end{split}
\end{equation}

\begin{equation}
\begin{split}
    \mathbb{E}_{\mathbb{S}\times\mathbb{\tilde{S}}}[{sp}({T}_{\boldsymbol{\beta}}(\boldsymbol{h}^{m}_{\boldsymbol{\theta}}(\boldsymbol{x}), \boldsymbol{h}^{l}(\boldsymbol{\tilde{t}}))] = 
    \\ \sum_{i}\sum_{i \neq j} \hat{p}^{ij}{sp}({T}_{\boldsymbol{\beta}}(\boldsymbol{h}^{m}_{\boldsymbol{\theta}}(\boldsymbol{x^i}), \boldsymbol{h}^{l}(\boldsymbol{\tilde{t}^j})).
\end{split}
\end{equation}
Here, $\hat{p}^{ij}$ is defined as $\frac{p_{I}^{i} \cdot p_{I}^{j}}{\sum_{i}\sum_{i \neq j}p_{I}^{i} \cdot p_{I}^{j}}$, adjusting for the non-uniform distribution of sample probabilities.
As $p_{I}^{i}$ is produced from the weighting network $\boldsymbol{\alpha}$, we can also write $I(\boldsymbol{\theta}, \boldsymbol{\beta})$ as $I(\boldsymbol{\theta}, \boldsymbol{\beta}, \boldsymbol{\alpha})$.

\subsection{Experimental Settings}
\label{appendix: settings}

following~\cite{shu2019meta}, the weighting network comprises a two-layer MLP with a hidden size of $100$, and we set the learning rate $\eta_2$ for this network at $0.001$.
The learning rate $\eta_0$ of the discriminator is set as $0.001$ at the first epoch and then decreases to $0.0001$ for the rest of epochs.

\subsection{Standard Deviation Results}
\label{appendix: ablation_std}

Table~\ref{tab: ablation_std} presents the results along with standard deviations to underscore the consistency and reliability of our method's performance.

\begin{table*}[htb]
\caption{Ablation results with standard deviation.}
\label{tab: ablation_std}
\vskip 0.15in
\begin{center}
\begin{small}
\scalebox{0.70}{
\setlength\tabcolsep{3pt}
\begin{tabular}{c|c|cc|cc|cc|cc}
\toprule

\multicolumn{2}{c|}{Methods}&\multicolumn{2}{c|}{Ours}&\multicolumn{2}{c|}{w/o mutual}&\multicolumn{2}{c|}{w/o reweight}&\multicolumn{2}{c|}{TimesNet}\\

\midrule

\multicolumn{2}{c|}{Metric} & MSE  & MAE & MSE & MAE& MSE & MAE& MSE  & MAE\\
\midrule

\multirow{4}{*}{\rotatebox{90}{$Weather$}}
& $96$ & $0.166 \pm 0.002$ & $0.217 \pm 0.002$ & $0.168\pm 0.003$ & $0.218\pm 0.005$ & $0.181 \pm 0.003$ & $0.232 \pm 0.001$ & $0.174\pm 0.003$ & $0.224\pm 0.002$ \\
& $192$ & $0.229 \pm 0.003$ & $0.269\pm 0.003$ & $0.227 \pm 0.004$ & $0.268\pm 0.003$ & $0.230 \pm 0.002$ & $0.270 \pm 0.004$ & $0.235\pm 0.001$ & $0.272\pm 0.003$ \\
& $336$ & $0.278\pm 0.002$ & $0.302\pm 0.003$ & $0.298\pm 0.004$ & $0.318\pm 0.003$ &  $0.283\pm 0.004$ & $0.306\pm 0.002$  & $0.285\pm 0.002$ & $0.307\pm 0.002$ \\
& $720$ & $0.354 \pm 0.001$ & $0.351\pm 0.001$ & $0.361 \pm 0.002$ & $0.356 \pm 0.001$ & $0.361\pm 0.002$ & $0.355\pm 0.001$ &  $0.365 \pm 0.001$ & $0.358 \pm 0.000$\\
\midrule

\multirow{4}{*}{\rotatebox{90}{$ETTh1$}}
& $96$  & $0.403\pm 0.005$ & $0.420\pm 0.003$ & $0.402\pm 0.004$ & $0.422\pm 0.002$ & $0.408\pm 0.003$ & $0.428\pm 0.002$ & $0.414\pm 0.006$ & $0.431\pm 0.004$ \\
& $192$ & $0.440\pm 0.009$ & $0.441\pm 0.004$ & $0.459\pm 0.006$ & $0.455\pm 0.005$ & $0.469\pm 0.005$ & $0.460\pm 0.003$ & $0.463\pm0.010$ & $0.456\pm 0.006$ \\
& $336$ & $0.471\pm 0.006$ & $0.457\pm 0.004$ & $0.471 \pm 0.005$ & $0.457\pm 0.004$ & $0.492\pm 0.004$ & $0.474\pm 0.004$ & $0.487\pm 0.007$ & $0.466\pm 0.005$  \\
& $720$ & $0.503 \pm 0.005$ & $0.487\pm 0.004$ & $0.535\pm 0.003$ & $0.507\pm 0.003$ & $0.485\pm 0.006$ & $0.478\pm 0.007$ & $0.517 \pm 0.004$ & $0.494\pm 0.004$ \\
\midrule

\multirow{4}{*}{\rotatebox{90}{$ETTm1$}}
& $96$  & $0.329\pm 0.014$ & $0.371\pm 0.006$ & $0.341\pm 0.010$ & $0.377\pm 0.008$ & $0.350\pm 0.011$ & $0.387\pm 0.005$ & $0.340\pm 0.011$ & $0.377\pm 0.007$ \\
& $192$ & $0.380\pm 0.009$ & $0.398\pm 0.004$ & $0.404\pm 0.010$ & $0.413\pm 0.005$ & $0.383\pm 0.010$ & $0.397\pm 0.005$ & $0.406\pm 0.012$ & $0.408 \pm 0.004$ \\
& $336$ & $0.418\pm 0.004$ & $0.425\pm 0.004$ & $0.432\pm 0.005$ & $0.428\pm 0.002$ & $0.410\pm 0.004$ & $0.411\pm 0.003$ & $0.424\pm 0.006$ & $0.425\pm 0.003$ \\
& $720$ & $0.476\pm 0.008$ & $0.440\pm 0.005$ & $0.468\pm 0.009$ & $0.449\pm 0.004$ & $0.467\pm 0.007$ & $0.448\pm 0.003$ & $0.485\pm 0.010$ & $0.461\pm 0.006$ \\
\midrule

\multirow{4}{*}{\rotatebox{90}{$ILI$}}
& $24$ & $1.921 \pm 0.201$ & $0.898\pm 0.033$ & $2.170\pm 0.174$ & $0.947\pm 0.039$ & $1.934\pm 0.170$ & $0.925\pm 0.034$ & $2.072\pm 0.211$ & $0.948\pm 0.026$\\
& $36$ & $2.151\pm 0.061$ & $0.933\pm 0.035$ & $2.093\pm 0.119$ & $0.889\pm 0.041$ & $2.505\pm 0.179$ & $1.020\pm 0.017$ & $2.494\pm 0.125$ & $1.019\pm 0.007$ \\
& $48$ & $2.062\pm 0.090$ & $0.892\pm 0.019$ & $2.418\pm 0.058$ & $0.959\pm 0.014$ & $2.325\pm 0.201$ & $0.948\pm 0.062$ & $2.298\pm 0.066$ & $0.964\pm 0.011$ \\
& $60$ & $1.759\pm 0.214$ & $0.853\pm 0.061$ & $2.203\pm 0.181$ & $0.971\pm 0.048$ & $1.926\pm 0.152$ & $0.896\pm 0.039$ & $2.198\pm0.070$ & $0.963\pm0.017$\\
\bottomrule
\end{tabular}
}
\end{small}
\end{center}
\vskip -0.1in
\end{table*}

\subsection{Showcases}
\label{appendix: show}
\begin{figure*}
\centering
\begin{minipage}[t]{.33\textwidth}
  \centering
    \includegraphics[width=1.00\columnwidth]{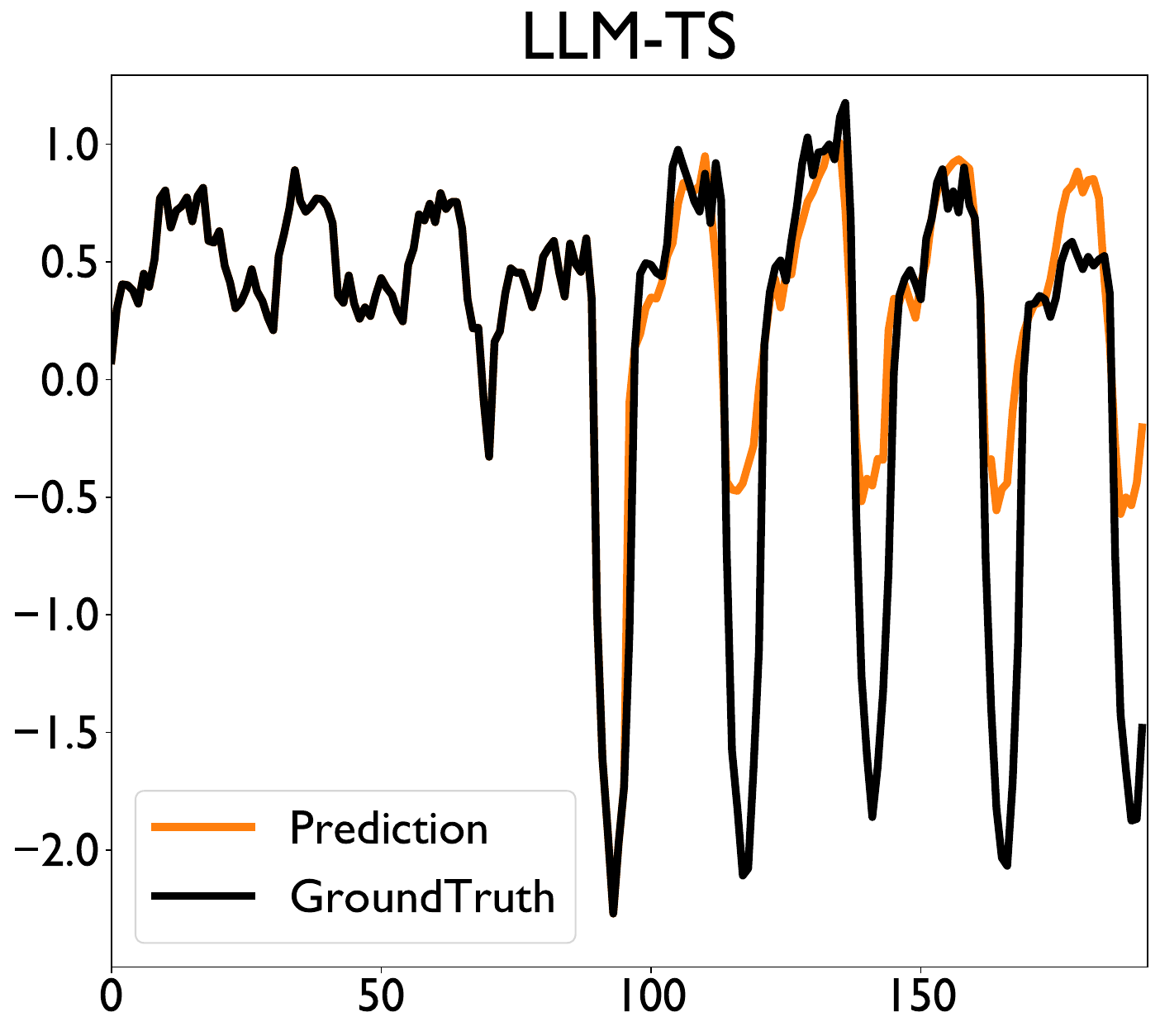}
    \captionsetup{width=1.05\linewidth}
    \label{fig: etth1_llmts}
\end{minipage}
\begin{minipage}[t]{.33\textwidth}
  \centering
    \includegraphics[width=1.0\columnwidth]{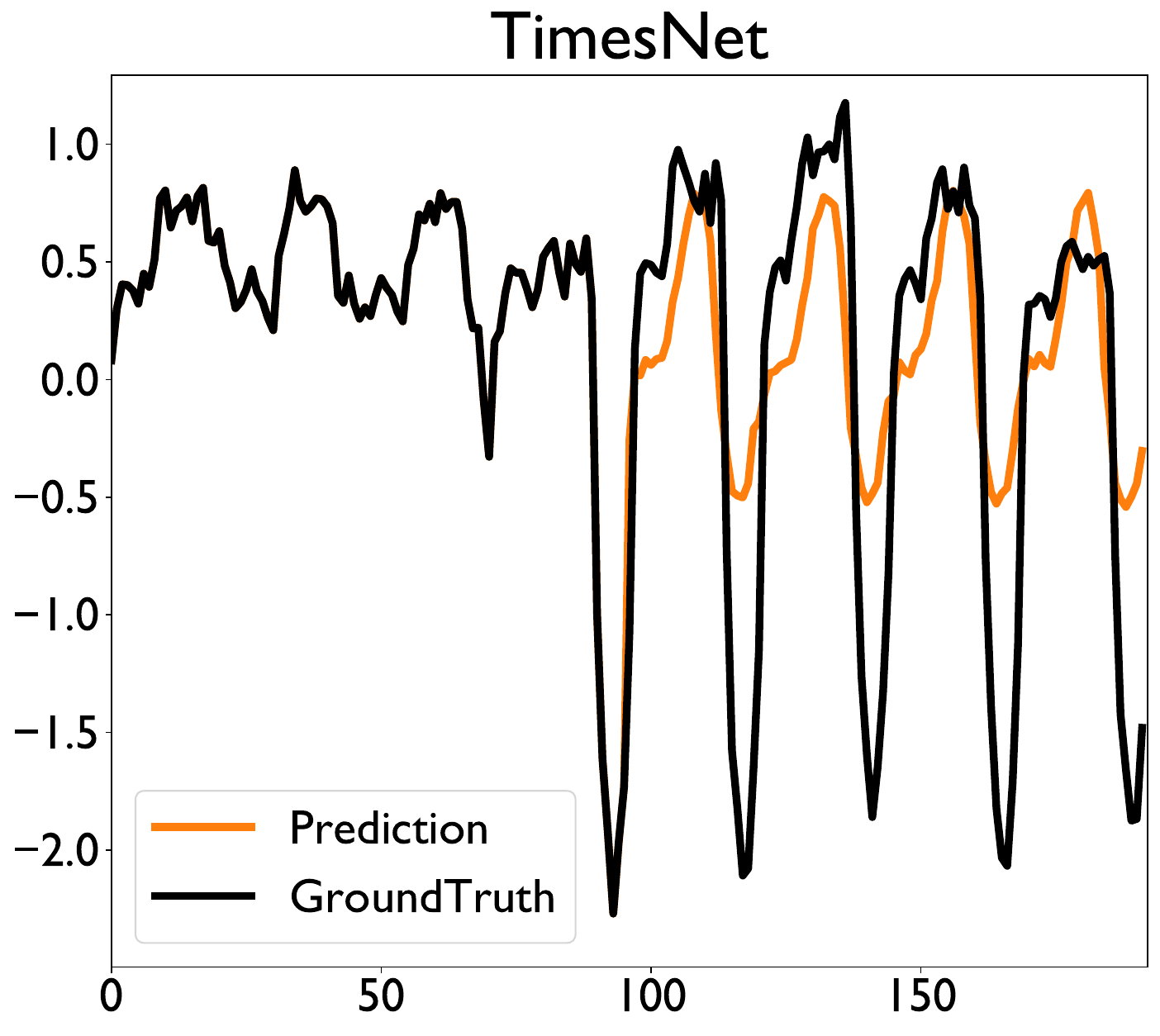}
    \captionsetup{width=.90\linewidth}
    %\caption{a}
    \label{fig: etth1_timesnet}
\end{minipage}%
\begin{minipage}[t]{.33\textwidth}
  \centering
    \includegraphics[width=1.00\columnwidth]{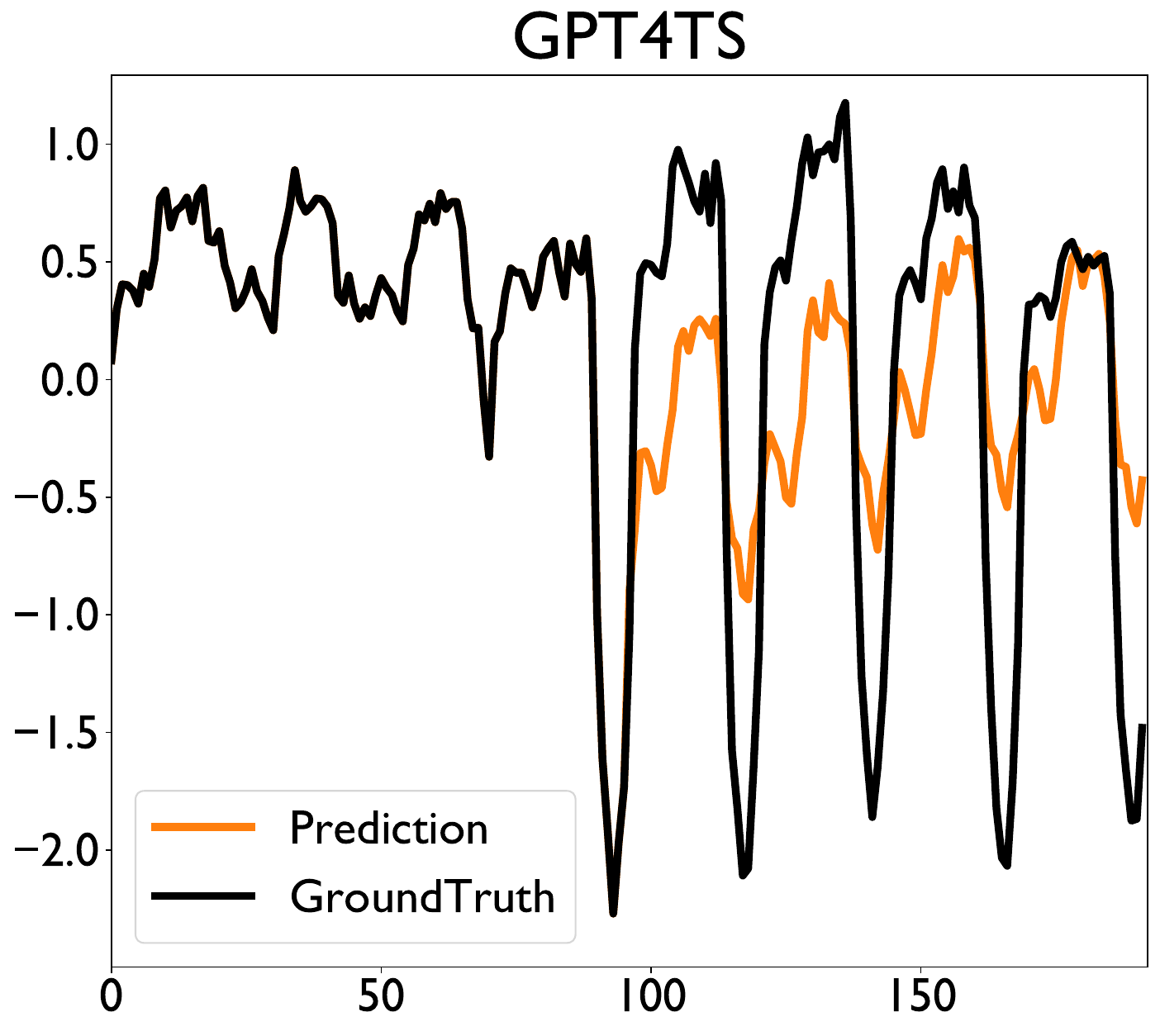}
    \captionsetup{width=.90\linewidth}
    \label{fig: etth1_gpt4ts}
\end{minipage}
\caption{ETTh1}
\label{fig: ETTh1_show}
\end{figure*}

\begin{figure*}
\centering
\begin{minipage}[t]{.33\textwidth}
  \centering
    \includegraphics[width=1.00\columnwidth]{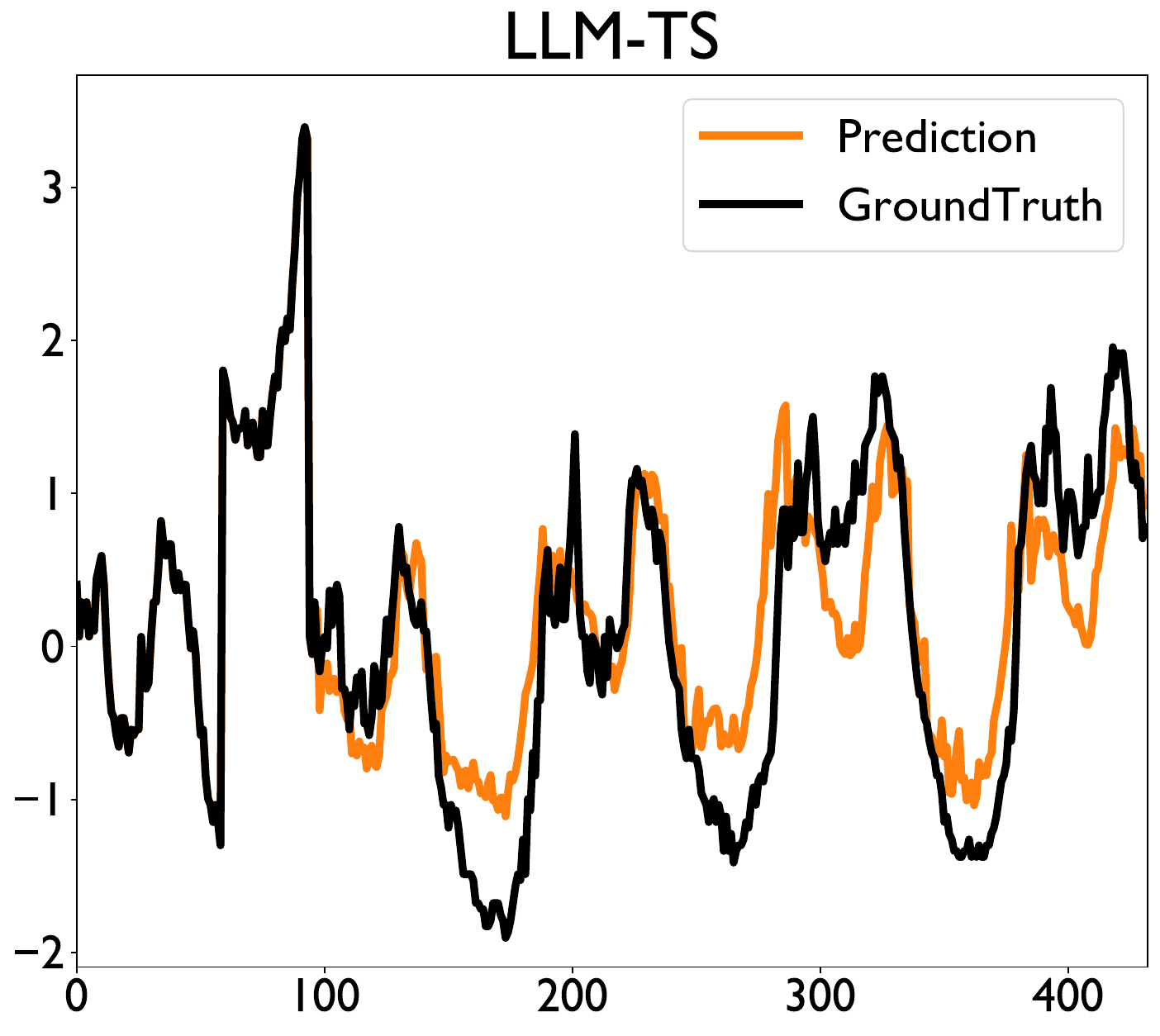}
    \captionsetup{width=1.05\linewidth}
    \label{fig: ettm1_llmts}
\end{minipage}
\begin{minipage}[t]{.33\textwidth}
  \centering
    \includegraphics[width=1.0\columnwidth]{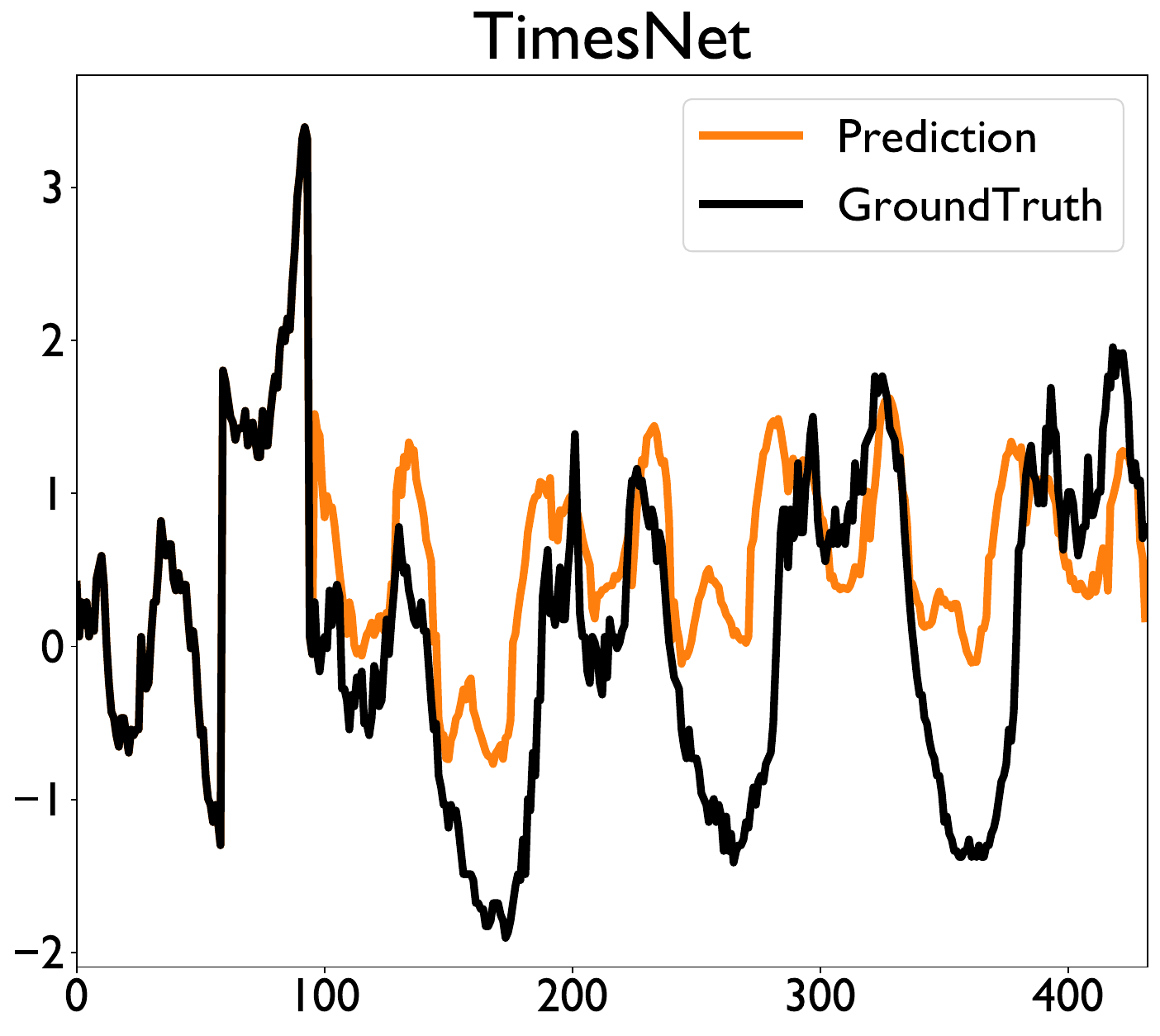}
    \captionsetup{width=.90\linewidth}
    %\caption{a}
    \label{fig: ettm1_timesnet}
\end{minipage}%
\begin{minipage}[t]{.33\textwidth}
  \centering
    \includegraphics[width=1.00\columnwidth]{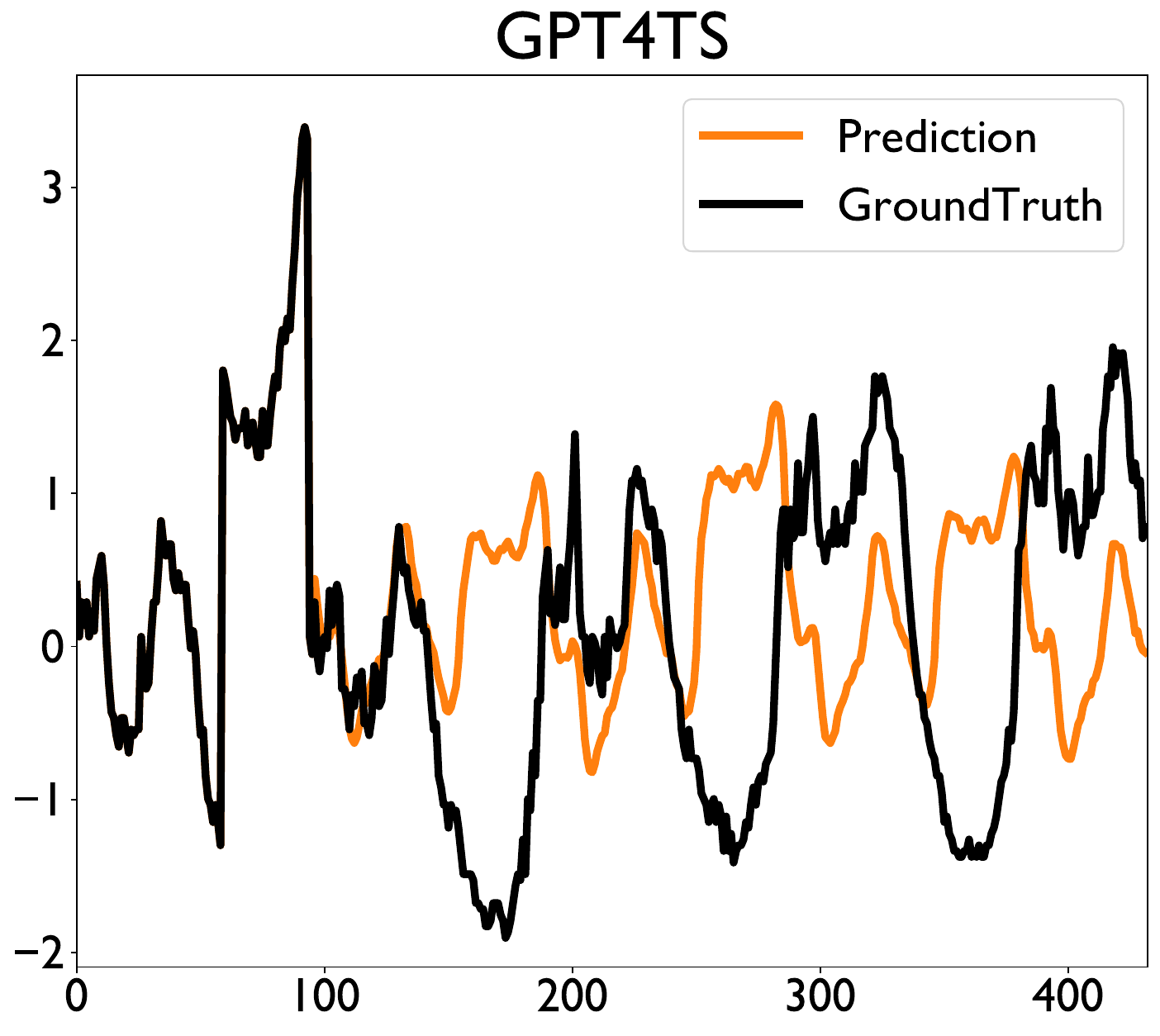}
    \captionsetup{width=.90\linewidth}
    \label{fig: ettm1_gpt4ts}
\end{minipage}
\caption{ETTm1}
\label{fig: ETTm1_show}
\end{figure*}

To provide a clear comparison among different models, we showcase the forecasting task results on ETTh1 ($96$-$96$) and ETTm1 ($96$-$336$) using three models: \textit{LLM-TS}, TimesNet, and GPT4TS. As shown in Figures~\ref{fig: ETTh1_show} and \ref{fig: ETTm1_show}, our LLM-TS model produces significantly more accurate predictions, demonstrating its effectiveness.

To illustrate the performance improvements achieved by the LLM-TS Integrator framework, we introduce a case study. We created a training set with a weighted sine function:
\begin{equation}
\sum_{i=1}^{4}\omega_i \sin(f_i t + p_i) + \epsilon N(0,1) 
\end{equation}
where \(w_1 = 0.1\), \(w_2 = 0.2\), \(w_3 = 0.3\), \(w_4 = 0.4\); \(f_1 = \frac{1}{40}\), \(f_2 = \frac{1}{45}\), \(f_3 = \frac{1}{50}\), \(f_4 = \frac{1}{55}\); \(p_1 = 0\), \(p_2 = 1\), \(p_3 = 2\), \(p_4 = 3\); and \(\epsilon = 0.1\) is the noise level. We generated a long sequence of length $10,000$ and then sampled a batch of size $64$ with a sequence length of $96$ and a prediction length of $336$ to train GPT4TS, TimesNet, and LLM-TS on this data for $1,000$ iterations. For testing, we created a test set with frequency \(f = \frac{1}{20}\), which is greater than \(\max(f_1, f_2, f_3, f_4)\), and used \(p = 2.5\), \(w = 1\) and \(\epsilon = 0.1\).

\begin{figure}
    \centering
    \includegraphics[width=0.9\linewidth]{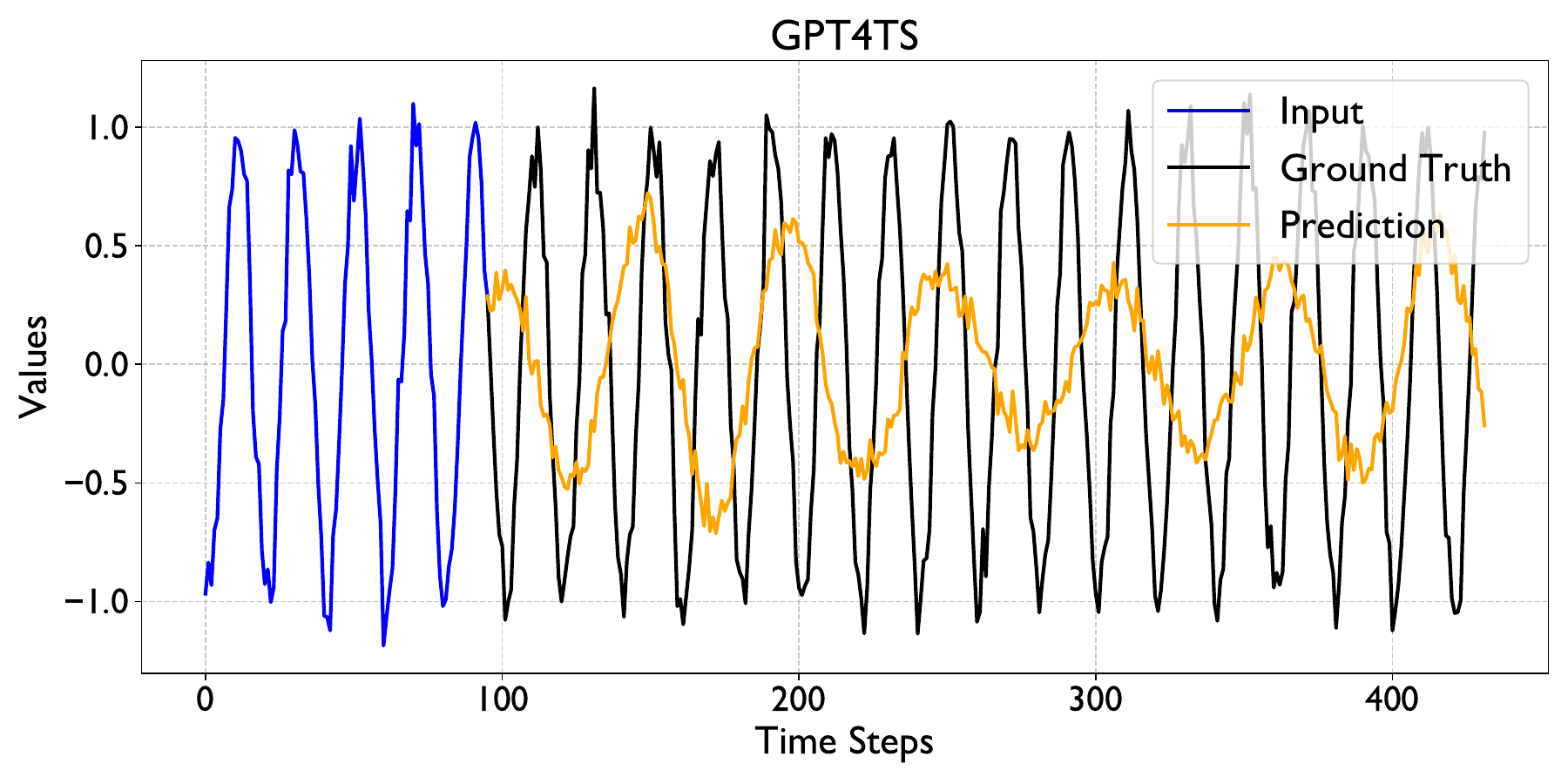}
    \caption{GPT4TS on synthetic data}
    \label{fig: rebuttal_gpt4ts}
\end{figure}

\begin{figure}
    \centering
    \includegraphics[width=0.9\linewidth]{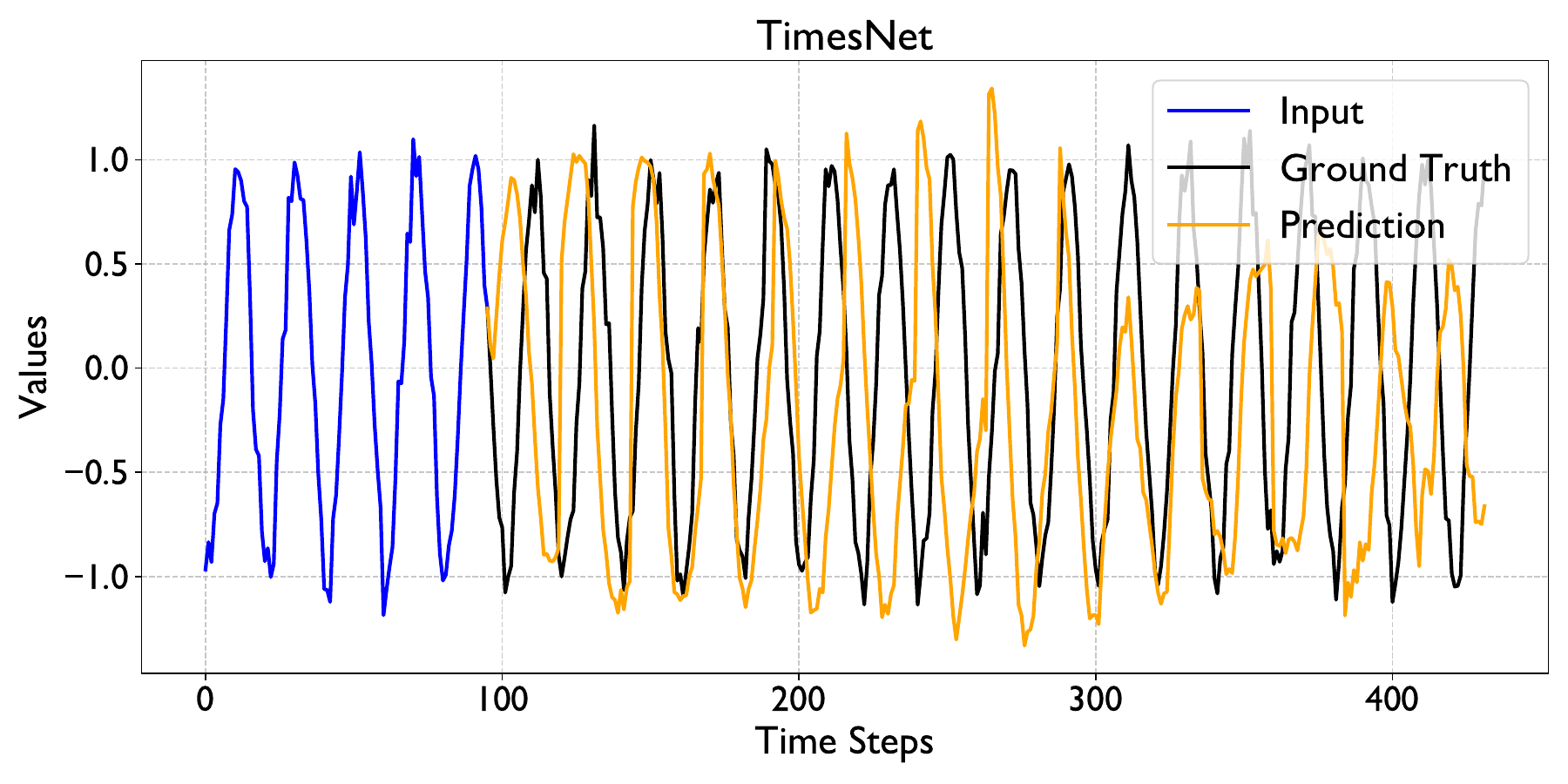}
    \caption{TimesNet on synthetic data}
    \label{fig: rebuttal_timesnet}
\end{figure}

\begin{figure}
    \centering
    \includegraphics[width=0.9\linewidth]{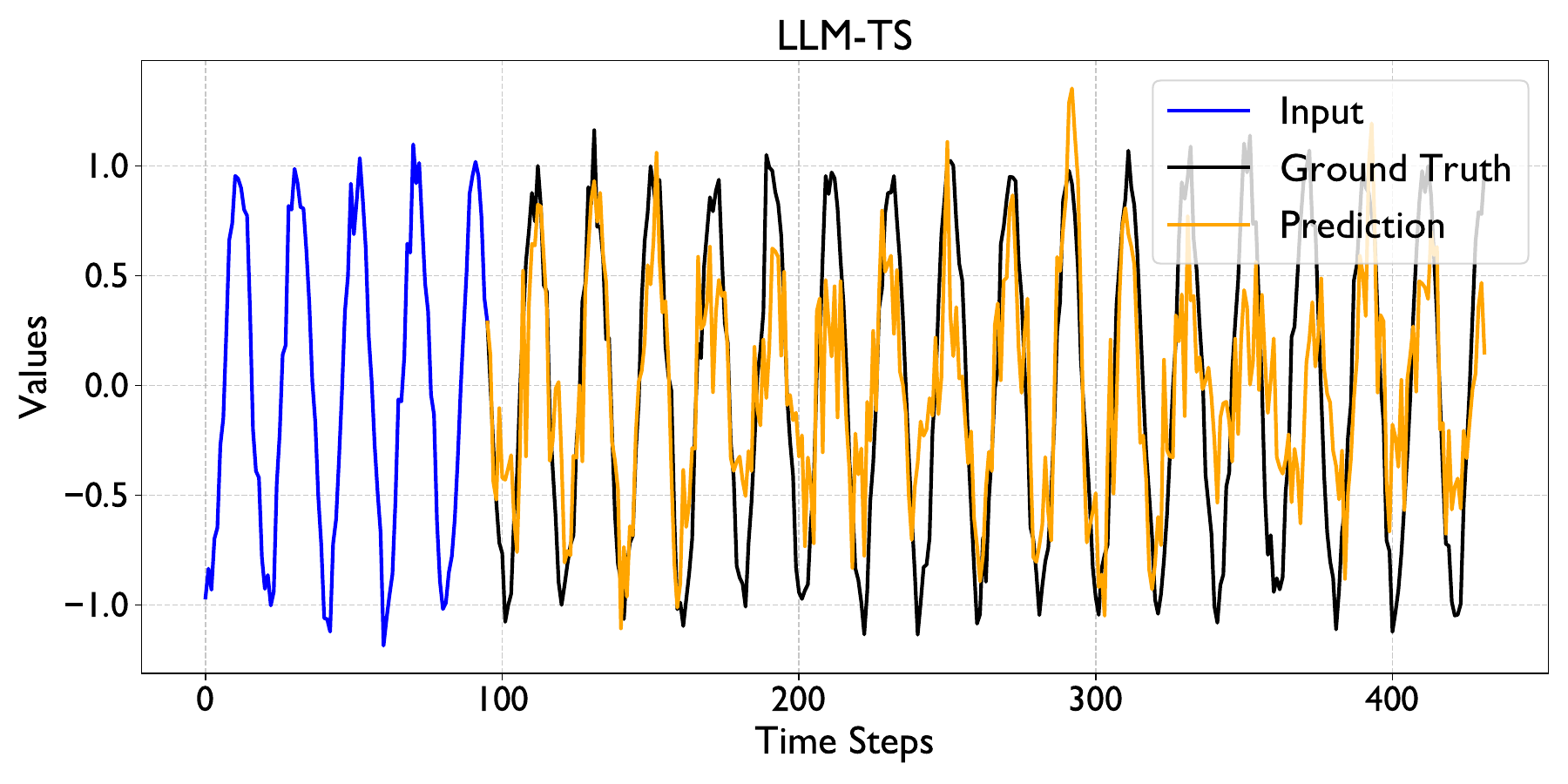}
    \caption{LLM-TS on synthetic data}
    \label{fig: rebuttal_llmts}
\end{figure}

As shown in Figure~\ref{fig: rebuttal_gpt4ts}, Figure~\ref{fig: rebuttal_timesnet} an Figure~\ref{fig: rebuttal_llmts}, we can know:
\begin{itemize}
    \item GPT4TS fails to accurately capture periodic information as it relies solely on a language model without incorporating traditional mathematical modelling.
    \item TimesNet generally captures periodic information due to the use of the FFT mathematical operator, but it still does not perfectly match the ground truth.
    \item LLM-TS captures periodic information and better matches the ground truth by integrating rich language model insights into the traditional TimesNet model.
\end{itemize}

This case study highlights how the LLM-TS Integrator framework benefits from both inherent properties of traditional TS models and pattern recognition abilities of LLMs, demonstrating the effectiveness of our approach.

\subsection{Template Variation}
\label{appendix: template}

We conducted additional experiments on the ETTh1 dataset for long-term forecasting with GPT2. The original template achieves a Mean Squared Error (MSE) of 0.464 and a Mean Absolute Error (MAE) of 0.458. We tested variations of the template by changing the original context from "The Electricity Transformer Temperature is a crucial indicator in electric power long-term deployment." to:
\begin{itemize}
    \item Variation 1: "The temperature of the electricity transformer is a vital metric for long-term electric power deployment."
    \item Variation 2: "Monitoring the temperature of electricity transformers is essential for the long-term deployment of electric power."
    \item Variation 3: "The temperature of electricity transformers serves as a key indicator in the long-term deployment of electric power."
    \item Variation 4: No template.
\end{itemize}
Besides, we also consider the following changes:
\begin{itemize}
    \item w/o Input Statistics: excluding input statistical data from our analysis.
    \item w/o Mean, Max, Median: remove mean, max and median informatino.
    \item w/o Lags: remove lags information.
\end{itemize}
    
The performance of these variations is summarized in Table~\ref{tab: template_variation}.
These results indicate that the performance is quite similar across different variations, supporting the robustness of our approach regardless of minor template modifications.
For further details on the template implementation, refer to our code repository at \url{https://anonymous.4open.science/r/llm_ts_anonymous-F07D/utils/tools.py}.

\begin{table}[h]
\caption{Performance across different template variations}
\label{tab: template_variation}
\centering
\begin{tabular}{@{}lcc@{}}
\toprule
Template Variation & MSE & MAE \\ 
\midrule
Original Template & $0.464 \pm 0.004$ & $0.458 \pm 0.005$ \\
\midrule
Variation 1 & $0.460 \pm 0.005$ & $0.456 \pm 0.003$ \\
Variation 2 & $0.465 \pm 0.006$ & $0.460 \pm 0.005$ \\
Variation 3 & $0.464 \pm 0.004$ & $0.459 \pm 0.003$ \\
Variation 4 (No template) & $0.466 \pm 0.005$ & $0.460 \pm 0.005$ \\
\midrule
w/o Input Statistics & $0.468 \pm 0.004$ & $0.462 \pm 0.004$ \\
w/o Mean/Max/Median & $0.465 \pm 0.004$ & $0.459 \pm 0.003$ \\
w/o Lags & $0.467 \pm 0.003$ & $0.460 \pm 0.005$ \\
\bottomrule
\end{tabular}
\end{table}

\subsection{Model Efficiency Analysis}
\label{appendix: efficiency}
Compared to TimesNet, our \textit{LLM-TS integrator} introduces additional costs due to the mutual information and sample weighting modules. However, after training, the inference cost of our method is the same as TimesNet. We detail the time cost of each component for ETTh1 and ETTm1 tasks, using a batch size of $32$ on a 32G V100 GPU.
As shown in Table~\ref{tab:cost_compare}, the training cost of our method is reasonable, given that it achieves the best performance across most tasks.

It is important to note that we use the pre-trained LLM to obtain the text embeddings only once. These embeddings can then be used throughout the training process. For instance, obtaining the embeddings for the ETTh1 dataset using the llama-3b model on an A100 GPU takes approximately $1$ hour. After this, the embeddings are utilized in our framework to train the model, and in the final output of the TimesNet model. This ensures that the inference time of our method is identical to that of the TimesNet model.

As detailed in the TimesNet paper, our backbone model TimesNet is relatively small with $0.067$ MB parameters. For comparison, other models have the following sizes: Non-stationary Transformer has $1.884$ MB, Autoformer has $1.848$ MB, FEDformer has $2.9$ MB, LightTS has $0.163$ MB, DLinear has $0.296$ MB, ETSformer has $1.123$ MB, Informer has $1.903$ MB, Reformer has $1.157$ MB, and Pyraformer has $1.308$ MB. The introduced mutual information network consists of only two linear layers of size $64x64$ and $64x4096$, which is negligible in terms of additional parameters. Similarly, the introduced MLP network consists of four layers: $1\times100$, $100\times1$, $1\times1$, and $1\times1$, and the number of parameters is also negligible.

Thus, our model remains very small and efficient, with inference time identical to TimesNet (as the mutual information component is only used during training). Given that many TS models are primarily used for inference, our approach offers effective performance gains with minimal additional computational cost.

\begin{table*}[htb]
\vskip -0.10in
\captionsetup{font=small} 
\caption{Cost Comparison per step(s).}
\label{tab:cost_compare}
%\vskip 0.15in
\begin{center}
\begin{small}
\scalebox{0.7}{
\setlength\tabcolsep{3pt}
\begin{tabular}{c|cccc}
\toprule

Methods & Overall & TimesNet & Mutual Information & Sample Reweighting  \\

\midrule

ETTh1 & $3.177$ & $0.126$ & $0.577$ & $2.474$ \\
Weather & $5.563$ & $0.436$ & $1.094$ & $4.033$ \\

\bottomrule

\end{tabular}
}
\end{small}
\end{center}
\end{table*}

%text2embed
%0.97s ETTh1
%0.85s ETTh2
%0.84s ETTm1
%0.62s ETTm2

\subsection{Full Results of Short-term Forecasting}
\label{appendix:short-term_full}

Table~\ref{tab: short_term_full} displays the comprehensive results for short-term forecasting.
\begin{table*}[htb]
\renewcommand\arraystretch{1.5}
\vskip -0.10in
\captionsetup{font=small} 
\caption{Full results of short-term forecasting.}
\label{tab: short_term_full}
%\vskip 0.15in
\begin{center}
\begin{small}
\scalebox{0.7}{
\setlength\tabcolsep{3pt}
\begin{tabular}{cc|cccccccccc}
\toprule

\multicolumn{2}{c|}{Methods}& LLM-TS &TimesNet& GPT4TS& TIME-LLM &PatchTST&N-HiTS&N-BEATS &FEDformer &Stationary & Autoformer  \\

% \multirow{2}{*}{Methods} 
% &\multicolumn{2}{c|}{GPT2(6)} & \multicolumn{2}{c|}{TimesNet}&\multicolumn{2}{c|}{ETSformer}&\multicolumn{2}{c|}{ETSformer}&\multicolumn{2}{c|}{LightTS}&\multicolumn{2}{c|}{DLinear}&\multicolumn{2}{c|}{FEDformer}&\multicolumn{2}{c|}{Stationary}&\multicolumn{2}{c|}{Autoformer}&\multicolumn{2}{c}{Informer}&\multicolumn{2}{c}{Reformer} \\

\midrule
\multirow{3}{*}{\rotatebox{90}{$Yearly$}}
&$SMAPE$&$13.369$&${13.512}$&$13.531$&$13.419$&$13.477$&$13.418$&$13.436$&$13.728$&$13.717$&$13.974$\\
&$MASE$&$3.021$&${3.065}$&$3.015$&$3.0050$&$3.019$&$3.045$&$3.043$&$3.048$&$3.078$&$3.134$\\
&$OWA$&$0.789$&${0.799}$&$0.793$&$0.789$&$0.792$&$0.793$&$0.794$&$0.803$&$0.807$&$0.822$\\
\bottomrule

\multirow{3}{*}{\rotatebox{90}{$Quarterly$}}
&$SMAPE$&$10.020$&${10.069}$&$10.177$&$10.110$&$10.38$&$10.202$&$10.124$&$10.792$&$10.958$&$11.338$\\
&$MASE$&$1.162$&${1.178}$&$1.194$&$1.178$&$1.233$&$1.194$&$1.169$&$1.283$&$1.325$&$1.365$\\
&$OWA$&$0.878$&${0.887}$&$0.898$&$0.889$&$0.921$&$0.899$&$0.886$&$0.958$&$0.981$&$1.012$\\
\bottomrule

\multirow{3}{*}{\rotatebox{90}{$Monthly$}}
&$SMAPE$&$12.696$&${12.783}$&$12.894$&$12.980$&$12.959$&$12.791$&$12.677$&$14.260$&$13.917$&$13.958$\\
&$MASE$&$0.936$&${0.949}$&$0.956$&$0.963$&$0.970$&$0.969$&$0.937$&$1.102$&$1.097$&$1.103$\\
&$OWA$&$0.880$&${0.889}$&$0.897$&$0.903$&$0.905$&$0.899$&$0.880$&$1.012$&$0.998$&$1.002$\\
\bottomrule

\multirow{3}{*}{\rotatebox{90}{$Others$}}
&$SMAPE$&$4.916$&${4.954}$&$4.940$&$4.795$&$4.952$&$5.061$&$4.925$&$4.954$&$6.302$&$5.485$\\
&$MASE$&$3.310$&$3.364$&$3.228$&$3.178$&$3.347$&${3.216}$&$3.391$&$3.264$&$4.064$&$3.865$\\
&$OWA$&$1.039$&${1.052}$&$1.029$&$1.006$&$1.049$&$1.040$&$1.053$&$1.036$&$1.304$&$1.187$\\
\bottomrule

\multirow{3}{*}{\rotatebox{90}{$Average$}}
&$SMAPE$&$11.819 $&${11.908}$&$11.991$&$11.983$&$12.059$&$ 11.927$&$ 11.851$&$ 12.840 $&$12.780 $&$12.909$ \\
&$MASE$&$1.588 $&$ {1.612} $&$ 1.600$&$1.595$&$1.623 $&$ 1.613 $&$ 1.599 $&$1.701 $&$1.756 $&$1.771$  \\
&$OWA$&$0.851 $&${0.860} $&$ 0.861$&$0.859$&$0.869 $&$0.861 $&$0.855 $&$0.918 $&$0.930 $&$0.939$ \\

\bottomrule

\end{tabular}
}
\end{small}
\end{center}
\end{table*}

\subsection{Full Results of Long-Term Forecasting}
\label{appendix:long-term_full}
Full results for long-term forecasting are presented in Table~\ref{tab:long-term_full}.

\begin{table*}[htb]
\caption{Full results for long-term forecasting. We use prediction length $O \in \{96, 192, 336, 720\}$ except for ILI and $O \in \{24, 36, 48, 60\}$ for ILI. A lower MSE indicates better performance.}
\label{tab:long-term_full}
\vskip 0.15in
\begin{center}
\begin{small}
\scalebox{0.70}{
\setlength\tabcolsep{3pt}
\begin{tabular}{cc|cc|cc|cc|cc|cc|cc|cc|cc|cc|cc}
\toprule

\multicolumn{2}{c|}{Methods}&\multicolumn{2}{c|}{LLM-TS}&\multicolumn{2}{c|}{TimesNet}&\multicolumn{2}{c|}{TIME-LLM}&\multicolumn{2}{c|}{DLinear}&\multicolumn{2}{c|}{PatchTST}&\multicolumn{2}{c|}{GPT4TS}&\multicolumn{2}{c|}{FEDformer}&\multicolumn{2}{c|}{TEST}&\multicolumn{2}{c|}{Stationary}&\multicolumn{2}{c}{ETSformer} \\

\midrule

\multicolumn{2}{c|}{Metric} & MSE  & MAE & MSE & MAE& MSE & MAE& MSE  & MAE& MSE  & MAE& MSE  & MAE& MSE  & MAE& MSE  & MAE& MSE & MAE& MSE  & MAE  \\
\midrule

\multirow{5}{*}{\rotatebox{90}{$Weather$}}
&$ 96  $&$ 0.166 $&$ 0.217 $&$ 0.174 $&$ 0.224 $&$ 0.202 $&$ 0.239 $&$ 0.196 $&$ 0.255 $&$ 0.186 $&$ 0.227 $&$ 0.196 $&$ 0.234$&$0.217$&$0.296$&$ 0.214 $&$ 0.264$&$0.173$&$0.223$&$0.197$&$0.281$\\
&$ 192 $&$ 0.229 $&$ 0.269 $&$ 0.235 $&$ 0.272 $&$ 0.245 $&$ 0.277 $&$ 0.237 $&$ 0.296 $&$ 0.234 $&$ 0.265 $&$ 0.241 $&$ 0.271$&$0.276$&$0.336$&$ 0.262 $&$ 0.298$&$0.245$&$0.285$&$0.237$&$0.312$\\
&$ 336 $&$ 0.278 $&$ 0.302 $&$ 0.235 $&$ 0.272 $&$ 0.300 $&$ 0.313 $&$ 0.283 $&$ 0.335 $&$ 0.284 $&$ 0.301 $&$ 0.296 $&$ 0.308$&$0.339$&$0.380$&$ 0.310 $&$ 0.329$&$0.321$&$0.338$&$0.298$&$0.353$\\
&$ 720 $&$ 0.354 $&$ 0.351 $&$ 0.365 $&$ 0.358 $&$ 0.369 $&$ 0.356 $&$ 0.345 $&$ 0.381 $&$ 0.356 $&$ 0.349 $&$ 0.367 $&$ 0.354$&$0.403$&$0.428$&$ 0.378 $&$ 0.370$&$0.414$&$0.410$&$0.352$&$0.288$\\
&$ Avg $&$ 0.257 $&$ 0.285 $&$ 0.265 $&$ 0.290 $&$ 0.279 $&$ 0.296 $&$ 0.265 $&$ 0.317 $&$ 0.265 $&$ 0.285 $&$ 0.275 $&$ 0.292$&$0.309$&$0.360$&$ 0.291 $&$ 0.315$&$0.288$&$0.314$&$0.271$&$0.334$\\
\midrule

\multirow{5}{*}{\rotatebox{90}{$ETTh1$}}
&$ 96  $&$ 0.403 $&$ 0.420 $&$ 0.414 $&$ 0.431 $&$ 0.414 $&$ 0.422 $&$ 0.386 $&$ 0.400 $&$ 0.460 $&$ 0.447 $&$ 0.409 $&$ 0.415$&$0.376$&$0.419$&$ 0.411 $&$ 0.426$&$0.513$&$0.491$&$0.494$&$0.479$\\
&$ 192 $&$ 0.440 $&$ 0.441 $&$ 0.463 $&$ 0.456 $&$ 0.466 $&$ 0.450 $&$ 0.437 $&$ 0.432 $&$ 0.512 $&$ 0.477 $&$ 0.468 $&$ 0.446$&$0.420$&$0.448$&$ 0.475 $&$ 0.461$&$0.534$&$0.504$&$0.538$&$0.504$\\
&$ 336 $&$ 0.471 $&$ 0.457 $&$ 0.487 $&$ 0.466 $&$ 0.515 $&$ 0.475 $&$ 0.481 $&$ 0.459 $&$ 0.546 $&$ 0.496 $&$ 0.503 $&$ 0.461$&$0.459$&$0.465$&$ 0.508 $&$ 0.482$&$0.588$&$0.535$&$0.574$&$0.521$\\
&$ 720 $&$ 0.503 $&$ 0.487 $&$ 0.517 $&$ 0.494 $&$ 0.503 $&$ 0.487 $&$ 0.519 $&$ 0.516 $&$ 0.544 $&$ 0.517 $&$ 0.510 $&$ 0.482$&$0.506$&$0.507$&$ 0.504 $&$ 0.494$&$0.643$&$0.616$&$0.562$&$0.535$\\
&$ Avg $&$ 0.454 $&$ 0.451 $&$ 0.470 $&$ 0.462 $&$ 0.474 $&$ 0.459 $&$ 0.456 $&$ 0.452 $&$ 0.516 $&$ 0.484 $&$ 0.473 $&$ 0.451$&$0.440$&$0.460$&$ 0.475 $&$ 0.466$&$0.570$&$0.537$&$0.542$&$0.510$\\
\midrule

\multirow{5}{*}{\rotatebox{90}{$ETTh2$}}
&$ 96  $&$ 0.322 $&$ 0.366 $&$ 0.340 $&$ 0.374 $&$ 0.306 $&$ 0.353 $&$ 0.333 $&$ 0.387 $&$ 0.308 $&$ 0.355$&$ 0.298 $&$ 0.350$&$0.358$&$0.397$&$ 0.328 $&$ 0.374$&$0.476$&$0.458$&$0.340$&$0.391$ \\
&$ 192 $&$ 0.400 $&$ 0.409 $&$ 0.399 $&$ 0.410 $&$ 0.386 $&$ 0.399 $&$ 0.477 $&$ 0.476 $&$ 0.393 $&$ 0.405$&$ 0.376 $&$ 0.399$&$0.429$&$0.439$&$ 0.403 $&$ 0.418$&$0.512$&$0.493$&$0.430$&$0.439$\\
&$ 336 $&$ 0.432 $&$ 0.435 $&$ 0.452 $&$ 0.452 $&$ 0.460 $&$ 0.458 $&$ 0.594 $&$ 0.541 $&$ 0.427 $&$ 0.436$&$ 0.430 $&$ 0.439$&$0.496$&$0.487$&$ 0.455 $&$ 0.458$&$0.552$&$0.551$&$0.485$&$0.479$ \\
&$ 720 $&$ 0.430 $&$ 0.442 $&$ 0.462 $&$ 0.468 $&$ 0.442 $&$ 0.451 $&$ 0.831 $&$ 0.657 $&$ 0.436 $&$ 0.450$&$ 0.428 $&$ 0.451$&$0.463$&$0.474$&$ 0.470 $&$ 0.477$&$0.562$&$0.560$&$0.500$&$0.497$\\
&$ Avg $&$ 0.396 $&$ 0.413 $&$ 0.413 $&$ 0.426 $&$ 0.398 $&$ 0.415 $&$ 0.559 $&$ 0.515 $&$ 0.391 $&$ 0.411$&$ 0.383 $&$ 0.410$&$0.437$&$0.449$&$ 0.414 $&$ 0.432$&$0.526$&$0.516$&$0.439$&$0.452$\\
\midrule

\multirow{5}{*}{\rotatebox{90}{$ETTm1$}}
&$ 96  $&$ 0.329 $&$ 0.371 $&$ 0.340 $&$ 0.377 $&$ 0.393 $&$ 0.398 $&$ 0.345 $&$ 0.372 $&$ 0.352 $&$ 0.374 $&$ 0.350 $&$ 0.369 $&$0.379$&$0.419$&$ 0.336 $&$ 0.373$&$0.386$&$0.398$&$0.375$&$0.398$ \\
&$ 192 $&$ 0.380 $&$ 0.398 $&$ 0.406 $&$ 0.408 $&$ 0.412 $&$ 0.405 $&$ 0.380 $&$ 0.389 $&$ 0.390 $&$ 0.393 $&$ 0.387 $&$ 0.387 $&$0.426$&$0.441$&$ 0.381 $&$ 0.399$&$0.459$&$0.444$&$0.408$&$0.410$\\
&$ 336 $&$ 0.418 $&$ 0.425 $&$ 0.424 $&$ 0.425 $&$ 0.442 $&$ 0.425 $&$ 0.413 $&$ 0.413 $&$ 0.421 $&$ 0.414 $&$ 0.418 $&$ 0.407 $&$0.445$&$0.459$&$ 0.411 $&$ 0.418$&$0.495$&$0.464$&$0.435$&$0.428$\\
&$ 720 $&$ 0.476 $&$ 0.440 $&$ 0.485 $&$ 0.461 $&$ 0.502 $&$ 0.457 $&$ 0.474 $&$ 0.453 $&$ 0.462 $&$ 0.449 $&$ 0.477 $&$ 0.437 $&$0.543$&$0.490$&$ 0.478 $&$ 0.454$&$0.585$&$0.516$&$0.499$&$0.462$\\
&$ Avg $&$ 0.401 $&$ 0.409 $&$ 0.414 $&$ 0.418 $&$ 0.437 $&$ 0.421 $&$ 0.403 $&$ 0.407 $&$ 0.406 $&$ 0.407 $&$ 0.408 $&$ 0.400 $&$0.448$&$0.452$&$ 0.402 $&$ 0.411$&$0.481$&$0.456$&$0.429$&$0.425$\\
\midrule

\multirow{5}{*}{\rotatebox{90}{$ETTm2$}}
&$ 96  $&$ 0.189 $&$ 0.266 $&$ 0.185 $&$ 0.264 $&$ 0.193 $&$ 0.281 $&$ 0.193 $&$ 0.292 $&$ 0.183 $&$ 0.270 $&$ 0.185 $&$ 0.271$&$0.203$&$0.287$&$ 0.230 $&$ 0.307$&$0.192$&$0.274$&$0.189$&$0.280$\\
&$ 192 $&$ 0.253 $&$ 0.307 $&$ 0.252 $&$ 0.306 $&$ 0.254 $&$ 0.315 $&$ 0.284 $&$ 0.363 $&$ 0.255 $&$ 0.314 $&$ 0.250 $&$ 0.312$&$0.269$&$0.328$&$ 0.284 $&$ 0.338$&$0.280$&$0.339$&$0.253$&$0.319$\\
&$ 336 $&$ 0.315 $&$ 0.345 $&$ 0.323 $&$ 0.350 $&$ 0.320 $&$ 0.355 $&$ 0.369 $&$ 0.427 $&$ 0.309 $&$ 0.347 $&$ 0.314 $&$ 0.351$&$0.325$&$0.366$&$ 0.340 $&$ 0.370$&$0.334$&$0.361$&$0.314$&$0.357$\\
&$ 720 $&$ 0.421 $&$ 0.408 $&$ 0.415 $&$ 0.403 $&$ 0.426 $&$ 0.416 $&$ 0.554 $&$ 0.522 $&$ 0.412 $&$ 0.404 $&$ 0.410 $&$ 0.408$&$0.421$&$0.415$&$ 0.436 $&$ 0.420$&$0.417$&$0.413$&$0.414$&$0.413$ \\
&$ Avg $&$ 0.295 $&$ 0.331 $&$ 0.294 $&$ 0.331 $&$ 0.298 $&$ 0.342 $&$ 0.350 $&$ 0.401 $&$ 0.290 $&$ 0.334 $&$ 0.290 $&$ 0.335$&$0.305$&$0.349$&$ 0.323 $&$ 0.359$&$0.306$&$0.347$&$0.293$&$0.342$\\
\midrule

\multirow{5}{*}{\rotatebox{90}{$ILI$}}
&$ 24  $&$ 1.921 $&$ 0.898 $&$ 2.072 $&$ 0.948 $&$ 2.589 $&$ 1.054 $&$ 2.398 $&$ 1.040 $&$ 2.229 $&$ 0.894 $&$ 5.259 $&$ 1.689$&$3.228$&$1.260$&$ 3.371 $&$ 1.231$&$2.294$&$0.945$&$2.527$&$1.020$\\
&$ 36  $&$ 2.151 $&$ 0.933 $&$ 2.494 $&$ 1.019 $&$ 2.996 $&$ 1.194 $&$ 2.646 $&$ 1.088 $&$ 2.330 $&$ 0.925 $&$ 6.136 $&$ 1.831$&$2.679$&$1.080$&$ 3.725 $&$ 1.322$&$1.825$&$0.848$&$2.615$&$1.007$\\
&$ 48  $&$ 2.062 $&$ 0.892 $&$ 2.298 $&$ 0.964 $&$ 2.714 $&$ 1.095 $&$ 2.614 $&$ 1.086 $&$ 2.140 $&$ 0.894 $&$ 4.670 $&$ 1.562$&$2.622$&$1.078$&$ 3.291 $&$ 1.237$&$2.010$&$0.900$&$2.359$&$0.972$\\
&$ 60  $&$ 1.759 $&$ 0.853 $&$ 2.198 $&$ 0.963 $&$ 2.605 $&$ 1.050 $&$ 2.804 $&$ 1.146 $&$ 2.037 $&$ 0.912 $&$ 4.402 $&$ 1.517$&$2.857$&$1.157$&$ 2.907 $&$ 1.136$&$2.178$&$0.963$&$2.487$&$1.016$ \\
&$ Avg $&$ 1.973 $&$ 0.894 $&$ 2.266 $&$ 0.974 $&$ 2.726 $&$ 1.098 $&$ 2.616 $&$ 1.090 $&$ 2.184 $&$ 0.906 $&$ 5.117 $&$ 1.650$&$2.847$&$1.144$&$ 3.324 $&$ 1.232$&$2.077$&$0.914$&$2.497$&$1.004$\\
\midrule

\multirow{5}{*}{\rotatebox{90}{$ECL$}}
&$ 96  $&$ 0.167 $&$ 0.271 $&$ 0.169 $&$ 0.273 $&$ 0.207 $&$ 0.292 $&$ 0.197 $&$ 0.282 $&$ 0.190 $&$ 0.296 $&$ 0.186 $&$ 0.273 $&$0.193$&$0.308$&$ 0.218 $&$ 0.309$&$0.169$&$0.273$&$0.187$&$0.304$ \\
&$ 192 $&$ 0.178 $&$ 0.280 $&$ 0.186 $&$ 0.288 $&$ 0.209 $&$ 0.297 $&$ 0.196 $&$ 0.285 $&$ 0.199 $&$ 0.304 $&$ 0.190 $&$ 0.278 $&$0.201$&$0.315$&$ 0.220 $&$ 0.311$&$0.182$&$0.286$&$0.199$&$0.315$\\
&$ 336 $&$ 0.198 $&$ 0.302 $&$ 0.206 $&$ 0.305 $&$ 0.224 $&$ 0.312 $&$ 0.209 $&$ 0.301 $&$ 0.217 $&$ 0.319 $&$ 0.204 $&$ 0.291 $&$0.214$&$0.329$&$ 0.234 $&$ 0.323$&$0.200$&$0.304$&$0.212$&$0.329$\\
&$ 720 $&$ 0.233 $&$ 0.344 $&$ 0.231 $&$ 0.327 $&$ 0.277 $&$ 0.359 $&$ 0.245 $&$ 0.333 $&$ 0.258 $&$ 0.352 $&$ 0.245 $&$ 0.297 $&$0.325$&$0.355$&$ 0.276 $&$ 0.354$&$0.222$&$0.321$&$0.233$&$0.345$\\
&$ Avg $&$ 0.194 $&$ 0.299 $&$ 0.198 $&$ 0.298 $&$ 0.229 $&$ 0.315 $&$ 0.212 $&$ 0.300 $&$ 0.216 $&$ 0.318 $&$ 0.206 $&$ 0.285 $&$0.214$&$0.327$&$ 0.237 $&$ 0.324$&$0.193$&$0.296$&$0.208$&$0.323$\\
\midrule

\multirow{5}{*}{\rotatebox{90}{$Traffic$}}
&$ 96  $&$ 0.587 $&$ 0.315 $&$ 0.589 $&$ 0.313 $&$ 0.609 $&$ 0.402 $&$ 0.650 $&$ 0.396 $&$ 0.526 $&$ 0.347 $&$ 0.563 $&$ 0.378$&$0.587$&$0.366$&$ 0.589 $&$ 0.390$&$0.612$&$0.338$&$0.607$&$0.392$\\
&$ 192 $&$ 0.612 $&$ 0.326 $&$ 0.627 $&$ 0.337 $&$ 0.586 $&$ 0.382 $&$ 0.598 $&$ 0.370 $&$ 0.522 $&$ 0.332 $&$ 0.549 $&$ 0.367$&$0.604$&$0.373$&$ 0.567 $&$ 0.380$&$0.613$&$0.340$&$0.621$&$0.399$\\
&$ 336 $&$ 0.634 $&$ 0.338 $&$ 0.635 $&$ 0.341 $&$ 0.593 $&$ 0.390 $&$ 0.605 $&$ 0.373 $&$ 0.517 $&$ 0.334 $&$ 0.566 $&$ 0.376$&$0.621$&$0.383$&$ 0.583 $&$ 0.389$&$0.618$&$0.328$&$0.622$&$0.396 $\\
&$ 720 $&$ 0.640 $&$ 0.351 $&$ 0.658 $&$ 0.349 $&$ 0.636 $&$ 0.405 $&$ 0.645 $&$ 0.394 $&$ 0.552 $&$ 0.352 $&$ 0.567 $&$ 0.372$&$0.626$&$0.382$&$ 0.585 $&$ 0.391$&$0.653$&$0.355$&$0.632$&$0.396$\\
&$ Avg $&$ 0.618 $&$ 0.333 $&$ 0.627 $&$ 0.335 $&$ 0.606 $&$ 0.395 $&$ 0.625 $&$ 0.383 $&$ 0.529 $&$ 0.341 $&$ 0.561 $&$ 0.373$&$0.610$&$0.376$&$ 0.581 $&$ 0.388$&$0.624$&$0.340$&$0.621$&$0.396$\\
\midrule
\multicolumn{2}{c|}{Average}&$ 0.574 $&$ 0.427 $&$0.618 $&$0.442$&$ 0.681 $&$0.468$&$0.686$&$0.483$&$0.600$&$0.436$&$0.964$&$0.525$&$0.701$&$0.489$&$ 0.756 $&$ 0.491$&$0.633$&$0.465$&$0.662$&$0.473$\\

\bottomrule
\end{tabular}
}
\end{small}
\end{center}
\end{table*}

\subsection{Full Results of Imputation.}
\label{appendix:imputation_full}
Table~\ref{tab:imputation_full} contains the detailed results of our imputation tasks. 
\begin{table*}[htb]
\caption{Full results for the imputation task. Randomly masked \{$12.5$\%, $25$\%, $37.5$\%, $50$\%\} of points in $96$-length series, averaging results over $4$ mask ratios.}
% We randomly mask {12.5\%, 25\%, 37.5\%, 50\%} to compare the model performance under different missing degrees. A lower MSE indicates better performance. \textbf{Black}: best,  {\textbf{Red}}: second best.}
\label{tab:imputation_full}
\begin{center}
\begin{small}
\scalebox{0.65}{
\setlength\tabcolsep{3pt}
\begin{tabular}{cc|cc|cc|cc|cc|cc|cc|cc|cc|cc|cc}
\toprule

\multicolumn{2}{c|}{Methods} &\multicolumn{2}{c|}{LLM-TS }
&\multicolumn{2}{c|}{TimesNet } & \multicolumn{2}{c|}{GPT4TS}&\multicolumn{2}{c|}{PatchTST}&\multicolumn{2}{c|}{LightTS}&\multicolumn{2}{c|}{DLinear}&\multicolumn{2}{c|}{FEDformer}&\multicolumn{2}{c|}{Stationary}&\multicolumn{2}{c|}{Autoformer}&\multicolumn{2}{c}{Reformer} \\
Mask&Ratio&MSE&MAE&MSE&MAE&MSE&MAE&MSE&MAE&MSE&MAE&MSE&MAE&MSE&MAE&MSE&MAE&MSE&MAE&MSE&MAE \\

\midrule
\multirow{5}{*}{\rotatebox{90}{$ETTm1$}}
&$ 12.5\% $&${ 0.018}$&${0.088} $&$0.023$&$0.101$&$ 0.018$&$0.089$&$0.041$&$0.130$&$0.093$&$0.206$&$0.080$&$0.193$&$0.052$&$0.166$&$0.032$&$0.119$&$0.046$&$0.144$&$0.042$&$0.146 $\\
&$ 25\% $&${ 0.022}$&${0.097}$&$0.023$&$0.101$&$ 0.023$&$0.099$&$0.044$&$0.135$&$0.093$&$0.206$&$0.080$&$0.193$&$0.052$&$0.166$&$0.032$&$0.119$&$0.046$&$0.144$&$0.042$&$0.146 $\\
&$ 37.5\% $&${ 0.027}$&${0.108}$&$0.029$&$0.112$&${0.030}$&${0.112}$&$0.049$&$0.143$&$0.113$&$0.231$&$0.103$&$0.219$&$0.069$&$0.191$&$0.039$&$0.131$&$0.057$&$0.161$&$0.063$&$0.182 $\\
&$ 50\% $&${ 0.033}$&${0.120}$&${0.035}$&${0.123}$&$0.042$&$0.131$&$0.055$&$0.151$&$0.134$&$0.255$&$0.132$&$0.248$&$0.089$&$0.218$&$0.047$&$0.145$&$0.067$&$0.174$&$0.082$&$0.208 $\\
&$ Avg $&${ 0.025}$&${0.103}$&${0.028}$&$0.109$&$0.028$&${0.108}$&$0.047$&$0.140$&$0.104$&$0.218$&$0.093$&$0.206$&$0.062$&$0.177$&$0.036$&$0.126$&$0.051$&$0.150$&$0.055$&$0.166 $\\
\midrule

\multirow{5}{*}{\rotatebox{90}{$ETTm2$}}
&$ 12.5\% $&${ 0.018}$&${0.079}$&$0.019$&$0.081$&${0.019}$&${0.078}$&$0.108$&$0.239$&$0.034 $&$ 0.127$&$0.062$&$0.166$&$0.056$&$0.159$&$0.021$&$0.088$&$0.023$&$0.092$&$0.108$&$0.228 $\\
&$ 25\% $&${ 0.020}$&${0.085}$&$0.021$&$0.087$&${0.021}$&${0.084}$&$0.028$&$0.099$&$0.042$&$0.143$&$0.085$&$0.196$&$0.080$&$0.195$&$0.024$&$0.096$&$0.026$&$0.101$&$0.136$&$0.262 $\\
&$ 37.5\% $&${ 0.022}$&${0.089}$&$0.023$&$0.092$&${0.024}$&${0.090}$&$0.030$&$0.104$&$0.051$&$0.159$&$0.106$&$0.222$&$0.110$&$0.231$&$0.027$&$0.103$&$0.030$&$0.108$&$0.175$&$0.300 $\\
&$ 50\% $&${ 0.025}$&${0.096}$&$0.025$&$0.097$&${0.027}$&${0.098}$&$0.034$&$0.110$&$0.059$&$0.174$&$0.131$&$0.247$&$0.156$&$0.276$&$0.030$&$0.108$&$0.035$&$0.119$&$0.211$&$0.329 $\\
&$ Avg $&${ 0.021}$&${0.087}$&$0.022$&$0.089$&${0.023}$&${0.088}$&$0.029$&$0.102$&$0.046$&$0.151$&$0.096$&$0.208$&$0.101$&$0.215$&$0.026$&$0.099$&$0.029$&$0.105$&$0.157$&$0.280 $\\
\midrule

\multirow{5}{*}{\rotatebox{90}{$ETTh1$}}
&$ 12.5\% $&${ 0.058}$&${0.165}$&$0.064$&$0.170$&${0.043}$&${0.141}$&$0.093$&$0.201$&$0.240$&$0.345$&$0.151$&$0.267$&$0.070$&$0.190$&$0.060$&$0.165$&$0.074$&$0.182$&$0.074$&$0.194 $\\
&$ 25\% $&${ 0.077}$&${0.189}$&$0.082$&$0.192$&${0.056}$&${0.159}$&$0.107$&$0.217$&$0.265$&$0.364$&$0.180$&$0.292$&$0.106$&$0.236$&$0.080$&$0.189$&$0.090$&$0.203$&$0.102$&$0.227 $\\
&$ 37.5\% $&${ 0.096}$&${0.209}$&$0.098$&$0.209$&${0.074}$&${0.182}$&$0.120$&$0.230$&$0.296$&$0.382$&$0.215$&$0.318$&$0.124$&$0.258$&$0.102$&$0.212$&$0.109$&$0.222$&$0.135$&$0.261 $\\
&$ 50\% $&${ 0.118}$&${0.228}$&${0.116}$&${0.226}$&$0.104$&$0.214$&$0.141$&$0.248$&$0.334$&$0.404$&$0.257$&$0.347$&$0.165$&$0.299$&$0.133$&$0.240$&$0.137$&$0.248$&$0.179$&$0.298 $\\
&$ Avg $&${ 0.087}$&${0.198}$&$0.090$&$0.199$&${0.069}$&${0.174}$&$0.115$&$0.224$&$0.284$&$0.373$&$0.201$&$0.306$&$0.117$&$0.246$&$0.094$&$0.201$&$0.103$&$0.214$&$0.122$&$0.245 $\\
\midrule

\multirow{5}{*}{\rotatebox{90}{$ETTh2$}}
&$ 12.5\% $&${ 0.039}$&${0.131}$&$0.040$&$0.132$&${0.041}$&${0.129}$&$0.057$&$0.152$&$0.101$&$0.231$&$0.100$&$0.216$&$0.095$&$0.212$&$0.042$&$0.133$&$0.044$&$0.138$&$0.163$&$0.289 $\\
&$ 25\% $&${ 0.046}$&${0.143}$&$0.048$&$0.146$&${0.046}$&${0.137}$&$0.061$&$0.158$&$0.115$&$0.246$&$0.127$&$0.247$&$0.137$&$0.258$&$0.049$&$0.147$&$0.050$&$0.149$&$0.206$&$0.331 $\\
&$ 37.5\% $&${ 0.053}$&${0.154}$&$0.055$&$0.156$&${0.053}$&${0.148}$&$0.067$&$0.166$&$0.126$&$0.257$&$0.158$&$0.276$&$0.187$&$0.304$&$0.056$&$0.158$&$0.060$&$0.163$&$0.252$&$0.370 $\\
&$ 50\% $&${ 0.061}$&${0.165}$&$0.061$&$0.165$&${0.060}$&${0.160}$&$0.073$&$0.174$&$0.136$&$0.268$&$0.183$&$0.299$&$0.232$&$0.341$&$0.065$&$0.170$&$0.068$&$0.173$&$0.316$&$0.419 $\\
&$ Avg $&${ 0.050}$&${0.148}$&$0.051$&$0.150$&${0.050}$&${0.144}$&$0.065$&$0.163$&$0.119$&$0.250$&$0.142$&$0.259$&$0.163$&$0.279$&$0.053$&$0.152$&$0.055$&$0.156$&$0.234$&$0.352 $\\
\midrule

\multirow{5}{*}{\rotatebox{90}{$ECL$}}
&$ 12.5\% $&${ 0.087}$&${0.203}$&$0.090$&$0.204$&${0.080}$&${0.194}$&$0.055$&$0.160$&$0.102$&$0.229$&$0.092$&$0.214$&$0.107$&$0.237$&$0.093$&$0.210$&$0.089$&$0.210$&$0.190$&$0.308 $\\
&$ 25\% $&${ 0.091}$&${0.207}$&$0.092$&$0.209$&${0.087}$&${0.203}$&$0.065$&$0.175$&$0.121$&$0.252$&$0.118$&$0.247$&$0.120$&$0.251$&$0.097$&$0.214$&$0.096$&$0.220$&$0.197$&$0.312 $\\
&$ 37.5\% $&${ 0.095}$&${0.213} $&${0.096}$&$0.213$&$0.094$&${0.211}$&$0.076$&$0.344$&$0.141$&$0.273$&$0.144$&$0.276$&$0.136$&$0.266$&$0.102$&$0.220$&$0.104$&$0.229$&$ 0.203$&$0.315$\\
&$ 50\% $&${ 0.101}$&${0.220}$&${0.102}$&$0.221$&$0.101$&${0.220}$&$0.091$&$0.208$&$0.160$&$0.293$&$0.175$&$0.305$&$0.158$&$0.284$&$0.108$&$0.228$&$0.113$&$0.239$&$0.210$&$0.319 $\\
&$ Avg $&${ 0.094}$&${0.211}$&$0.095$&$0.212$&${0.091}$&${0.207}$&$0.072$&$0.183$&$0.131$&$0.262$&$0.132$&$0.260$&$0.130$&$0.259$&$0.100$&$0.218$&$0.101$&$0.225$&$0.200$&$0.313 $\\
\midrule

\multirow{5}{*}{\rotatebox{90}{$Weather$}}
&$ 12.5\%$&${ 0.026}$&${0.048} $&${0.025}$&${0.047}$&$0.027$&$0.049$&$0.029$&$0.049$&$0.047$&$0.101$&$0.039$&$0.084$&$0.041$&$0.107$&$0.027$&$0.051$&$0.026$&$0.047$&$0.031$&$0.076 $\\
&$ 25\% $&${ 0.029}$&${0.055}$&$0.031$&$0.062$&${0.030}$&${0.054}$&$0.031$&$0.053$&$0.052$&$0.111$&$0.048$&$0.103$&$0.064$&$0.163$&$0.029$&$0.056$&$0.030$&$0.054$&$0.035$&$0.082 $\\
&$ 37.5\% $&${ 0.032}$&${0.059}$&${0.034}$&${0.064}$&$0.034$&$0.062$&$0.035$&$0.058$&$0.058$&$0.121$&$0.057$&$0.117$&$0.107$&$0.229$&$0.033$&$0.062$&$0.032$&$0.060$&$0.040$&$0.091 $\\
&$ 50\% $&${ 0.033}$&${0.061}$&${0.035}$&${0.062}$&$0.037$&$0.066$&$0.038$&$0.063$&$0.065$&$0.133$&$0.066$&$0.134$&$0.183$&$0.312$&$0.037$&$0.068$&$0.037$&$0.067$&$0.046$&$0.099 $\\
&$ Avg $&${ 0.030}$&${0.056}$&$0.031$&$0.059$&${0.032}$&${0.058}$&$0.060$&$0.144$&$0.055$&$0.117$&$0.052$&$0.110$&$0.099$&$0.203$&$0.032$&$0.059$&$0.031$&$0.057$&$0.038$&$0.087 $\\
\bottomrule

\end{tabular}
}
\end{small}
\end{center}
\end{table*}

\subsection{Full Results of Classification}
\label{appendix: classification}

Table~\ref{tab:classification} contains the comprehensive results for classification.
\begin{table*}[htb]
\caption{Complete classification task results. $\ast$. in the Transformers indicates the name of $\ast$former.}
\label{tab:classification}
\vskip 0.15in
\begin{center}
\begin{small}
\scalebox{0.65}{
\setlength\tabcolsep{3pt}
\begin{tabular}{c|cc|cc|c|ccccccccc|cc|c|cc|c}
\toprule

\multirow{2}{*}{Methods} &
\multicolumn{2}{c|}{Classical} & \multicolumn{2}{c|}{RNN} & \multirow{2}{*}{TCN}& \multicolumn{9}{c|}{Transformers} &\multicolumn{2}{c|}{MLP} & \multirow{2}{*}{TimesNet} & \multicolumn{2}{c|}{LLM} & \multirow{2}{*}{LLM-TS}\\
&XGB&Roc&LSTNet& LSSL& & Trans.& Re.& In.& Pyra.& Auto.& Station.& FED.& ETS.& Flow. &DL &LTS. & & GPT4TS & TEST &  \\

\midrule
Ethanol&$43.7 $&$45.2 $&$39.9 $&$31.1 $&$28.9 $&$32.7 $&$31.9 $&$31.6 $&$30.8 $&$31.6 $&$32.7 $&$31.2 $&$28.1 $&$33.8 $&$32.6 $&$29.7 $&$30.4 $&$26.2$&$25.1 $&$31.9$\\
FaceD&$63.3 $&$64.7 $&$65.7 $&$66.7 $&$52.8 $&$67.3 $&$68.6 $&$67.0 $&$65.7 $&$68.4 $&$68.0 $&$66.0 $&$66.3 $&$67.6 $&$68.0 $&$67.5 $&$68.6 $&$67.8$&$50.1 $&$68.9$\\
HandW&$15.8 $&$58.8 $&$25.8 $&$24.6 $&$53.3 $&$32.0 $&$27.4 $&$32.8 $&$29.4 $&$36.7 $&$31.6 $&$28.0 $&$32.5 $&$33.8 $&$27.0 $&$26.1 $&$32.1 $&$28.9$&$20.1 $&$32.7$\\
HeartB&$73.2 $&$75.6 $&$77.1 $&$72.7 $&$75.6 $&$76.1 $&$77.1 $&$80.5 $&$75.6 $&$74.6 $&$73.7 $&$73.7 $&$71.2 $&$77.6 $&$75.1 $&$75.1 $&$77.6 $&$72.2$&$73.7 $&$77.1$\\
JapanV&$86.5 $&$96.2 $&$98.1 $&$98.4 $&$98.9 $&$98.7 $&$97.8 $&$98.9 $&$98.4 $&$96.2 $&$99.2 $&$98.4 $&$95.9 $&$98.9 $&$96.2 $&$96.2 $&$97.2 $&$98.4$&$78.4 $&$98.1$\\
PEMS&$98.3 $&$75.1 $&$86.7 $&$86.1 $&$68.8 $&$82.1 $&$82.7 $&$81.5 $&$83.2 $&$82.7 $&$87.3 $&$80.9 $&$86.0 $&$83.8 $&$75.1 $&$88.4 $&$89.6 $&$79.2$&$59.5 $&$90.8$\\
SCP1&$84.6 $&$90.8 $&$84.0 $&$90.8 $&$84.6 $&$92.2 $&$90.4 $&$90.1 $&$88.1 $&$84.0 $&$89.4 $&$88.7 $&$89.6 $&$92.5 $&$87.3 $&$89.8 $&$90.4 $&$90.1$&$84.0 $&$91.8$\\
SCP2&$48.9 $&$53.3 $&$52.8 $&$52.2 $&$55.6 $&$53.9 $&$56.7 $&$53.3 $&$53.3 $&$50.6 $&$57.2 $&$54.4 $&$55.0 $&$56.1 $&$50.5 $&$51.1 $&$57.1 $&$50.0$&$54.4 $&$57.8$\\
SpokenA&$69.6 $&$71.2 $&$100.0 $&$100.0 $&$95.6 $&$98.4 $&$97.0 $&$100.0 $&$99.6 $&$100.0 $&$100.0 $&$100.0 $&$100.0 $&$98.8 $&$81.4 $&$100.0 $&$98.6 $&$97.9$&$82.1 $&$98.6$\\
UWave&$75.9 $&$94.4 $&$87.8 $&$85.9 $&$88.4 $&$85.6 $&$85.6 $&$85.6 $&$83.4 $&$85.9 $&$87.5 $&$85.3 $&$85.0 $&$86.6 $&$82.1 $&$80.3 $&$85.5 $&$85.6$&$84.4 $&$86.6$\\
\midrule
Avg &$66.0 $&$72.5 $&$71.8 $&$70.9 $&$70.3 $&$71.9 $&$71.5 $&$72.1 
$&$70.8 $&$71.1 $&$72.7 $&$70.7 $&$71.0 $&\underline{$73.0 $}&$67.5 $&$70.4 $&${72.7} $&${69.5}$&$61.2 $&\boldsymbol{$73.4$}\\
\bottomrule

\end{tabular}
}
\end{small}
\end{center}
\end{table*}

\subsection{Full Results of Anamoly Detection}
\label{appendix:anomaly_full}

Full results for anamoly detection are detailed in Table~\ref{tab:anomaly_full}.
\begin{table*}[htb]
\caption{Full results for the anomaly detection.}
\label{tab:anomaly_full}
\vskip 0.15in
\begin{center}
\begin{small}
\scalebox{0.65}{
\begin{threeparttable}[b]
\begin{tabular}{c|ccc|ccc|ccc|ccc|ccc|c}
\toprule

Methods &
\multicolumn{3}{c|}{SMD} & \multicolumn{3}{c|}{MSL} & \multicolumn{3}{c|}{SMAP}& \multicolumn{3}{c|}{SWaT} &\multicolumn{3}{c|}{PSM} & Avg F1 \\
Metrics&P&R&F1&P&R&F1&P&R&F1&P&R&F1&P&R&F1&\%  \\

\midrule
LLM-TS &$ 88.09$&$ 81.54$&$ 84.69$&$ 89.04$&$ 74.49$&$ 81.11$&$ 89.95$&$ 56.51$&$ 69.41$&$ 91.16$&$ 95.40$&$93.23 $&$ 98.44$&$ 96.45$&$ 97.43 $&$ 85.17 $\\
TimesNet &$ 87.93$&$ 81.45$&$ 84.57$&$ 88.62$&$ 73.48$&$ 80.34$&$ 89.59$&$ 56.35$&$ 69.18$&$ 91.00$&$ 95.33$&$ 93.12 $&$ 98.40$&$ 96.18$&$ 97.27 $&$ 84.90 $\\
GPT4TS&${87.70}$&${81.19}$&${84.32}$&$82.15$&${81.32}$&${81.73}$&${90.04}$&${55.75}$&${68.86}$&${92.12}$&${93.06}$&${92.59}$&${98.37}$&$96.34$&$97.34$&${84.97}$\\
PatchTST&$87.26$&$82.14$&$84.62$&$88.34$&$70.96$&$78.70$&$90.64$&$55.46$&$68.82$&$91.10$&$80.94$&$85.72$&$98.84$&$93.47$&$96.08$&$82.79$\\
ETSformer&$87.44$&$79.23$&$83.13$&$85.13$&$84.93$&$85.03$&$92.25$&$55.75$&$69.50$&$90.02$&$80.36$&$84.91$&$99.31$&$85.28$&$91.76$&$82.87$\\
FEDformer&$87.95$&$82.39$&$85.08$&$77.14$&$80.07$&$78.57$&$90.47$&$58.10$&$70.76$&$90.17$&$96.42$&$93.19$&$97.31$&$97.16$&$97.23$&$84.97$\\
LightTS&$87.10$&$78.42$&$82.53$&$82.40$&$75.78$&$78.95$&$92.58$&$55.27$&$69.21$&$91.98$&$94.72$&$93.33$&$98.37$&$95.97$&$97.15$&$84.23 $\\
DLinear&$83.62$&$71.52$&$77.10$&$84.34$&$85.42$&$84.88$&$92.32$&$55.41$&$69.26$&$80.91$&$95.30$&$87.52$&$98.28$&$89.26$&$93.55$&$82.46 $\\
Stationary&$88.33$&$81.21$&$84.62$&$68.55$&$89.14$&$77.50$&$89.37$&$59.02$&$71.09$&$68.03$&$96.75$&$79.88$&$97.82$&$96.76$&$97.29$&$82.08 $\\
Autoformer&$88.06$&$82.35$&$85.11$&$77.27$&$80.92$&$79.05$&$90.40$&$58.62$&$71.12$&$89.85$&$95.81$&$92.74$&$99.08$&$88.15$&$93.29$&$84.26 $\\
Pyraformer&$85.61$&$80.61$&$83.04$&$83.81$&$85.93$&$84.86$&$92.54$&$57.71$&$71.09$&$87.92$&$96.00$&$91.78$&$71.67$&$96.02$&$82.08$&$82.57 $\\
Anomaly Transformer&$88.91$&$82.23$&$85.49$&$79.61$&$87.37$&$83.31$&$91.85$&$58.11$&$71.18$&$72.51$&$97.32$&$83.10$&$68.35$&$94.72$&$79.40$&$80.50 $\\
Informer&$86.60$&$77.23$&$81.65$&$81.77$&$86.48$&$84.06$&$90.11$&$57.13$&$69.92$&$70.29$&$96.75$&$81.43$&$64.27$&$96.33$&$77.10$&$78.83 $\\
Reformer&$82.58$&$69.24$&$75.32$&$85.51$&$83.31$&$84.40$&$90.91$&$57.44$&$70.40$&$72.50$&$96.53$&$82.80$&$59.93$&$95.38$&$73.61$&$77.31 $\\
Transformer&$83.58$&$76.13$&$79.56$&$71.57$&$87.37$&$78.68$&$89.37$&$57.12$&$69.70$&$68.84$&$96.53$&$80.37$&$62.75$&$96.56$&$76.07$&$76.88 $\\

\bottomrule

\end{tabular}

\end{threeparttable}

}
\end{small}
\end{center}
\vskip -0.1in
\end{table*}

\begin{table*}[htb]
\caption{Different traditional models. We use prediction length $O \in \{96, 192, 336, 720\}$ for ILI and $O \in \{24, 36, 48, 60\}$ for others. }
\label{tab: trad_ablation}
\vskip 0.15in
\begin{center}
\begin{small}
\scalebox{0.70}{
\setlength\tabcolsep{3pt}
\begin{tabular}{c|c|cc|cc|cc|cc|cc|cc|cc|cc}
\toprule

\multicolumn{2}{c|}{Methods}&\multicolumn{2}{c|}{PatchTST} &\multicolumn{2}{c|}{PatchTST INT}&\multicolumn{2}{c|}{ETSformer}&\multicolumn{2}{c|}{ETS INT}&\multicolumn{2}{c|}{Stationary}&\multicolumn{2}{c|}{Stat INT}&\multicolumn{2}{c|}{FreTS} &\multicolumn{2}{c}{FreTS INT} \\

\midrule

\multicolumn{2}{c|}{Metric}  & MSE & MAE& MSE & MAE& MSE  & MAE& MSE  & MAE& MSE & MAE& MSE & MAE& MSE  & MAE& MSE  & MAE  \\
\midrule

\multirow{5}{*}{\rotatebox{90}{$Weather$}}
& 96  & 0.174 & 0.216 & 0.172 & 0.214 & 0.196 & 0.282 & 0.200 & 0.285& 0.178 & 0.226 & 0.201 & 0.246 & 0.187 & 0.243 & 0.179 & 0.235 \\
& 192 & 0.222 & 0.258 & 0.219 & 0.255 & 0.282 & 0.364 & 0.278 & 0.361& 0.235 & 0.278 & 0.238 & 0.280 & 0.227 & 0.274 & 0.221 & 0.278\\
& 336 & 0.280 & 0.298 & 0.279 & 0.298 & 0.344 & 0.409 & 0.322 & 0.382& 0.327 & 0.339 & 0.312 & 0.329 & 0.281 & 0.325 & 0.276 & 0.320 \\
& 720 & 0.356 & 0.349 & 0.356 & 0.348 & 0.430 & 0.472 & 0.427 & 0.470& 0.387 & 0.383 & 0.386 & 0.383 & 0.352 & 0.382 & 0.344 & 0.376 \\
& Avg & 0.258 & 0.280 & 0.257 & 0.279 & 0.313 & 0.382 & 0.307 & 0.375& 0.282 & 0.307 & 0.284 & 0.309 & 0.262 & 0.306 & 0.255 & 0.302 \\
\midrule

\multirow{5}{*}{\rotatebox{90}{$ETTh1$}}
& 96  & 0.381 & 0.398 & 0.382 & 0.401& 0.554 & 0.536 & 0.550 & 0.532& 0.534 & 0.499 & 0.523 & 0.486& 0.398 & 0.412& 0.395 & 0.409 \\
& 192 & 0.421 & 0.426 & 0.422 & 0.428& 0.686 & 0.619 & 0.690 & 0.621& 0.639 & 0.560 & 0.609 & 0.560& 0.454 & 0.449& 0.455 & 0.451 \\
& 336 & 0.464 & 0.449 & 0.460 & 0.441& 0.869 & 0.730 & 0.868 & 0.728& 0.790 & 0.648 & 0.780 & 0.634& 0.512 & 0.483& 0.502 & 0.474 \\
& 720 & 0.527 & 0.500 & 0.510 & 0.496& 1.085 & 0.849 & 1.054 & 0.830& 0.706 & 0.620 & 0.701 & 0.606& 0.572 & 0.547& 0.560 & 0.530 \\
& Avg & 0.448 & 0.443 & 0.444 & 0.442& 0.799 & 0.684 & 0.791 & 0.678& 0.667 & 0.582 & 0.653 & 0.572& 0.484 & 0.473& 0.478 & 0.466 \\
\midrule

\multirow{5}{*}{\rotatebox{90}{$ETTm1$}}
& 96  & 0.332 & 0.368 & 0.332 & 0.372 & 0.526 & 0.495 & 0.424 & 0.434 & 0.417 & 0.417 & 0.412 & 0.410 & 0.340 & 0.375 & 0.339 & 0.374\\
& 192 & 0.368 & 0.388 & 0.367 & 0.388 & 0.565 & 0.538 & 0.458 & 0.461 & 0.446 & 0.437 & 0.445 & 0.435 & 0.395 & 0.408 & 0.384 & 0.399\\
& 336 & 0.397 & 0.405 & 0.396 & 0.405 & 0.658 & 0.603 & 0.537 & 0.519 & 0.582 & 0.507 & 0.570 & 0.491 & 0.431 & 0.433 & 0.420 & 0.423 \\
& 720 & 0.457 & 0.445 & 0.460 & 0.446 & 0.801 & 0.696 & 0.802 & 0.696 & 0.661 & 0.546 & 0.660 & 0.546 & 0.494 & 0.470 & 0.484 & 0.462 \\
& Avg & 0.389 & 0.402 & 0.389 & 0.403 & 0.638 & 0.583 & 0.555 & 0.528 & 0.527 & 0.477 & 0.522 & 0.471 & 0.415 & 0.422 & 0.407 & 0.415\\
\midrule

\multirow{5}{*}{\rotatebox{90}{$ILI$}}
& 24 & 2.229 & 0.894 & 2.172 & 0.856 & 4.043 & 1.410 & 3.607 & 1.305 & 2.722 & 1.024 & 1.905 & 0.872 & 3.226 & 1.231 & 3.202 & 1.213\\
& 36 & 2.330 & 0.925 & 2.347 & 0.978 & 3.809 & 1.358 & 3.705 & 1.315 & 3.026 & 1.071 & 2.790 & 1.068 & 3.363 & 1.259 & 3.000 & 1.173\\
& 48 & 2.140 & 0.894 & 1.984 & 0.869 & 3.851 & 1.351 & 3.714 & 1.309 & 2.622 & 1.032 & 2.132 & 0.900 & 3.456 & 1.285 & 3.132 & 1.213\\
& 60 & 2.037 & 0.912 & 1.770 & 0.831 & 3.983 & 1.349 & 3.935 & 1.350 & 2.520 & 1.035 & 1.991 & 0.901 & 3.749 & 1.340 & 3.298 & 1.243\\
& Avg& 2.184 & 0.906 & 2.068 & 0.884 & 3.922 & 1.367 & 3.740 & 1.320 & 2.722 & 1.041 & 2.205 & 0.935 & 3.449 & 1.279 & 3.158 & 1.211\\
\bottomrule
\end{tabular}
}
\end{small}
\end{center}
\vskip -0.1in
\end{table*}
\begin{table*}[htb]
\caption{Different LLM embeddings. We use prediction length $O \in \{96, 192, 336, 720\}$ for ILI and $O \in \{24, 36, 48, 60\}$ for others. }
\label{tab: lm_ablation}
\begin{center}
\begin{small}
\scalebox{0.70}{
\setlength\tabcolsep{3pt}
\begin{tabular}{c|c|cc|cc|cc|cc|cc}
\toprule

\multicolumn{2}{c|}{Methods}&\multicolumn{2}{c|}{LLM-TS (LLaMA)}&\multicolumn{2}{c|}{LLaMA w/o text}&\multicolumn{2}{c|}{GPT2}&\multicolumn{2}{c|}{BERT}&\multicolumn{2}{c}{No LLM} \\

\midrule

\multicolumn{2}{c|}{Metric} & MSE  & MAE& MSE  & MAE & MSE & MAE& MSE & MAE& MSE  & MAE  \\
\midrule

\multirow{5}{*}{\rotatebox{90}{$Weather$}}
& 96  & 0.166 & 0.217 & 0.170 & 0.218& 0.168 & 0.218 & 0.167 & 0.217 & 0.168 & 0.218 \\
& 192 & 0.229 & 0.269 & 0.227 & 0.266& 0.226 & 0.267 & 0.229 & 0.270 & 0.227 & 0.268 \\
& 336 & 0.278 & 0.302 & 0.295 & 0.314& 0.292 & 0.310 & 0.283 & 0.305 & 0.298 & 0.318 \\
& 720 & 0.354 & 0.351 & 0.360 & 0.354& 0.359 & 0.354 & 0.360 & 0.354 & 0.361 & 0.356 \\
& Avg & 0.257 & 0.285 & 0.263 & 0.288& 0.261 & 0.287 & 0.260 & 0.287 & 0.264 & 0.290 \\
\midrule

\multirow{5}{*}{\rotatebox{90}{$ETTh1$}}
& 96  & 0.403 & 0.420 & 0.409 & 0.427& 0.408 & 0.426 & 0.402 & 0.421 & 0.402 & 0.422 \\
& 192 & 0.440 & 0.441 & 0.445 & 0.445& 0.442 & 0.444 & 0.452 & 0.450 & 0.459 & 0.455 \\
& 336 & 0.471 & 0.457 & 0.490 & 0.472& 0.487 & 0.467 & 0.494 & 0.472 & 0.471 & 0.457 \\
& 720 & 0.503 & 0.487 & 0.518 & 0.496& 0.517 & 0.494 & 0.520 & 0.497 & 0.535 & 0.507 \\
& Avg & 0.454 & 0.451 & 0.465 & 0.460& 0.464 & 0.458 & 0.467 & 0.460 & 0.467 & 0.460 \\
\midrule

\multirow{5}{*}{\rotatebox{90}{$ETTm1$}}
& 96  & 0.329 & 0.371 & 0.350 & 0.387& 0.338 & 0.370 & 0.340 & 0.375 & 0.341 & 0.377 \\
& 192 & 0.380 & 0.398 & 0.383 & 0.398& 0.392 & 0.404 & 0.401 & 0.408 & 0.404 & 0.413 \\
& 336 & 0.418 & 0.425 & 0.423 & 0.426& 0.416 & 0.423 & 0.414 & 0.421 & 0.432 & 0.428 \\
& 720 & 0.476 & 0.440 & 0.467 & 0.449& 0.477 & 0.454 & 0.470 & 0.445 & 0.468 & 0.449 \\
& Avg & 0.401 & 0.409 & 0.406 & 0.415& 0.406 & 0.413 & 0.406 & 0.412 & 0.411 & 0.417 \\
\midrule

\multirow{5}{*}{\rotatebox{90}{$ILI$}}
& 24 & 1.921 & 0.898 & 1.998 & 0.929& 1.997 & 0.929 & 1.917 & 0.915 & 2.170 & 0.947 \\
& 36 & 2.151 & 0.933 & 2.422 & 0.957& 2.333 & 0.958 & 2.431 & 1.004 & 2.093 & 0.889 \\
& 48 & 2.062 & 0.892 & 2.198 & 0.964& 2.269 & 0.937 & 2.333 & 0.961 & 2.418 & 0.959 \\
& 60 & 1.759 & 0.853 & 2.072 & 0.948& 2.077 & 0.921 & 2.089 & 0.926 & 2.203 & 0.971 \\
& Avg & 1.973 & 0.894 & 2.173 & 0.950& 2.169 & 0.936 & 2.193 & 0.952 & 2.221 & 0.942 \\
\bottomrule
\end{tabular}
}
\end{small}
\end{center}
\vskip -0.1in
\end{table*}

\subsection{Further Ablation Studies}
\label{appendix: further_ablation}

\noindent \textbf{Mutual Information Estimator.}
In the main paper, we utilize the Jensen-Shannon mutual information (MI) estimator. Additionally, we explore the Mutual Information Neural Estimator (MINE)~\cite{hjelm2018learning}. We evaluate both estimators on two tasks, ETTh1 and ETTm1, with results averaged over four prediction lengths.
For ETTh1, the MSE and MAE using the original Jensen-Shannon estimator are $0.454$ and $0.451$, respectively, compared to $0.460$ and $0.457$ with MINE. For ETTm1, the MSE and MAE are $0.401$ and $0.409$ with the original estimator, and $0.402$ and $0.410$ with MINE.
These comparisons highlight the robustness of our method across different mutual information estimators.

\begin{figure*}[htb]
\centering
\begin{minipage}[t]{.49\textwidth}
  \centering
    \includegraphics[width=1.00\columnwidth]{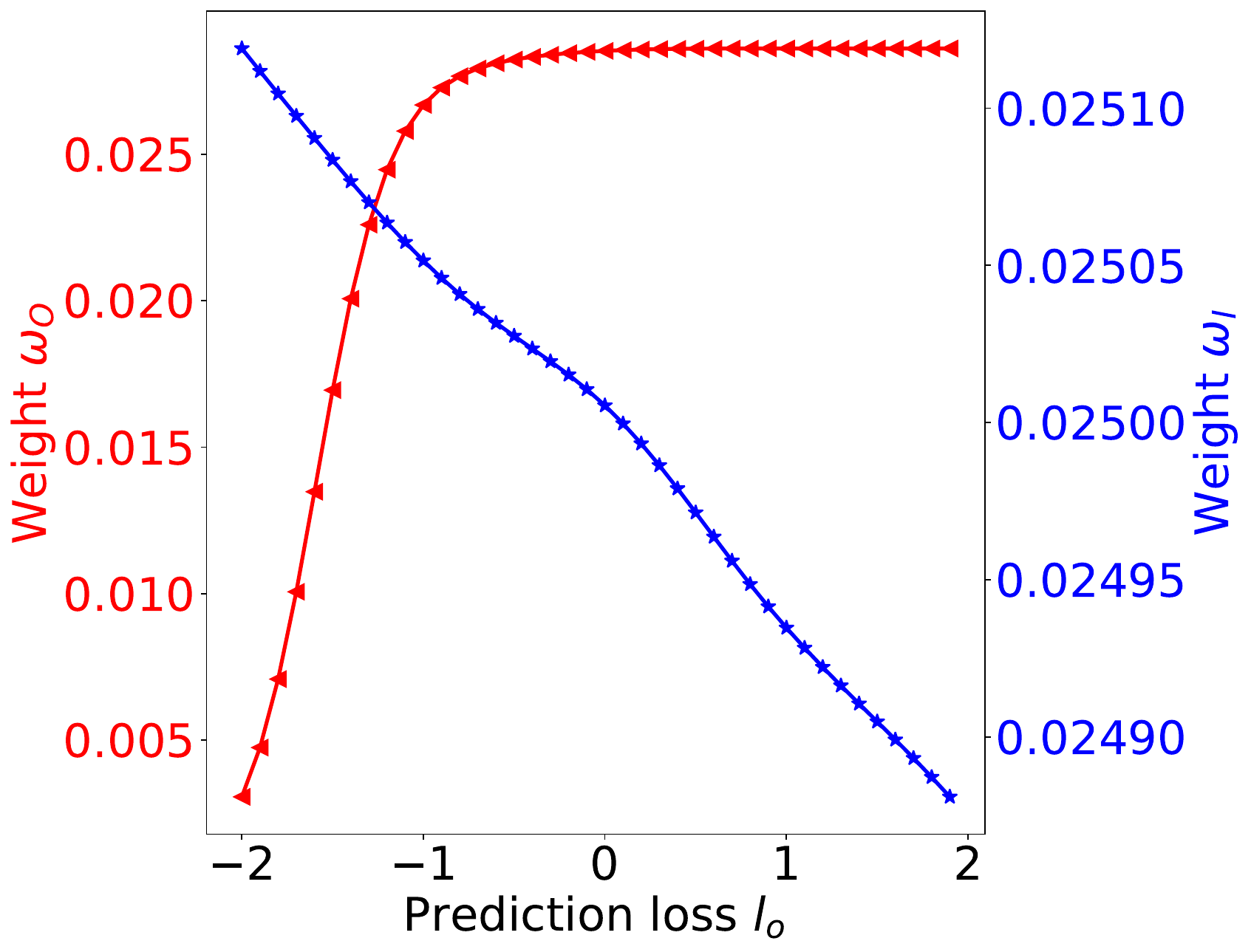}
    \captionsetup{width=1.05\linewidth}
    \caption{Forecasting.
    }
    \label{fig: forecast}
\end{minipage}
\begin{minipage}[t]{.49\textwidth}
  \centering
    \includegraphics[width=1.0\columnwidth]{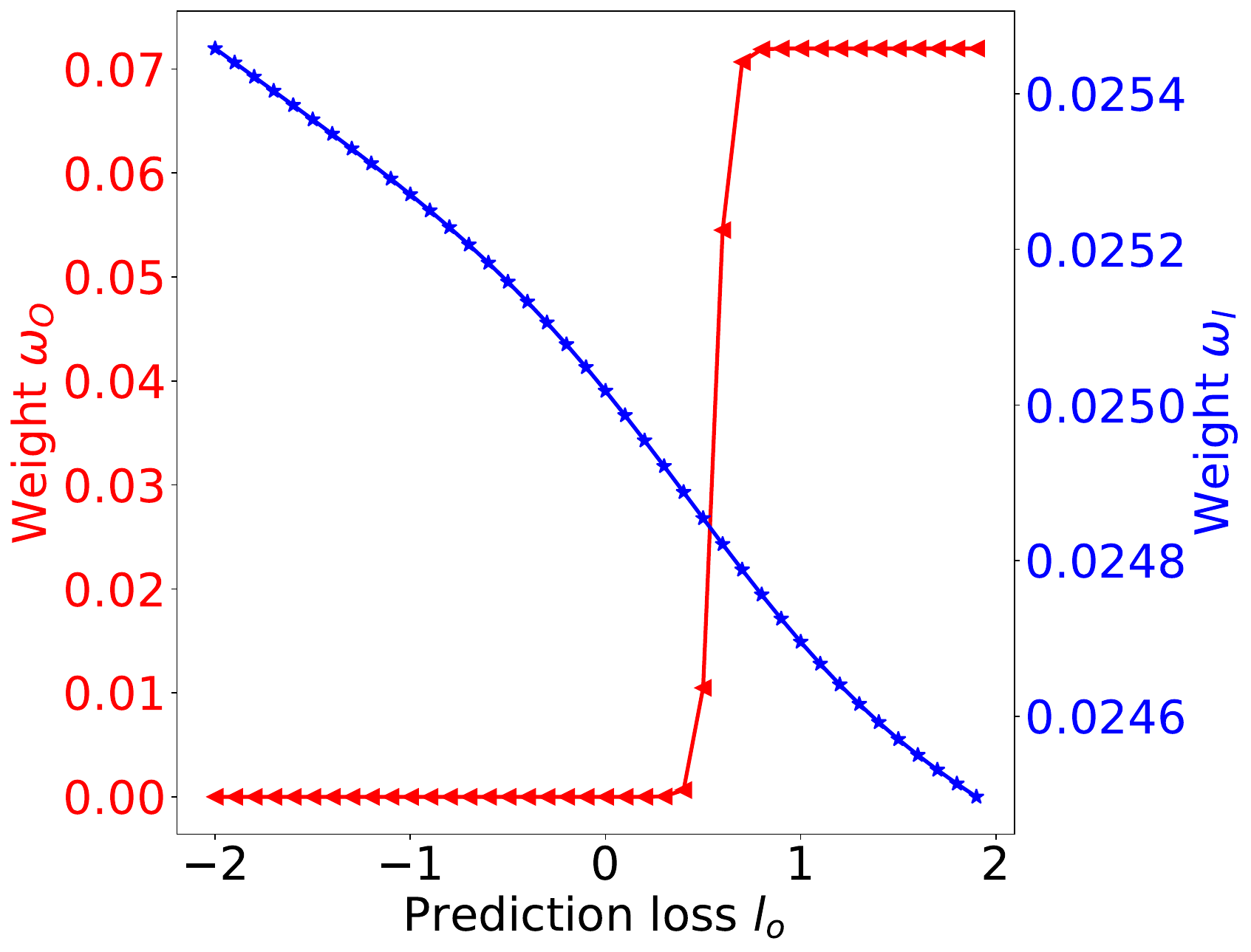}
    \captionsetup{width=.90\linewidth}
    %\caption{a}
    \caption{Anomaly detection}
    %\caption{The ratio of the high-scoring design score for different predefined target scores to that of the score $10$.  }
    \label{fig: anomaly}
\end{minipage}%
\end{figure*}

\begin{figure*}[htb]
\centering
\begin{minipage}[t]{.49\textwidth}
  \centering
    \includegraphics[width=1.00\columnwidth]{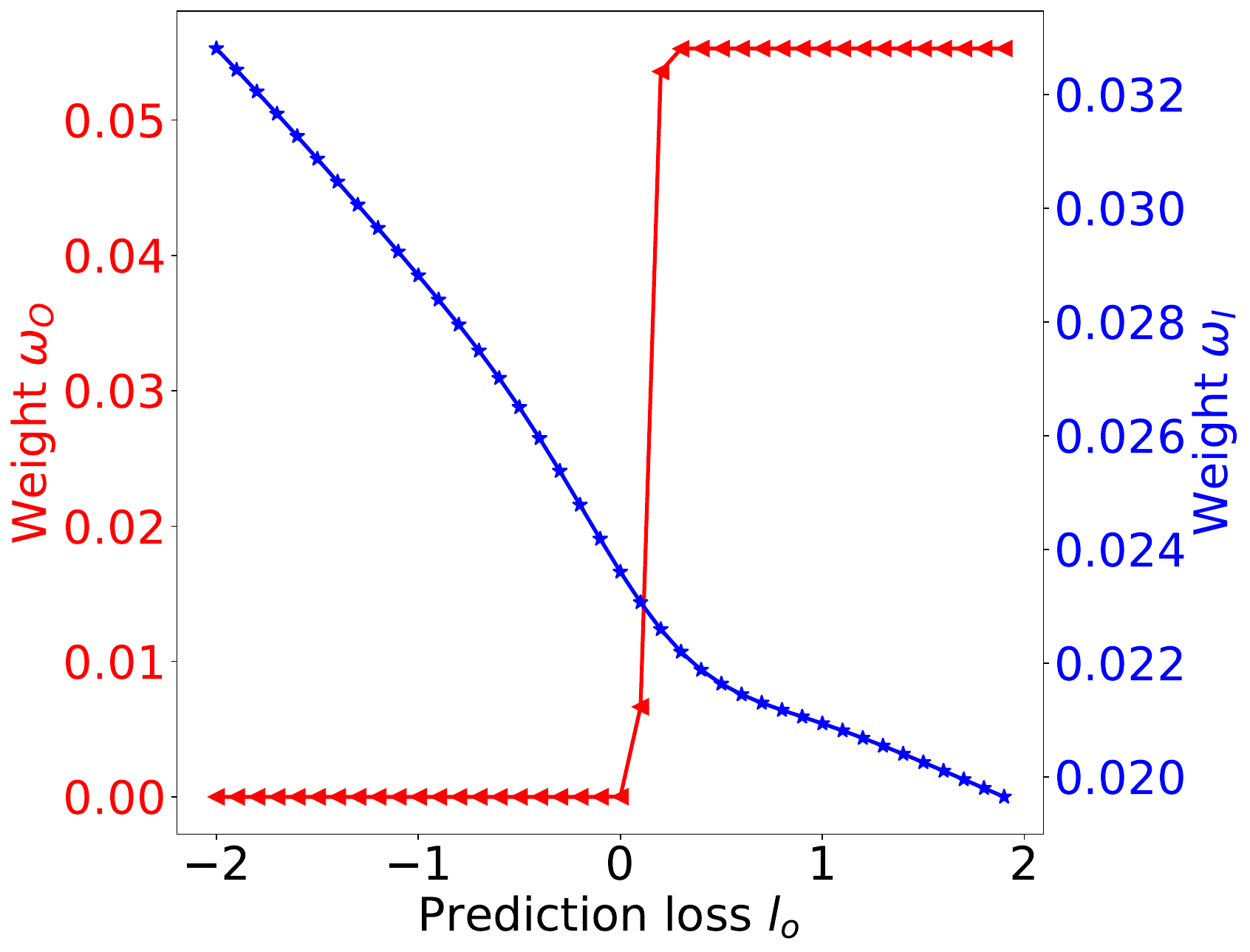}
    \captionsetup{width=1.05\linewidth}
    \caption{imputation.
    }
    \label{fig: imputation}
\end{minipage}
\begin{minipage}[t]{.49\textwidth}
  \centering
    \includegraphics[width=1.0\columnwidth]{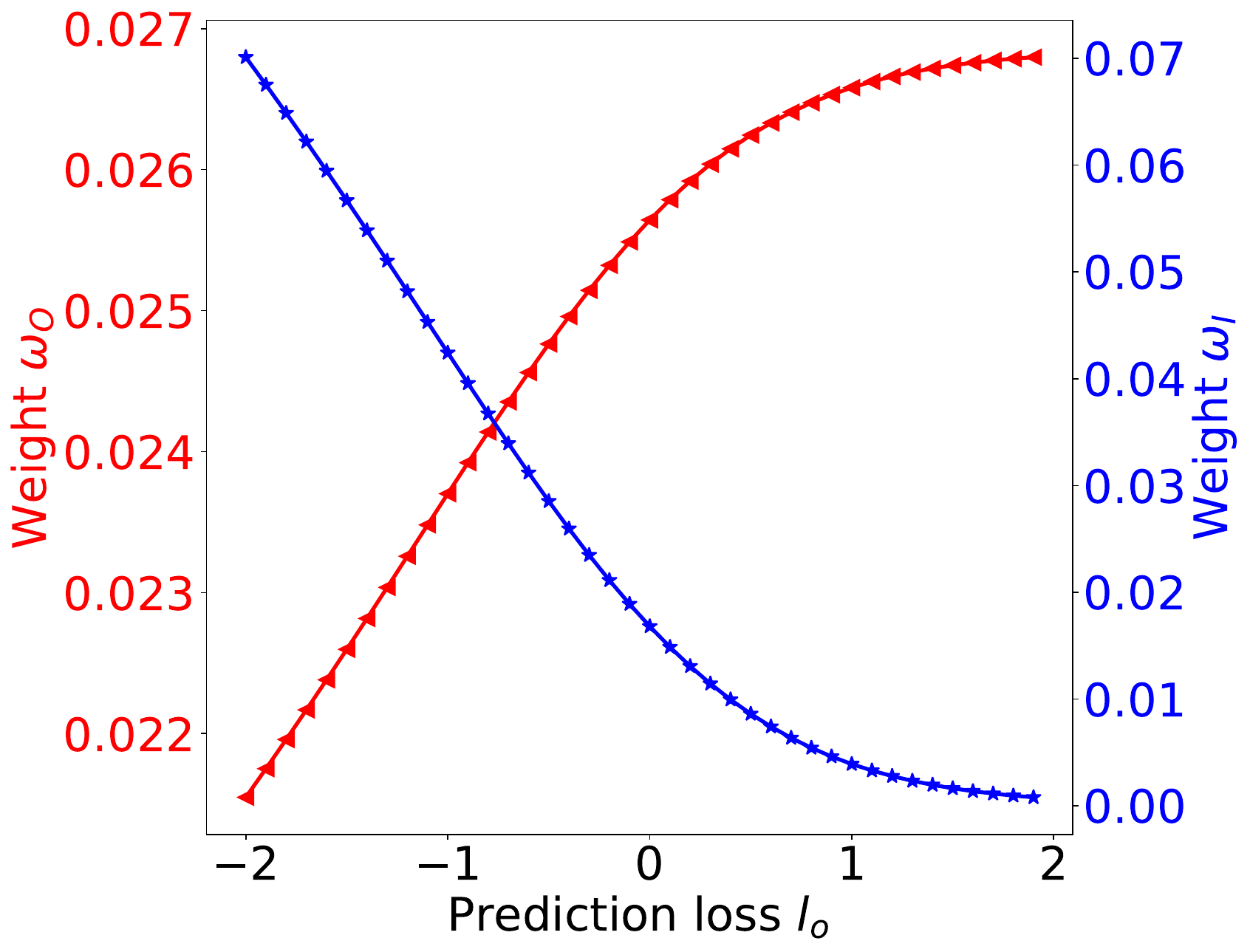}
    \captionsetup{width=.90\linewidth}
    %\caption{a}
    \caption{classification}
    %\caption{The ratio of the high-scoring design score for different predefined target scores to that of the score $10$.  }
    \label{fig: classification}
\end{minipage}%
\end{figure*}

\noindent \textbf{Sample Reweighting Illustration.}
Figures~\ref{fig: anomaly}, \ref{fig: forecast}, \ref{fig: imputation}, and \ref{fig: classification} display the learned weighting network applied to various datasets: MSL for anomaly detection, Weather for forecasting, ETTh1 for imputation, and PEMS-SF for classification. These visualizations corroborate our hypothesis: the sample weight $\omega_O$ increases with the prediction loss $l_O$, while the weight $\omega_I$ decreases as $l_O$ increases. This observed pattern supports the efficacy of our reweighting strategy.

\noindent \textbf{Static Weighting Scheme.}
We also explore a static weighting scheme as a contrast to the dynamic weighting used in our sample reweighting module. 
This scheme balances the prediction loss and mutual information loss, with a ratio of $0.0$ representing pure prediction loss and $1.0$ representing pure mutual information loss. 
As shown in Table~\ref{tab: static_weight}, the static approach underperforms relative to our dynamic sample weighting module, demonstrating the superior effectiveness of our method.

\subsubsection{Comprehensive Results.}
The detailed performance of various traditional TS models and LLMs is presented in Table~\ref{tab: trad_ablation} and Table~\ref{tab: lm_ablation}.

\begin{table*}[htb]
\caption{Static Weighting Scheme with Different ratios.  }
\label{tab: static_weight}
\vskip 0.15in
\begin{center}
\begin{small}
\scalebox{0.70}{
\setlength\tabcolsep{3pt}
\begin{tabular}{c|cc|cc|cc|cc|cc|cc|cc|cc|cc}
\toprule

\multicolumn{1}{c|}{Ratio}&\multicolumn{2}{c|}{$0.0$} &\multicolumn{2}{c|}{$0.2$}&\multicolumn{2}{c|}{$0.4$}&\multicolumn{2}{c|}{$0.6$}&\multicolumn{2}{c|}{$0.8$}&\multicolumn{2}{c|}{$1.0$}&\multicolumn{2}{c|}{Ours} \\

\midrule

\multicolumn{1}{c|}{Metric}  & MSE & MAE& MSE & MAE& MSE  & MAE& MSE  & MAE& MSE & MAE& MSE & MAE& MSE & MAE  \\
\midrule

\multirow{1}{*}{\rotatebox{0}{$ETTh1$}}
 & $0.478$ & $0.468$ & $0.471$ & $0.459$ & $0.465$ & $0.462$ & $0.470$ & $0.463$& $0.473$ & $0.450$ & $0.471$ & $0.463$ & $0.454$ & $0.451$ \\
\midrule

\multirow{1}{*}{\rotatebox{0}{$ETTm1$}}
& $0.415$ & $0.417$ & $0.408$ & $0.414$& $0.405$ & $0.412$ & $0.406$ & $0.412$& $0.417$ & $0.419$ & $0.416$ & $0.419$ & $0.401$ & $0.409$\\

\bottomrule
\end{tabular}
}
\end{small}
\end{center}
\vskip -0.1in
\end{table*}

% DLinear from TimesNet
% PatchTST (illness) from TimeMixer

\end{document}